\documentclass[final,5p,times,twocolumn,numbers]{elsarticle}

\usepackage{amsmath,amssymb,amsfonts}

\usepackage{graphicx}
\usepackage{float}
\usepackage{caption}

\usepackage{booktabs}
\usepackage{multirow}
\usepackage{array}
\usepackage{tabularx}
\usepackage{longtable}
\usepackage{supertabular}
\usepackage{threeparttable}
\usepackage{rotating}
\usepackage{pdflscape}

\usepackage{xcolor}
\usepackage{orcidlink}

\tikzset{
    y27box/.style={
        rectangle,
        draw=YoloDark,
        rounded corners=2pt,
        line width=0.8pt,
        minimum height=0.72cm,
        align=center,
        font=\scriptsize
    },
    y27api/.style={
        y27box,
        minimum width=5.6cm,
        minimum height=0.90cm,
        fill=YoloNavy,
        draw=YoloNavy,
        text=white,
        font=\scriptsize\bfseries
    },
    y27cnn/.style={
        y27box,
        minimum width=2.8cm,
        fill=YoloAmber,
        draw=YoloOrange,
        text=YoloDark,
        font=\scriptsize\bfseries
    },
    y27cnnhead/.style={
        y27box,
        minimum width=3.4cm,
        minimum height=0.85cm,
        fill=YoloOrange,
        draw=YoloOrange,
        text=white,
        font=\scriptsize\bfseries
    },
    y27query/.style={
        y27box,
        minimum width=3.1cm,
        fill=YoloCyan,
        draw=YoloBlue,
        text=YoloDark,
        font=\scriptsize\bfseries
    },
    y27queryhead/.style={
        y27box,
        minimum width=3.6cm,
        minimum height=0.85cm,
        fill=YoloBlue,
        draw=YoloBlue,
        text=white,
        font=\scriptsize\bfseries
    },
    y27outputcnn/.style={
        y27box,
        minimum width=3.3cm,
        fill=YoloRed,
        draw=YoloRed,
        text=white,
        font=\scriptsize\bfseries
    },
    y27outputquery/.style={
        y27box,
        minimum width=3.3cm,
        fill=YoloGreen,
        draw=YoloGreen,
        text=white,
        font=\scriptsize\bfseries
    },
    y27arrow/.style={
        -{Stealth[length=2.2mm]},
        line width=0.95pt,
        draw=YoloDark
    },
    y27cnnarrow/.style={
        -{Stealth[length=2.2mm]},
        line width=1.0pt,
        draw=YoloOrange
    },
    y27queryarrow/.style={
        -{Stealth[length=2.2mm]},
        line width=1.0pt,
        draw=YoloBlue
    },
    y27trainarrow/.style={
        -{Stealth[length=2.2mm]},
        dashed,
        line width=0.95pt,
        draw=YoloRed
    }
}
\usepackage{tikz}
\usetikzlibrary{
    mindmap,
    shadows,
    positioning,
    backgrounds,
    arrows.meta,
    fit,
    shapes.geometric,
    trees
}

\usepackage{pgfplots}
\usepackage{pgfplotstable}
\pgfplotsset{compat=1.18}

\usepackage{hyperref}
\usepackage{url}
\usepackage{xcolor}
\usepackage{tikz}

\usetikzlibrary{
    arrows.meta,
    positioning,
    calc,
    fit,
    backgrounds,
    shapes.geometric
}

\definecolor{YoloNavy}{HTML}{17365D}
\definecolor{YoloBlue}{HTML}{1565C0}
\definecolor{YoloCyan}{HTML}{D9F2FA}
\definecolor{YoloOrange}{HTML}{E65100}
\definecolor{YoloAmber}{HTML}{FFF0D5}
\definecolor{YoloRed}{HTML}{B71C1C}
\definecolor{YoloGreen}{HTML}{2E7D32}
\definecolor{YoloGray}{HTML}{F2F4F7}
\definecolor{YoloDark}{HTML}{202124}

\usepackage{hyperref}

\journal{Elsevier}

\begin{document}


\begin{frontmatter}


\title{\textbf{Optimizing YOLO27, YOLO26, YOLO11, and YOLOv8 for Fine-Grained Small-Object Detection and Segmentation in Complex Orchard Environments}}


\author[1]{\textbf{Ranjan Sapkota}\,\orcidlink{0000-0002-5417-6744}
\corref{cor1}}
\ead{rs2672@cornell.edu}

\author[1]{\textbf{Manoj Karkee}\,\orcidlink{0000-0001-5337-4848}
\corref{cor1}}
\ead{mk2684@cornell.edu}


\affiliation[1]{
    organization={Cornell University, AgRobotics Lab},
    addressline={134 Riley-Robb Hall},
    city={Ithaca},
    state={NY},
    postcode={14853},
    country={USA}
}


\cortext[cor1]{Corresponding authors: Ranjan Sapkota and Manoj Karkee.}

\begingroup
\setlength{\leftskip}{0.5cm}
\setlength{\rightskip}{-0.5cm}

\begin{abstract}

This study presents an architectural and experimental cross-generation analysis of Ultralytics YOLO27 (YOLOv27), YOLO26 (YOLOv26), YOLO11 (YOLOv11), and YOLOv8 for fine-grained robotic orchard perception. Fine-grained detection and instance segmentation of early-stage fruit anatomy remain challenging in complex orchard environments because of limited pixel footprints, green-on-green similarity, occlusion, and substantial scale variation. Because YOLO27 has been announced but its public implementation and trainable segmentation models are not yet available, the present study provides an architectural analysis of YOLO27, while controlled experiments benchmark YOLOv8, YOLO11, and YOLO26; YOLO27 experiments will be incorporated following public model availability. Five model scales-nano (n), small (s), medium (m), large (l), and extra-large (x)-were evaluated for fruitlet, calyx, and peduncle detection and segmentation using conventional $640\times640$ and small-object-focused $960\times960$ configurations, yielding 30 experiments. YOLO11s-960 achieved the highest observed mask mAP$_{50:95}$ (0.402) and box mAP$_{50:95}$ (0.426), whereas YOLO26s-960 achieved comparable values of 0.397 and 0.425 with only 10.37~M parameters and 34.1~GFLOPs. Peduncle remained the most challenging class, and increasing model capacity did not consistently improve accuracy. Overall, compact-to-moderate YOLO models combined with small-object-focused training provided favorable accuracy--efficiency trade-offs, establishing a reproducible benchmark for fine-grained agricultural robotic perception. Code, trained models, and experimental configurations are publicly available, and will be updated through our \href{https://github.com/rnjnspkt/Optimizing-and-Comparing-Ultralytics-YOLO26-YOLO11-and-YOLOv8-for-Small-Object-Detection-and-Seg}{GitHub repository}.
\end{abstract}

\endgroup


\end{frontmatter}

\section{Introduction}
\begin{figure*}[t!]
\centering

\resizebox{0.90\textwidth}{!}{%
\begin{tikzpicture}[
    node distance=0.55cm and 0.75cm,
    >=Stealth
]


\node[y27api] (api)
{\textbf{Unified Ultralytics YOLO27 API}};


\node[
    y27cnnhead,
    below left=1.00cm and 1.50cm of api
] (ns)
{\textbf{YOLO27n/s}\\
Streamlined CNN};

\node[
    y27queryhead,
    below right=1.00cm and 1.50cm of api
] (ml)
{\textbf{YOLO27m/l}\\
Query-Based Detection};


\node[
    above=0.18cm of ns,
    text=YoloOrange,
    font=\scriptsize\bfseries
]
{\hspace{0.2 cm}OEFFICIENCY- \hspace{0.3 cm}ORIENTED PATH};

\node[
    above=0.18cm of ml,
    text=YoloBlue,
    font=\scriptsize\bfseries
]
{\hspace{1.7 cm}ACCURACY / CONTEXT-ORIENTED PATH};


\node[
    y27cnn,
    below=0.58cm of ns
] (high)
{Strengthened\\
High-Resolution Stage};

\node[
    y27cnn,
    below=0.52cm of high
] (fpn1)
{FPN/PAN\\
Feature Fusion};

\node[
    y27cnn,
    minimum width=2.0cm,
    below left=0.60cm and 0.27cm of fpn1
] (pf)
{\textbf{Fine Scale}\\
$P_f$};

\node[
    y27cnn,
    minimum width=2.0cm,
    below right=0.60cm and 0.27cm of fpn1
] (pc)
{\textbf{Coarse Scale}\\
$P_c$};

\node[
    y27outputcnn,
    below=2.3cm of fpn1
] (dense)
{One-to-Many + NMS\\
or One-to-One NMS-Free};

\node[
    y27box,
    minimum width=2.45cm,
    fill=white,
    draw=YoloRed,
    text=YoloRed,
    font=\scriptsize\bfseries,
    left=2.2cm of dense
] (fg)
{Training-Only\\
Foreground Alignment};


\node[
    y27query,
    below left=0.60cm and 0.18cm of ml
] (mback)
{\textbf{YOLO27m}\\
CNN Backbone};

\node[
    y27query,
    below right=0.60cm and 0.18cm of ml
] (lback)
{\textbf{YOLO27l}\\
UltraViT Backbone};

\node[
    y27query,
    below=1.10cm of ml
] (fpn2)
{Multi-Scale FPN/PAN\\
Features};

\node[
    y27query,
    below=0.50cm of fpn2
] (queries)
{\textbf{Learned Object Queries}};

\node[
    y27query,
    below=0.50cm of queries
] (decoder)
{\textbf{Transformer Decoder}};

\node[
    y27outputquery,
    below=0.50cm of decoder
] (directout)
{Class + Box Predictions\\
\textbf{Native NMS-Free}};


\draw[y27arrow]
(api.south) -- ++(0,-0.28)
-| (ns.north);

\draw[y27arrow]
(api.south) -- ++(0,-0.28)
-| (ml.north);


\draw[y27cnnarrow]
(ns.south) -- (high.north);

\draw[y27cnnarrow]
(high.south) -- (fpn1.north);

\draw[y27cnnarrow]
(fpn1.south) -- ++(0,-0.18)
-| (pf.north);

\draw[y27cnnarrow]
(fpn1.south) -- ++(0,-0.18)
-| (pc.north);

\draw[y27cnnarrow]
(pf.south) |- (dense.north);

\draw[y27cnnarrow]
(pc.south) |- (dense.north);

\draw[y27trainarrow]
(fg.east) -- (dense.west);


\draw[y27queryarrow]
(ml.south) -- ++(0,-0.18)
-| (mback.north);

\draw[y27queryarrow]
(ml.south) -- ++(0,-0.18)
-| (lback.north);

\draw[y27queryarrow]
(mback.south) |- (fpn2.west);

\draw[y27queryarrow]
(lback.south) |- (fpn2.east);

\draw[y27queryarrow]
(fpn2.south) -- (queries.north);

\draw[y27queryarrow]
(queries.south) -- (decoder.north);

\draw[y27queryarrow]
(decoder.south) -- (directout.north);


\begin{scope}[on background layer]

\node[
    fit=(ns)(high)(fpn1)(pf)(pc)(dense)(fg),
    rounded corners=5pt,
    draw=YoloOrange,
    line width=1.0pt,
    fill=YoloAmber,
    fill opacity=0.12,
    inner sep=0.23cm
] {};

\node[
    fit=(ml)(mback)(lback)(fpn2)(queries)(decoder)(directout),
    rounded corners=5pt,
    draw=YoloBlue,
    line width=1.0pt,
    fill=YoloCyan,
    fill opacity=0.14,
    inner sep=0.23cm
] {};

\end{scope}


\node[
    below=0.78cm of api,
    fill=white,
    text=YoloNavy,
    font=\scriptsize\bfseries,
    inner xsep=6pt,
    inner ysep=6.5pt
]
{SCALE-DEPENDENT ARCHITECTURAL DIVERGENCE};

\end{tikzpicture}%
}


\caption{Scale-adaptive dual architecture of YOLO27.
Compact YOLO27n/s detectors employ a streamlined CNN-based pathway with
strengthened high-resolution representations, FPN/PAN feature fusion,
fine- and coarse-scale prediction, foreground-alignment supervision, and
either one-to-many NMS-based or one-to-one NMS-free detection. In contrast,
YOLO27m/l employ query-based detection with learned object queries and
transformer decoding. YOLO27m retains a CNN backbone, whereas YOLO27l
incorporates an UltraViT backbone. Dashed connections denote
training-specific supervision rather than the principal inference pathway.}

\label{fig:y27dual}

\end{figure*}
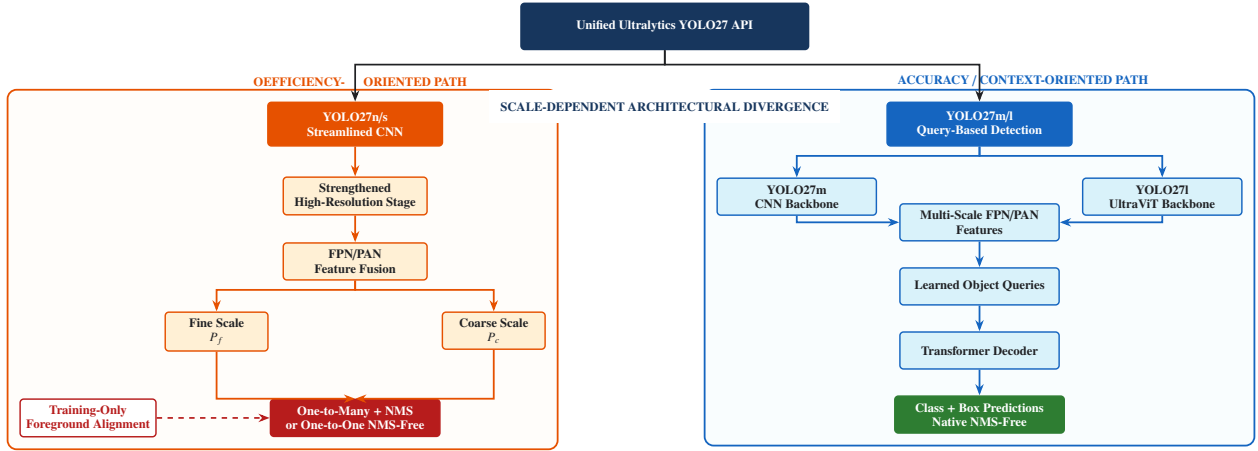

\begingroup
\setlength{\leftskip}{0.3cm}
\setlength{\rightskip}{-0.3cm}
Real-time object detection and instance segmentation have become fundamental components of modern computer vision systems, providing the perceptual foundation for robotics, autonomous systems, environmental monitoring, industrial inspection, and precision agriculture \cite{guo2025visual, hafiz2020survey}. Object detection determines the location and semantic identity of targets within an image, whereas instance segmentation extends this capability by delineating the pixel-level spatial extent of individual objects. The latter is particularly important in applications where subsequent decisions depend not only on recognizing an object but also on understanding its geometry, boundaries, orientation, or interaction with surrounding structures. Among single-stage detectors, YOLO has advanced real-time vision by balancing accuracy and computational efficiency since its original formulation \cite{redmon2016you,sapkota2025yolo,sapkota2025ultralytics}. Figure~\ref{fig:mapflops_extended} compares detection accuracy against computational complexity across successive YOLO generations, showing how performance and efficiency have evolved from earlier models to the latest YOLO27 architecture (Refer to Figure \ref{fig:y27dual}).
\endgroup

\begin{figure}[H]
\centering

\resizebox{0.48\textwidth}{!}{%
\begin{tikzpicture}[
    x=0.047cm,
    y=0.42cm,
    >=Stealth
]


\definecolor{y27color}{RGB}{190,35,35}
\definecolor{y26color}{RGB}{20,48,170}
\definecolor{y11color}{RGB}{242,155,82}
\definecolor{y10color}{RGB}{105,175,105}
\definecolor{y9color}{RGB}{220,120,120}
\definecolor{y8color}{RGB}{170,145,205}
\definecolor{y7color}{RGB}{170,145,130}


\draw[black,thick]
    (0,0) rectangle (280,26);


\foreach \x in {0,40,80,120,160,200,240,280}
{
    \draw[gray!25,thin]
        (\x,0) -- (\x,26);

    \draw[black,thick]
        (\x,0) -- (\x,-0.35);

    \node[
        anchor=north,
        font=\scriptsize
    ] at (\x,-0.45) {\x};
}


\foreach \yy/\lab in {
1.5/37.5,
4/40,
6.5/42.5,
9/45,
11.5/47.5,
14/50,
16.5/52.5,
19/55,
21.5/57.5,
24/60
}
{
    \draw[gray!25,thin]
        (0,\yy) -- (280,\yy);

    \draw[black,thick]
        (0,\yy) -- (-1.6,\yy);

    \node[
        anchor=east,
        font=\scriptsize
    ] at (-2.0,\yy) {\lab};
}


\node[
    anchor=north,
    font=\small
] at (140,-2.2)
{\textbf{FLOPs (B) at 640-pixel input}};

\node[
    rotate=90,
    anchor=south,
    font=\small
] at (-12,13)
{\textbf{COCO mAP$_{50:95}^{val}$}};


\draw[
    y27color,
    line width=1.8pt
]
(7.2,6.3) --
(28.2,13.6) --
(65.0,19.8) --
(165.3,24.4);

\filldraw[fill=y27color,draw=white,line width=0.7pt]
    (7.2,6.3) circle (3.5pt);
\filldraw[fill=y27color,draw=white,line width=0.7pt]
    (28.2,13.6) circle (3.5pt);
\filldraw[fill=y27color,draw=white,line width=0.7pt]
    (65.0,19.8) circle (3.5pt);
\filldraw[fill=y27color,draw=white,line width=0.7pt]
    (165.3,24.4) circle (3.5pt);

\node[text=y27color,font=\small\bfseries,anchor=south east]
    at (6.2,6.7) {27n};
\node[text=y27color,font=\small\bfseries,anchor=south]
    at (28.2,14.1) {27s};
\node[text=y27color,font=\small\bfseries,anchor=south]
    at (65.0,20.3) {27m};
\node[text=y27color,font=\small\bfseries,anchor=south]
    at (165.3,24.9) {27l};


\draw[y26color,line width=1.55pt]
(5.4,4.9) --
(20.7,12.6) --
(68.2,17.1) --
(86.4,19.0) --
(193.9,21.5);

\filldraw[fill=y26color,draw=white,line width=0.7pt]
    (5.4,4.9) circle (3.2pt);
\filldraw[fill=y26color,draw=white,line width=0.7pt]
    (20.7,12.6) circle (3.2pt);
\filldraw[fill=y26color,draw=white,line width=0.7pt]
    (68.2,17.1) circle (3.2pt);
\filldraw[fill=y26color,draw=white,line width=0.7pt]
    (86.4,19.0) circle (3.2pt);
\filldraw[fill=y26color,draw=white,line width=0.7pt]
    (193.9,21.5) circle (3.2pt);

\node[text=y26color,font=\scriptsize\bfseries,anchor=north west]
    at (6.2,4.5) {26n};
\node[text=y26color,font=\scriptsize\bfseries,anchor=south]
    at (20.7,13.0) {26s};
\node[text=y26color,font=\scriptsize\bfseries,anchor=south east]
    at (67.0,17.5) {26m};
\node[text=y26color,font=\scriptsize\bfseries,anchor=south]
    at (86.4,19.4) {26l};
\node[text=y26color,font=\scriptsize\bfseries,anchor=south]
    at (193.9,21.9) {26x};


\draw[y11color,line width=1.05pt]
(6.5,3.5) --
(21.5,11.0) --
(68.0,15.5) --
(86.9,17.4) --
(194.9,18.7);

\fill[y11color] (6.5,3.5) circle (2.3pt);
\fill[y11color] (21.5,11.0) circle (2.3pt);
\fill[y11color] (68.0,15.5) circle (2.3pt);
\fill[y11color] (86.9,17.4) circle (2.3pt);
\fill[y11color] (194.9,18.7) circle (2.3pt);

\node[text=y11color,font=\scriptsize,anchor=north]
    at (6.5,3.1) {11n};
\node[text=y11color,font=\scriptsize,anchor=south west]
    at (22.2,11.3) {11s};
\node[text=y11color,font=\scriptsize,anchor=north west]
    at (69.0,15.2) {11m};
\node[text=y11color,font=\scriptsize,anchor=south]
    at (86.9,17.8) {11l};
\node[text=y11color,font=\scriptsize,anchor=south]
    at (194.9,19.1) {11x};


\draw[y10color,line width=1.05pt]
(6.7,2.5) --
(21.6,10.3) --
(59.1,15.1) --
(92.0,16.5) --
(120.3,17.2) --
(160.4,18.4);

\fill[y10color] (6.7,2.5) circle (2.3pt);
\fill[y10color] (21.6,10.3) circle (2.3pt);
\fill[y10color] (59.1,15.1) circle (2.3pt);
\fill[y10color] (92.0,16.5) circle (2.3pt);
\fill[y10color] (120.3,17.2) circle (2.3pt);
\fill[y10color] (160.4,18.4) circle (2.3pt);

\node[text=y10color,font=\scriptsize,anchor=east]
    at (5.8,2.5) {10n};
\node[text=y10color,font=\scriptsize,anchor=north west]
    at (22.3,10.0) {10s};
\node[text=y10color,font=\scriptsize,anchor=south east]
    at (58.4,15.5) {10m};
\node[text=y10color,font=\scriptsize,anchor=south]
    at (92.0,16.9) {10b};
\node[text=y10color,font=\scriptsize,anchor=south]
    at (120.3,17.6) {10l};
\node[text=y10color,font=\scriptsize,anchor=south]
    at (160.4,18.8) {10x};


\draw[y9color,line width=1.05pt]
(7.7,2.3) --
(26.7,10.8) --
(76.8,15.4) --
(102.8,17.0) --
(192.5,19.6);

\fill[y9color] (7.7,2.3) circle (2.3pt);
\fill[y9color] (26.7,10.8) circle (2.3pt);
\fill[y9color] (76.8,15.4) circle (2.3pt);
\fill[y9color] (102.8,17.0) circle (2.3pt);
\fill[y9color] (192.5,19.6) circle (2.3pt);

\node[text=y9color,font=\scriptsize,anchor=north west]
    at (8.4,2.0) {9t};
\node[text=y9color,font=\scriptsize,anchor=south]
    at (26.7,11.2) {9s};
\node[text=y9color,font=\scriptsize,anchor=south west]
    at (77.5,15.7) {9m};
\node[text=y9color,font=\scriptsize,anchor=south]
    at (102.8,17.4) {9c};
\node[text=y9color,font=\scriptsize,anchor=south]
    at (192.5,20.0) {9e};


\draw[y8color,line width=1.05pt]
(8.7,1.3) --
(28.6,8.9) --
(78.9,14.2) --
(165.1,16.9) --
(257.8,17.9);

\fill[y8color] (8.7,1.3) circle (2.3pt);
\fill[y8color] (28.6,8.9) circle (2.3pt);
\fill[y8color] (78.9,14.2) circle (2.3pt);
\fill[y8color] (165.1,16.9) circle (2.3pt);
\fill[y8color] (257.8,17.9) circle (2.3pt);

\node[text=y8color,font=\scriptsize,anchor=north]
    at (8.7,0.9) {8n};
\node[text=y8color,font=\scriptsize,anchor=south]
    at (28.6,9.3) {8s};
\node[text=y8color,font=\scriptsize,anchor=north west]
    at (79.7,13.9) {8m};
\node[text=y8color,font=\scriptsize,anchor=north]
    at (165.1,16.5) {8l};
\node[text=y8color,font=\scriptsize,anchor=south]
    at (257.8,18.3) {8x};


\draw[y7color,line width=1.05pt]
(104.7,15.4) --
(189.9,17.1);

\fill[y7color] (104.7,15.4) circle (2.3pt);
\fill[y7color] (189.9,17.1) circle (2.3pt);

\node[text=y7color,font=\scriptsize,anchor=north]
    at (104.7,15.0) {7l};
\node[text=y7color,font=\scriptsize,anchor=north]
    at (189.9,16.7) {7x};


\draw[
    y27color,
    dashed,
    thick,
    -{Stealth[length=2mm]}
]
(75,22.5)
.. controls (100,24.0) and (125,24.8) ..
(157,24.4);

\node[
    text=y27color,
    font=\scriptsize\bfseries,
    align=center,
    fill=white,
    inner sep=2pt
]
at (112,24.5)
{YOLO27 query-based\\m/l regime};


\draw[
    gray!60,
    fill=white,
    rounded corners=2pt
]
(216,3.0) rectangle (276,13.2);

\draw[y27color,line width=1.8pt]
    (220,12.1) -- (230,12.1);
\fill[y27color] (225,12.1) circle (2.6pt);
\node[anchor=west,font=\scriptsize\bfseries]
    at (233,12.1) {YOLO27};

\draw[y26color,line width=1.6pt]
    (220,10.8) -- (230,10.8);
\fill[y26color] (225,10.8) circle (2.4pt);
\node[anchor=west,font=\scriptsize\bfseries]
    at (233,10.8) {YOLO26};

\draw[y11color,line width=1.05pt]
    (220,9.5) -- (230,9.5);
\fill[y11color] (225,9.5) circle (2.1pt);
\node[anchor=west,font=\scriptsize]
    at (233,9.5) {YOLO11};

\draw[y10color,line width=1.05pt]
    (220,8.2) -- (230,8.2);
\fill[y10color] (225,8.2) circle (2.1pt);
\node[anchor=west,font=\scriptsize]
    at (233,8.2) {YOLOv10};

\draw[y9color,line width=1.05pt]
    (220,6.9) -- (230,6.9);
\fill[y9color] (225,6.9) circle (2.1pt);
\node[anchor=west,font=\scriptsize]
    at (233,6.9) {YOLOv9};

\draw[y8color,line width=1.05pt]
    (220,5.6) -- (230,5.6);
\fill[y8color] (225,5.6) circle (2.1pt);
\node[anchor=west,font=\scriptsize]
    at (233,5.6) {YOLOv8};

\draw[y7color,line width=1.05pt]
    (220,4.3) -- (230,4.3);
\fill[y7color] (225,4.3) circle (2.1pt);
\node[anchor=west,font=\scriptsize]
    at (233,4.3) {YOLOv7};

\end{tikzpicture}%
}

\caption{%
Accuracy--computational-complexity comparison of YOLO27, YOLO26, YOLO11,
YOLOv10, YOLOv9, YOLOv8, and YOLOv7 detection models on COCO. The vertical
axis reports mAP$_{50:95}^{val}$, while the horizontal axis represents
floating-point operations at approximately 640-pixel input resolution.
YOLO27 and YOLO26 are emphasized to illustrate the transition from
deployment-oriented CNN/end-to-end detection toward scale-adaptive
architectures, including query-based YOLO27m/l detection. FLOPs provide a
hardware-independent approximation of arithmetic complexity but do not
capture memory movement, operator fusion, attention-kernel efficiency, or
accelerator utilization. YOLO27 results are preliminary.
}
\label{fig:mapflops_extended}

\end{figure}
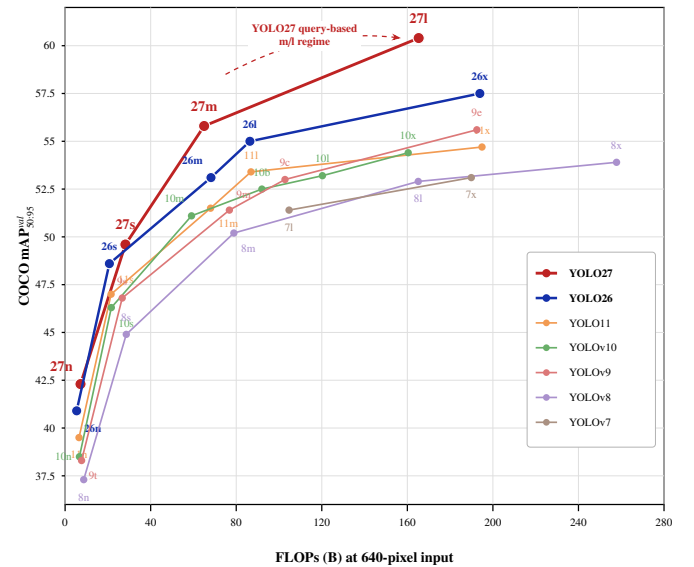



The YOLO family has undergone substantial architectural evolution since its early generations. YOLOv3 \cite{redmon2018yolov3} strengthened multi-scale prediction and became an influential reference architecture for real-time object detection, while YOLOv5 (\href{https://docs.ultralytics.com/models/yolov5}{Source Link}) contributed to the practical accessibility and deployment of YOLO-based pipelines. Subsequent architectures, including YOLOv7 \cite{wang2023yolov7}, YOLOv8 (\href{https://docs.ultralytics.com/models/yolov8}{Source Link}), and YOLO11 (\href{https://docs.ultralytics.com/models/yolo11}{Source Link}), progressively incorporated improvements in feature extraction, feature aggregation, prediction heads, optimization, and computational efficiency \cite{sapkota2025yolo}. Collectively, these developments reflect a broader transition from conventional single-stage detectors toward increasingly efficient multi-task perception architectures capable of supporting object detection, instance segmentation, pose estimation, and other visual-recognition tasks while remaining suitable for real-time and edge-oriented deployment \cite{ali2024yolo}.

Across successive generations, improvements in backbone design, multi-scale feature aggregation, anchor-free detection, decoupled prediction mechanisms, and task-specific segmentation components have substantially increased the representational capacity of YOLO models \cite{palaniappan2025yolo, casas2023assessing}. These architectural changes are particularly relevant to instance segmentation because successful prediction requires simultaneous recognition of object identity, accurate spatial localization, and preservation of sufficiently detailed representations for pixel-level mask reconstruction. However, improvements between YOLO generations are not necessarily uniform across object sizes or application domains. A model that performs strongly on conventional benchmark objects may not provide equivalent improvements when the targets are extremely small, elongated, partially occluded, densely clustered, or weakly contrasted against their backgrounds. Consequently, evaluating successive YOLO generations under the same application-specific dataset and training conditions is important for distinguishing improvements attributable to architectural evolution from those associated with model capacity, input representation, or training strategy.

YOLO26 (\href{https://docs.ultralytics.com/models/yolo26/}{Source Link}) represents a recent stage in this architectural evolution and places particular emphasis on deployment-oriented efficiency and streamlined end-to-end prediction \cite{sapkota2025yolo26}. Its design prioritizes simplified inference, reduced post-processing dependence, and efficient transfer from model development to practical deployment. These characteristics are relevant to robotic and autonomous systems in which perception latency, computational resource requirements, and predictable inference behavior can be as important as absolute accuracy.

Compared with earlier high-performing generations such as YOLO11
(\href{https://docs.ultralytics.com/models/yolo11/}{Source Link}),
YOLOv12 \cite{tian2025yolov12}, and YOLOv13 \cite{lei2025yolov13},
which introduced advances in feature extraction, multi-scale representation,
and computational efficiency \cite{sapkota2025yolo}, YOLO26 further
emphasized end-to-end efficiency, simplified prediction mechanisms, and
reduced post-processing overhead
\cite{sapkota2025ultralytics,sapkota2025yolo26}.

More recently, YOLO27 represents the latest generation of the Ultralytics
YOLO family, extending its architectural evolution toward efficient real-time
object detection and segmentation \cite{Sapkota2026YOLO27}. Rather than
treating model scale solely as a change in network depth or width, YOLO27
introduces a more fundamental distinction at the \textit{model-family}
level. Whereas YOLO26 follows a common detection design across its
n/s/m/l/x capacities, YOLO27 currently organizes its n/s/m/l models into
two architectural regimes
(\href{https://docs.ultralytics.com/models/yolo27}{Ultralytics YOLO27}).
Accordingly, meaningful assessment of this architectural transition requires
comparison with established predecessors under identical datasets, target
classes, training conditions, and evaluation protocols rather than
evaluation of the newest generation in isolation.

The scale-dependent YOLO27 organization can be represented compactly as

\begin{equation}
\mathcal{A}_{27}(s)=
\begin{cases}
\mathcal{A}^{\mathrm{dual}}_{\mathrm{CNN}},
& s\in\{n,s\},\\[3pt]
\mathcal{A}_{\mathrm{query}},
& s\in\{m,l\},
\end{cases}
\label{eq:y27scale}
\end{equation}

where $\mathcal{A}^{\mathrm{dual}}_{\mathrm{CNN}}$ denotes the streamlined
CNN-based detection regime and $\mathcal{A}_{\mathrm{query}}$ denotes the
query-based detection regime. Consequently, YOLO27m/l should not be
interpreted simply as deeper or wider extensions of YOLO27n/s; instead,
model scaling crosses an architectural boundary that changes the underlying
detection mechanism. Figure~\ref{fig:y27dual} illustrates this scale-adaptive
organization and the principal information flow within the two YOLO27
architectural paths.

This need for cross-generation comparison is particularly important for small-object perception. Despite substantial architectural advances, reliable detection and segmentation of extremely small targets remain difficult because repeated resizing and spatial downsampling can progressively reduce the pixel-level representation of fine visual structures \cite{tao2024enhanced, liu2025yolo, zhao2024small}. When an object occupies only a limited region of the original image, resizing the complete scene to a fixed neural-network input resolution can reduce the target to relatively few pixels. Subsequent feature extraction further decreases spatial resolution as information propagates through the network. Fine edges, narrow structures, subtle textures, and weak object--background differences can therefore become progressively less distinguishable. For instance segmentation, the problem is particularly consequential because insufficient spatial representation affects not only whether an object is recognized but also whether its boundary can be reconstructed accurately.

A wide range of approaches has consequently been developed to improve small-object perception. Architectural strategies such as Feature Pyramid Networks (FPN) and Path Aggregation Networks (PANet) enhance information exchange among multiple feature-map resolutions, enabling high-resolution spatial information to be combined with deeper semantic representations \cite{guo2023multi, zheng2022gated, ding2026improvement}. Training-oriented approaches such as multi-scale training expose models to objects represented at different effective scales \cite{zhu2025cross, wang2025method}, while data augmentation increases variability in object appearance, location, scale, and imaging conditions \cite{hammami2020cycle, chen2024nhd, lin2025mixed}. Test-time augmentation (TTA) has similarly been investigated as a means of improving robustness by aggregating predictions from transformed representations of the same input image \cite{zhang2021vit, garta2024improved}. Attention mechanisms \cite{chung2024yolo, chen2022object, lan2025crack} and transformer-inspired modules have further expanded the ability of detection architectures to incorporate contextual relationships and long-range information, with potential benefits for small, partially occluded, or visually ambiguous objects \cite{rashid2026enhancing, rani2025extreme}. Nevertheless, increasingly elaborate architectural modifications can introduce additional parameters, computational operations, memory requirements, or training complexity, which may limit their suitability for resource-constrained or real-time systems.

\textcolor{black}{Small-object detection and fine-grained instance segmentation remain strongly dependent on the spatial information preserved throughout the perception pipeline. Limited feature-map resolution \cite{li2025ses, zeng2025multiscale, jing2024effective} and progressive spatial downsampling \cite{liu2025yolo, zhang2025adaptive} can substantially degrade representations when targets occupy only a small fraction of an image, particularly under occlusion, low foreground--background contrast, irregular geometry, and dense clustering. Consequently, increasing network capacity alone may provide limited benefit when discriminative spatial information has already been lost, motivating approaches that improve the effective representation of small targets.}

\textcolor{black}{Slicing-Aided Hyper Inference (SAHI) illustrates this broader principle by processing overlapping image regions so that small targets occupy a greater proportion of the model input \cite{akyon2022slicing}. Its reported effectiveness with Faster R-CNN and YOLO-based detectors \cite{mazen2025small, chen2022improved, zhorif2024implementation, zhang2024detection} further demonstrates the importance of effective target resolution. Similar resolution-oriented benefits have been reported for small vehicles in remote sensing \cite{vu2025object, liu2023traffic, zhang2023adaptive}, micro-lesions in medical imaging \cite{mihajlovic2025medical, pereira2023improved}, agricultural pests and diseases \cite{fotouhi2024persistent, ye2024improved, zeng2023feasibility}, distant pedestrians and traffic signs \cite{gia2024enhancing, simeonov2025evaluating}, marine vessels \cite{caricchio2024yolov8, de2025small}, UAV-based wildlife monitoring \cite{dat2025wildlive, chen2026cams}, and small industrial defects \cite{cabral2026automated, liu2026precision, zhang2026vision}. Related challenges also occur in wildfire monitoring \cite{tran2026kinematic, del2025smoke}, illegal-logging detection \cite{ye2023nonsalient, simoes2025human}, and large-area surveillance \cite{iqra2024small, alshoail2025dual}.}

\textcolor{black}{Motivated by this general principle rather than by slicing-based inference itself, the present study investigates spatial-resolution optimization directly during model training. Specifically, conventional $640\times640$-pixel training is compared with a small-object-focused $960\times960$-pixel configuration across the n, s, m, l, and x variants of YOLOv8, YOLO11, and YOLO26. This controlled design enables the effects of YOLO generation, model capacity, and training resolution to be disentangled on the same fine-grained orchard dataset, while determining whether higher spatial resolution provides greater benefit than simply increasing network capacity.}

\subsection{Small-Object Detection and Segmentation in Complex Natural Scenes}

\textcolor{black}{Small-object detection is widely recognized as one of the most challenging problems in visual perception \cite{mirzaei2023small, cheng2023towards}. Small targets contain intrinsically limited visual information, and this information can be further degraded by image resizing, network downsampling, and hierarchical feature extraction \cite{talebi2021learning, danon2021image}. Deep convolutional networks employing fixed-stride downsampling are particularly susceptible to this problem because an object represented by relatively few pixels at the input can become extremely compact at deeper feature levels \cite{iqra2024small}. As spatial support decreases, discriminative texture, contour, and shape information may be lost or become indistinguishable from surrounding features.}

\textcolor{black}{Natural outdoor environments introduce additional difficulties beyond object size. Illumination changes, cluttered backgrounds, partial occlusion, weak contrast, motion, variable viewpoints, and interactions among neighboring objects can all influence the visual appearance of small targets \cite{iqra2024small, batlle2000review, muzammul2025comprehensive}. Small targets may also occur in dense groups, causing object boundaries to overlap or become partially hidden. Under these conditions, even high-performing detectors can exhibit reduced recall \cite{tang2025learning, he2024recalling}, unstable localization \cite{ye2022dense, mirzaei2023small}, and fragmented or incomplete masks \cite{iqra2024small}. A model can therefore achieve strong aggregate performance while still performing inadequately on the smallest or most operationally important classes, emphasizing the importance of class-specific evaluation.}

\textcolor{black}{Instance segmentation places an even stronger requirement on spatial representation than bounding-box detection. Whereas detection requires sufficiently accurate localization of an object region, segmentation requires pixel-level delineation of individual object boundaries \cite{zhang2021refinemask, hafiz2020survey}. Thin, elongated, irregular, or partially visible targets are especially difficult because minor localization errors can represent a substantial fraction of the complete object. Predicted masks may consequently exhibit erosion, discontinuities, incomplete structures, or leakage into visually similar background regions \cite{mohammed2024insights, li2025instance}. These challenges arise across many domains, including small manufactured-component detection \cite{yang2019real, block2022image}, biological structure recognition \cite{huang2024brstd, iqra2024small}, material-defect inspection \cite{huang2022small, gong2022transfer}, and early-stage crop perception \cite{li2024object, junos2021optimized}. Small-object segmentation therefore requires methods capable of preserving fine spatial information \cite{muzammul2025comprehensive} while remaining computationally practical for the intended application.}

\textcolor{black}{The relationship between object scale and model scale is also important. Larger neural networks generally provide greater representational capacity, but increased parameter count does not necessarily recover information that has been removed through input resizing or spatial downsampling. Conversely, increasing input resolution may preserve additional spatial information but also increases training and inference computation. Determining the appropriate balance between these factors requires controlled experiments in which model generation, network capacity, and training configuration are varied systematically rather than independently across unrelated datasets. This consideration forms a central motivation for evaluating five model capacities within each of three YOLO generations under both conventional and small-object-focused training configurations.}

\subsection{Green-on-Green Tree Fruit Canopy Perception as a Representative Challenge}

\begin{figure*}[ht!]
     \centering
     \includegraphics[width=0.9\linewidth]{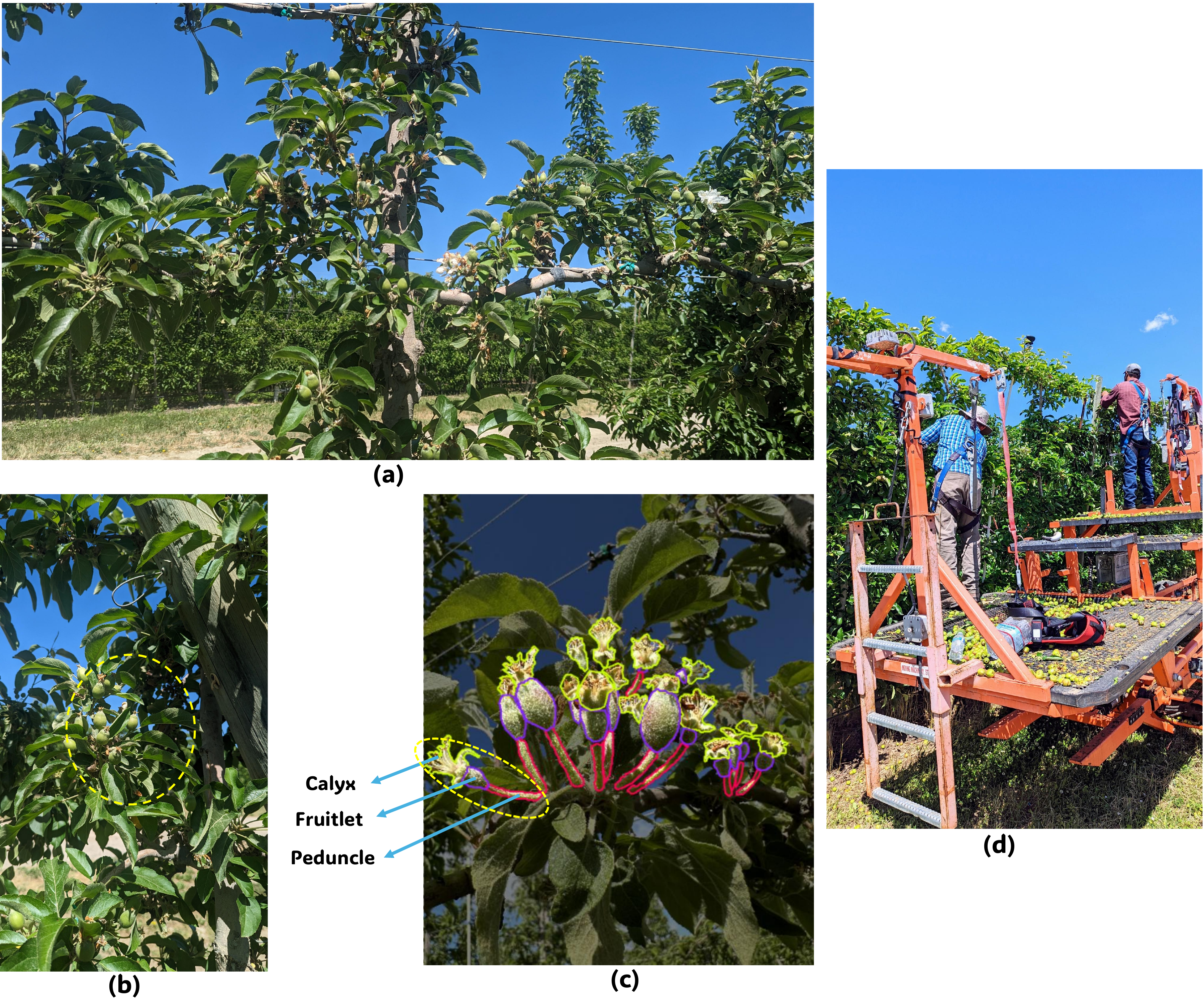}
     \caption{\textbf{Representative examples of complex green fruit orchard scenes used in this study.} (a) Full-view image of a commercial apple orchard during the early green fruit (fruitlet) development stage, where immature fruitlets are visually camouflaged within dense green canopy foliage. (b) Enlarged canopy region highlighting a dense cluster of approximately ten green fruitlets (yellow dashed circle), illustrating high object proximity, clustering, and background clutter. (c) Fine-grained anatomical annotation of clustered fruitlets into calyx, fruit main body, and peduncle, each representing a small and low-contrast target. (d) Workers performing manual fruitlet thinning in the field, a labor-intensive and costly but essential tree-fruit production operation increasingly constrained by worker shortages.}
     \label{fig:greenfruits}
\end{figure*}

\textcolor{black}{To ground the investigation in a realistic and operationally demanding environment, this study focuses on visual perception during the early green-fruit, or fruitlet, development stage in commercial apple orchards. At this stage, immature fruits and their anatomical structures must be recognized against a predominantly green canopy. The resulting green-on-green appearance creates a particularly difficult combination of low color contrast, dense vegetation, occlusion, scale variation, irregular illumination, and object clustering. Figure~\ref{fig:greenfruits} presents representative scenes from the dataset used in this study and illustrates the progression from complete orchard imagery to fine-grained anatomical annotation and the intended field-level application.}

\textcolor{black}{As shown in Fig.~\ref{fig:greenfruits}a, immature apple fruitlets are distributed throughout a visually heterogeneous canopy containing leaves, branches, shoots, stems, and variable illumination. Fruitlets and foliage exhibit similar green coloration and can share comparable reflectance and texture characteristics, substantially reducing foreground--background separability. The difficulty increases when direct sunlight, shadows, depth variation, and leaf orientation produce large within-class appearance changes. Under these conditions, simple color-based segmentation or thresholding becomes unreliable, and a learning-based model must instead exploit combinations of shape, texture, boundary, and contextual information.}

\textcolor{black}{Figure~\ref{fig:greenfruits}b illustrates an additional challenge arising from dense fruit clustering. The highlighted region contains approximately ten immature fruitlets in close spatial proximity. Fruitlets may overlap one another, be partially covered by foliage, or exhibit boundaries that visually merge with neighboring instances. Dense clustering therefore challenges both object localization and instance separation. The problem is particularly important for instance segmentation because the model must distinguish adjacent objects rather than merely identify that a cluster contains fruit. Similar conditions have been reported as major difficulties for green-fruit perception in orchard environments \cite{hussain2023green, gonzalez2023evaluation, sapkota2024immature}.}

\textcolor{black}{The perception problem considered here extends beyond whole-fruit recognition. As illustrated in Fig.~\ref{fig:greenfruits}c, each fruitlet is decomposed into three anatomically meaningful classes: the calyx, fruitlet body, and peduncle. These classes differ substantially in apparent size, geometry, visibility, and functional significance. The fruitlet body generally provides the largest target region, whereas the calyx occupies a considerably smaller and geometrically irregular region. The peduncle is frequently the most spatially constrained target because of its narrow, elongated geometry. Consequently, a single image can contain target instances spanning markedly different pixel areas, requiring the model to preserve both broader semantic context and fine spatial information.}

\textcolor{black}{The peduncle is particularly challenging because its appearance can resemble nearby leaf stems, shoots, and thin branches. Its limited width means that only a small number of pixels may represent the structure after image resizing, and partial occlusion can further reduce the visible region. Small errors in localization can therefore result in substantial mask degradation or complete omission of the structure. The calyx presents a related but geometrically different problem: its compact radial morphology, small spatial footprint, variable orientation, and partial occlusion can make its boundaries difficult to distinguish from the fruit surface or nearby canopy elements. These characteristics make the three-class formulation a more stringent evaluation of small-object perception than conventional whole-fruit detection.}

\textcolor{black}{Existing orchard-vision studies have extensively investigated whole-fruit detection, counting, and segmentation using earlier YOLO architectures and two-stage detection frameworks \cite{lyu2022green,gremes2023system, kang2025integrating, lu2023mask, qiang2024detecting, wang2024greenfruitdetector, SAPKOTA2026100125}. However, fine-grained multi-class instance segmentation of early-stage fruitlet anatomy---particularly the simultaneous delineation of the calyx, fruitlet body, and peduncle---has received substantially less attention. This distinction is important for robotic manipulation. Whole-fruit localization may be sufficient for counting or crop-load estimation, but selective physical interaction can require knowledge of attachment structures and local fruit geometry. The present benchmark therefore moves beyond whole-fruit perception toward anatomical-level scene understanding.}

\textcolor{black}{The inclusion of YOLO26 further introduces a timely cross-generation question. Although the recent architecture is designed to improve efficiency and deployment behavior, its performance should not be inferred solely from results reported for other datasets or YOLO generations. In particular, the extent to which its architectural advances translate to fine-grained green-on-green anatomical segmentation remains unclear. Direct comparison with YOLOv8 and YOLO11 under the same dataset, target classes, model-capacity levels, and training strategies is therefore necessary to establish whether newer architecture alone produces consistent benefits for small targets. The present study consequently uses YOLOv8 and YOLO11 as earlier-generation reference families rather than reporting YOLO26 as an isolated baseline.}

\textcolor{black}{The agricultural importance of this perception problem is illustrated in Fig.~\ref{fig:greenfruits}d. During early fruit development, growers perform green-fruit thinning to regulate crop load and reduce excessive competition among developing fruits. Maintaining an appropriate number of fruits per cluster contributes to desirable fruit size, quality attributes such as color and sugar accumulation, and overall tree productivity. Manual thinning requires workers to inspect dense canopies and selectively remove excess fruitlets, making the operation labor intensive and dependent on seasonal workforce availability. Accurate machine perception of fruitlets and their attachment anatomy can therefore contribute to automated or semi-autonomous thinning, selective manipulation, crop-load assessment, and other precision orchard operations \cite{miranda2024open, scalisi2024detecting, mirbod2023tree, ji2025green}.}

The visual conditions represented in Fig.~\ref{fig:greenfruits} also provide a useful model problem for small-object perception beyond agriculture. A peduncle embedded within green foliage presents a conceptually similar challenge to a micro-lesion embedded in heterogeneous tissue, a small vehicle embedded within a large aerial scene, or a fine surface defect embedded within repetitive industrial texture. In each case, the target occupies a small proportion of the complete image and may exhibit only subtle differences from its surroundings. Dense clustering and partial occlusion further increase the importance of retaining local spatial information while simultaneously exploiting broader contextual features.

\textcolor{black}{For this reason, training resolution represents a particularly relevant experimental variable. When a high-resolution orchard image is resized to $640\times640$ pixels, small anatomical structures necessarily receive fewer pixels than they would at a $960\times960$ input resolution. Increasing input resolution does not guarantee improved performance, because the additional spatial information is accompanied by increased computational demand and may interact differently with different model architectures and capacities. Nevertheless, the underlying hypothesis is experimentally testable: if loss of small-object spatial information is a major limitation, a small-object-focused higher-resolution training configuration should provide measurable benefits, particularly for the calyx and peduncle classes. Evaluating this hypothesis across three YOLO generations and five model scales provides a more rigorous basis than demonstrating it for a single architecture.}

\textcolor{black}{Accordingly, the dataset shown in Fig.~\ref{fig:greenfruits} serves as both an application-specific benchmark and a representative test case for low-contrast, fine-grained small-object perception. Its combination of anatomical decomposition, severe foreground--background similarity, dense clustering, scale variation, occlusion, and operational relevance makes it suitable for examining whether increased training resolution produces consistent improvements across different levels of architectural capacity. Importantly, the revised experimental framework does not attribute the $960\times960$ configuration to SAHI. Instead, SAHI and related slicing approaches provide background evidence motivating the broader importance of effective object resolution, whereas the experiments themselves directly compare conventional and small-object-focused model-training strategies.}

\subsection{Research Gap}
\textcolor{black}{Despite continued progress in YOLO architectures and extensive research on small-object detection, three interconnected gaps remain. First, much of the existing literature evaluates individual YOLO versions or proposes modifications to a particular architecture, limiting the ability to distinguish improvements arising from architectural generation from those produced by model scaling or experimental configuration. The emergence of YOLO26 makes this issue especially relevant because performance improvements claimed for a recent generation require direct comparison with established predecessors under identical data and evaluation conditions.}

\textcolor{black}{Second, the interaction between \emph{model capacity} and \emph{small-object-oriented training resolution} remains insufficiently characterized. Increasing network capacity from nano to small, medium, large, or extra-large variants increases representational and computational capacity, whereas increasing image resolution changes the amount of spatial information presented to the network. These interventions address different aspects of the perception problem. It is therefore important to determine whether high-capacity models inherently compensate for small-object information loss, whether higher-resolution training disproportionately benefits lightweight models, or whether gains persist across the complete capacity spectrum. Such questions cannot be answered through comparisons involving only one or two model scales.}

\textcolor{black}{Third, fine-grained anatomical segmentation in green-on-green orchard environments remains considerably less studied than whole-fruit detection. Existing work has established the feasibility of detecting and segmenting fruits under challenging canopy conditions \cite{lyu2022green,gremes2023system, kang2025integrating, lu2023mask, qiang2024detecting, wang2024greenfruitdetector, SAPKOTA2026100125}, but robotic interaction can require more detailed information about attachment structures and local anatomy. Calyxes and peduncles represent especially demanding targets because their small spatial footprints and irregular or elongated geometries amplify the consequences of spatial downsampling. A benchmark incorporating these structures therefore provides a more demanding test of cross-generation small-object perception.}

\textcolor{black}{To address these gaps, this study presents a comprehensive cross-generation comparison and optimization of Ultralytics YOLOv8, YOLO11, and YOLO26 for fine-grained small-object detection and instance segmentation in complex orchard environments. All five model scales---nano (n), small (s), medium (m), large (l), and extra-large (x)---are considered within each generation, producing 15 architecture--scale combinations. Each combination is investigated under a conventional $640\times640$-pixel training configuration and a small-object-focused $960\times960$-pixel configuration. This design establishes a controlled framework for analyzing the joint influence of architectural generation, network capacity, and input-resolution-oriented optimization while simultaneously considering predictive accuracy, class-specific performance, computational complexity, processing efficiency, training behavior, and deployment feasibility.}

\subsection{Objectives and Contributions}

\textcolor{black}{The primary objective of this study is to comprehensively compare and optimize successive generations of Ultralytics YOLO models for fine-grained small-object detection and instance segmentation in complex orchard environments. Three YOLO generations: YOLOv8, YOLO11, and YOLO26 are investigated across five model scales: nano (n), small (s), medium (m), large (l), and extra-large (x), resulting in 15 distinct architecture--scale combinations. All models are evaluated using the same green-on-green orchard dataset containing instance-level annotations of calyx, fruitlet, and peduncle structures. Rather than introducing a new network architecture, the study examines how YOLO generation, model capacity, and training strategy influence small-object perception under a controlled cross-generation experimental framework.}

Two complementary training strategies are investigated for each experimentally evaluated YOLO architecture and model scale. The first represents a conventional training configuration using $640\times640$-pixel inputs, whereas the second employs a small-object-focused optimization strategy using $960\times960$-pixel inputs together with targeted training and augmentation settings intended to preserve and enhance the representation of small anatomical structures. In the present study, these controlled experiments are conducted for YOLOv8, YOLO11, and YOLO26 across their n, s, m, l, and x model scales. YOLO27 is additionally examined from an architectural perspective as the latest generation considered in this cross-generation analysis; however, because its public trainable segmentation implementation is not yet available, YOLO27 is not included in the current controlled experimental benchmark. The experimental framework has therefore been structured so that YOLO27 can be incorporated under the same dataset partitions, training configurations, evaluation metrics, and computational-analysis protocol once the corresponding public models become available. This design enables systematic investigation of whether increasing the spatial representation of small objects during model optimization improves detection and segmentation performance across YOLO generations and computational capacities. Accordingly, the study aims to: (i) establish a controlled cross-generation experimental benchmark for YOLOv8, YOLO11, and YOLO26 while contextualizing the architectural evolution toward YOLO27; (ii) quantify the influence of model scale from nano to extra-large configurations within the experimentally evaluated generations; (iii) determine the effectiveness of small-object-focused training relative to conventional training; and (iv) characterize predictive-performance and computational-efficiency trade-offs relevant to robotic and precision orchard applications. Following public availability, YOLO27 will be evaluated using the same experimental protocol to extend the benchmark without altering its methodological basis.

The principal contributions of this work are summarized as follows:

\begin{itemize}

\item We present a cross-generation analysis of Ultralytics YOLOv8, YOLO11, YOLO26, and the latest YOLO27 for fine-grained small-object detection and instance segmentation. The controlled experimental component currently benchmarks YOLOv8, YOLO11, and YOLO26, with all five model scales-n, s, m, l, and x systematically investigated for each generation. This results in 15 architecture--scale combinations evaluated using a common orchard dataset and standardized experimental framework. YOLO27 is presently incorporated through architectural analysis and will be added to the controlled benchmark when its corresponding public trainable segmentation models become available.

\item We analyze the architectural progression from YOLOv8 through YOLO11 and YOLO26 to YOLO27, highlighting changes in feature extraction, multi-scale representation, prediction mechanisms, and deployment-oriented design. Particular attention is given to the scale-dependent organization of YOLO27 and its architectural differences relative to the experimentally evaluated predecessors. This separation between architectural analysis and empirical benchmarking allows the study to discuss the latest YOLO generation without attributing experimental results to models that have not yet been evaluated on the orchard dataset.

\item We investigate two complementary model-training strategies for every currently evaluated architecture--scale combination: a conventional $640\times640$-pixel configuration and a small-object-focused $960\times960$-pixel configuration. The higher-resolution strategy is designed to preserve greater spatial information for small anatomical targets during model learning. This controlled comparison provides a systematic assessment of whether resolution-oriented small-object optimization remains beneficial across different YOLO generations and model capacities.

\item We characterize the joint influence of architectural generation, model scale, and training strategy on predictive performance. Detection and instance-segmentation results are compared across YOLOv8, YOLO11, and YOLO26 and their five capacity levels to determine whether improvements associated with small-object-focused optimization remain consistent from lightweight nano models to extra-large networks or depend on the underlying architecture and computational capacity. The same analytical protocol is intended for subsequent YOLO27 evaluation, enabling direct extension of the benchmark.

\item We establish a fine-grained orchard perception benchmark involving three anatomically and visually distinct target classes: calyx, fruitlet, and peduncle. These structures are evaluated under challenging green-on-green orchard conditions characterized by small object footprints, substantial scale variation, partial occlusion, dense vegetation, clustering, and weak foreground--background contrast. This formulation extends conventional whole-fruit perception toward anatomical-level understanding relevant to robotic fruit thinning, selective manipulation, and other precision orchard operations.

\item We provide a deployment-oriented analysis that jointly considers predictive performance and computational characteristics, including model complexity, parameter count, GFLOPs, image-processing time, training duration, and convergence behavior. The resulting analysis characterizes accuracy--efficiency trade-offs across the experimentally evaluated generations, model scales, and training strategies, providing practical guidance for selecting model configurations under different computational and operational constraints. Once experimentally available, YOLO27 will be assessed using the same criteria to enable a consistent comparison with its predecessors.

\item Finally, we establish a reproducible and extensible cross-generation experimental framework encompassing dataset preparation, instance-level annotation, fixed training-validation-test partitioning, pretrained model initialization, conventional and small-object-focused training configurations, augmentation settings, checkpoint selection, performance evaluation, and computational analysis. Importantly, the framework is designed to accommodate subsequent YOLO generations without modifying the underlying dataset or evaluation protocol. YOLO27 can be incorporated once trainable segmentation models become available.

\end{itemize}

\section{Methods}

\begin{figure*}[ht!]
     \centering
     \includegraphics[width= 0.9\linewidth]{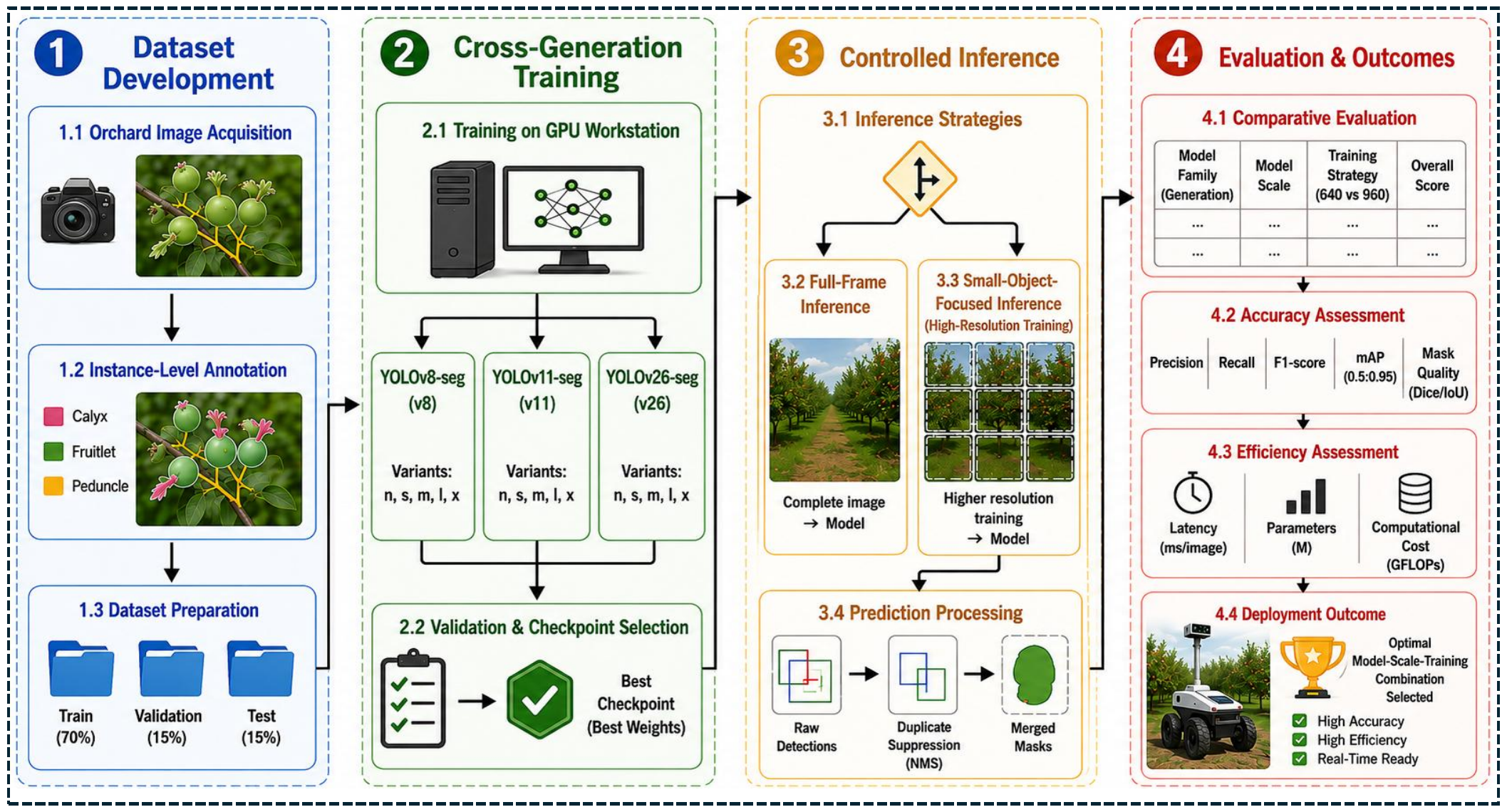}
  \caption{\textbf{Experimental workflow for cross-generation YOLO optimization.} Orchard data preparation is followed by training and validation of YOLOv8, YOLO11, and YOLO26 variants under conventional and small-object-focused configurations, with subsequent comparative assessment of detection, segmentation, computational efficiency, and deployment suitability.}
    \label{fig:methods}
\end{figure*}

\textcolor{black}{A structured multi-stage experimental methodology was developed to comprehensively compare and optimize Ultralytics YOLOv8, YOLO11, and YOLO26 for fine-grained small-object detection and instance segmentation in complex orchard environments, as illustrated in Fig.~\ref{fig:methods}. First, orchard images representing challenging green-on-green conditions were collected and curated, followed by instance-level annotation of three anatomically distinct classes: calyx, fruitlet, and peduncle---and organization into fixed training, validation, and test subsets. Second, all five segmentation scales: nano (n), small (s), medium (m), large (l), and extra-large (x), of each YOLO generation were systematically investigated, resulting in 15 architecture--scale combinations. Each combination was evaluated under two complementary training strategies: a conventional configuration using $640\times640$-pixel inputs and a small-object-focused configuration using $960\times960$-pixel inputs with targeted training and augmentation settings intended to improve the representation and learning of small anatomical structures. This factorial organization resulted in 30 experimental model-training configurations and enabled the effects of YOLO generation, model capacity, and training strategy to be examined within a unified experimental framework. Pretrained segmentation weights corresponding to each architecture and scale were fine-tuned using consistent dataset partitions and controlled optimization procedures, and the best-performing checkpoint from each training run was retained for evaluation. Finally, the resulting models were systematically compared using detection and instance-segmentation metrics, class-wise performance, model complexity, training and convergence characteristics, and computational-efficiency measures. The overall methodology therefore provides a controlled framework for determining whether small-object-focused optimization consistently improves fine-grained perception across successive YOLO generations and model capacities, while simultaneously characterizing the accuracy-efficiency trade-offs relevant to robotic perception and precision orchard applications.}

\begin{figure}[ht!]
     \centering
     \includegraphics[width=\linewidth]{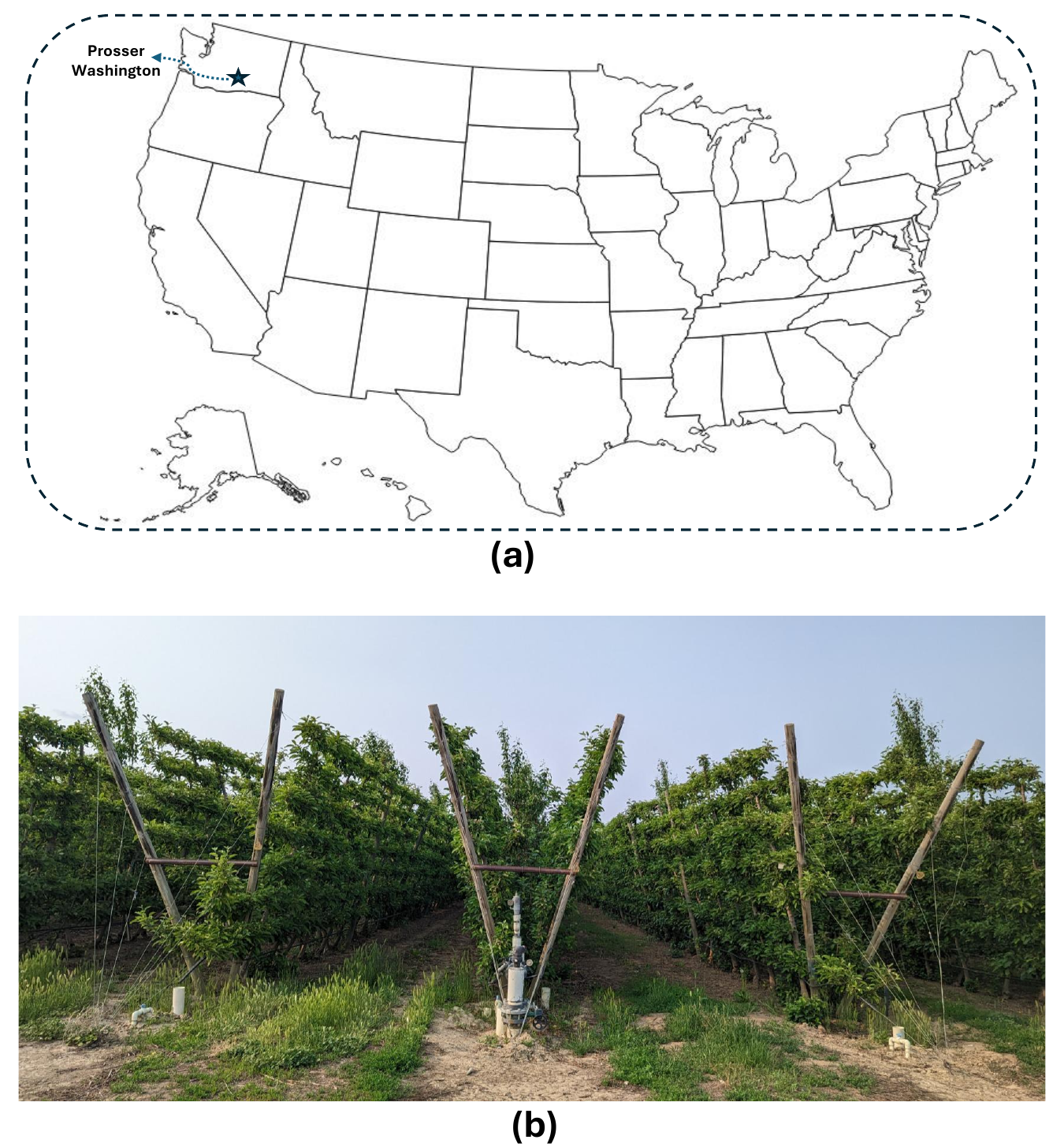}
    \caption{Overview of the study site and data collection environment for small-object detection and segmentation; (a) Geographical map of Washington State highlighting the location where the data were collected (Prosser, WA); b) Commercial apple orchard used in this study, illustrating the real-world field conditions during image acquisition.}
    \label{fig:site}
\end{figure}

\subsection{Data collection and Preparation}
\subsubsection{Image Acquisition and Study Site}

High-resolution RGB images were acquired from a commercial apple orchard located in Prosser, Washington State, USA (Figure~\ref{fig:site}), using an \textit{iPhone 14 Pro} imaging sensor. Image acquisition was conducted on \textit{May 6, 2024}, during the early fruit development stage and prior to commercial fruitlet thinning operations. This timing was deliberately selected, as fruitlets are smallest at this stage and exhibit minimal visual contrast with the surrounding canopy, representing one of the most challenging regimes for small-object detection and segmentation.

\subsubsection{Class Definitions and Annotation Procedure}

Each image contains three fine-grained anatomical components of early-stage apple fruitlets (Refer to  Figure \ref{fig:greenfruits}): \textit{calyx}, \textit{fruitlet} (main fruit body), and \textit{peduncle}. These structures are compact, frequently occluded, and visually similar to surrounding leaves, stems, and branches, making their identification particularly difficult. Accurately distinguishing these components is essential for downstream applications such as robotic fruitlet thinning, where precise knowledge of stem attachment points and anatomical boundaries is required to enable safe and targeted manipulation of target fruitlet.

All 600 selected images were manually annotated using the Roboflow annotation platform. Due to the small size, close spatial proximity, and multi-class nature of the objects, annotation was labor-intensive and required careful visual inspection. A team of four human annotators performed the labeling over a period of approximately two months. To ensure annotation quality and anatomical correctness, all labeled instances were cross-verified by a horticulture domain expert, focusing on class consistency, tight spatial coverage, and accurate separation from visually similar background structures. In total, the dataset contains 3,706 labeled instances of \textit{calyx}, 3,714 instances of \textit{fruitlet}, and 839 instances of \textit{peduncle}, reflecting the inherent class imbalance and varying visibility/occlusion of anatomical components.

\subsubsection{Dataset Organization and Preparation}
Following annotation, the dataset was partitioned into training, validation, and test subsets comprising 503 (83.8\%), 49 (8.2\%), and 48 (8.0\%) images, respectively. This split was designed to ensure sufficient data for model learning while maintaining independent subsets for hyperparameter tuning and unbiased performance evaluation. The annotated dataset was then exported into the \textcolor{black}{YOLOv8, YOLO11 and YOLO26} compatible format to support both detection and instance segmentation training. The finalized dataset and all accompanying training and inference code have been made publicly available to support reproducibility and further research, with access provided via the project repository (\href{https://github.com/rnjnspkt/YOLO26-Optimization-with-Slicing-Aided-Hyper-Inference-for-Small-Object-Detection-and-Segmentation}{GitHub Link}).

\subsection{Cross-Generation YOLO Architectures for Small-Object Detection and Instance Segmentation}
YOLO27, YOLO26, YOLO11, and YOLOv8 represent successive generations within the Ultralytics YOLO ecosystem and are considered to characterize the architectural evolution of modern YOLO detectors for fine-grained small-object detection and instance segmentation. Among these models, YOLOv8, YOLO11, and YOLO26 form the controlled experimental benchmark, whereas YOLO27 is included for architectural analysis pending availability of its public trainable segmentation implementation. Although these model families retain the general backbone--neck--prediction organization of contemporary one-stage detectors, they differ substantially in feature-extraction blocks, multi-scale feature fusion, prediction mechanisms, training strategies, and deployment characteristics \cite{palaniappan2025yolo,casas2023assessing,hidayatullah2025yolov8,safonova2026deep}. The backbone transforms input imagery into hierarchical feature representations at multiple spatial resolutions, the neck integrates complementary spatial and semantic information across feature levels, and the task-specific prediction components generate localization, classification, and instance-mask outputs. YOLO27 further extends this evolution through a scale-dependent architectural organization; particularly, the compact YOLO27n/s variants employ a streamlined CNN-based pathway with strengthened high-resolution representations, FPN/PAN fusion, and dual-scale prediction, as illustrated in Figure~\ref{fig:27architecture}. These architectural differences provide the basis for examining whether successive YOLO generations, alternative prediction mechanisms, and increasing model capacity improve the representation of small, low-contrast anatomical structures under complex orchard conditions. The detailed YOLO27n/s architecture is presented separately in Figure~\ref{fig:27architecture}, yet to experiment in this version while the experimentally evaluated YOLOv8, YOLO11, and YOLO26 architectures are illustrated in Fig.~\ref{fig:architecture}(a)-(c).


\subsubsection{Ultralytics YOLO27}

\begin{figure*}[h!]
    \centering
    \includegraphics[width=\linewidth]{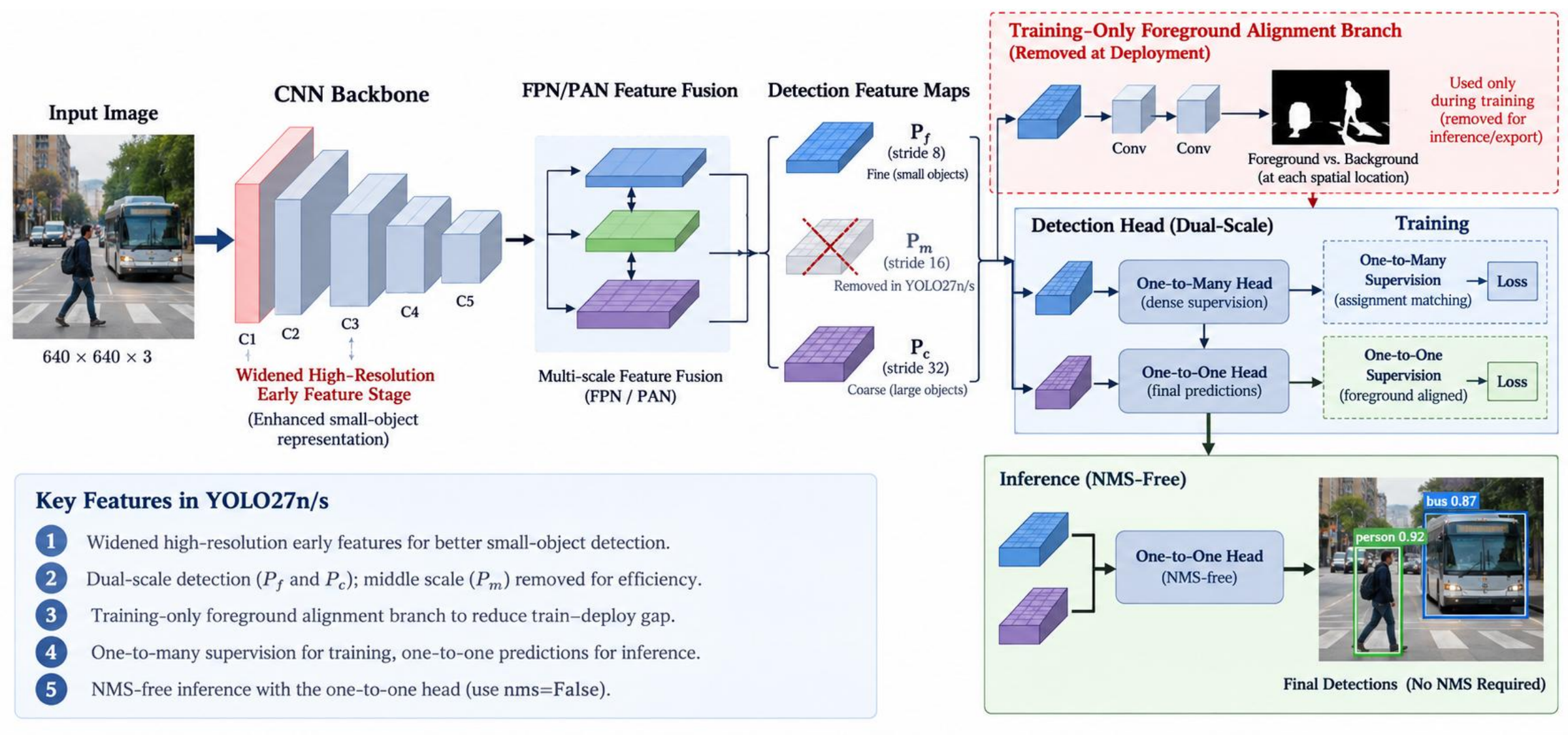}
    \caption{\textbf{Detailed architecture of YOLO27n/s.}
    The architecture incorporates a streamlined CNN backbone, widened
    high-resolution features, FPN/PAN feature fusion, and dual-scale
    prediction using fine (\(P_f\)) and coarse (\(P_c\)) feature maps.
    Training combines one-to-many supervision with a foreground-alignment
    branch, whereas deployment uses one-to-one predictions for efficient
    NMS-free inference.}
    \label{fig:27architecture}
\end{figure*}

Figure~\ref{fig:27architecture} summarizes the compact YOLO27n/s architecture and highlights the design choices that differentiate these variants from conventional uniformly scaled YOLO detectors. The backbone follows a streamlined convolutional design while preserving strengthened high-resolution representations, which are subsequently integrated through FPN/PAN feature fusion to support small- and multi-scale object localization. As illustrated in Figure~\ref{fig:27architecture}, prediction is performed from two principal feature levels, namely a fine-resolution map \(P_f\) and a coarser map \(P_c\), allowing the network to balance localization detail with broader contextual information. The training strategy further combines one-to-many supervision with a dedicated foreground-alignment branch, thereby providing richer supervisory signals during optimization while keeping the deployment pathway comparatively simple. In contrast to the training configuration, inference relies on one-to-one predictions, enabling direct prediction without conventional non-maximum suppression and thereby reducing post-processing overhead. This separation between training-time supervision and deployment-time prediction is important for real-time applications because model capacity can be concentrated on representation learning during training while maintaining a simplified inference graph during deployment. For agricultural perception, such a design is particularly relevant when small structures must be recognized under clutter, occlusion, and limited pixel resolution, because strengthened high-resolution features and the \(P_f\)--\(P_c\) dual-scale pathway can preserve localized information that may otherwise be attenuated in deeper feature hierarchies. Overall, Figure~\ref{fig:27architecture} illustrates that YOLO27n/s emphasize an efficiency-oriented CNN pathway in which high-resolution feature preservation, multi-scale fusion, dual-scale prediction, and NMS-free deployment are jointly integrated within a compact detection architecture.

\subsubsection{Ultralytics YOLO26}

\begin{figure}[h!]
    \centering
    \includegraphics[width=0.99\linewidth]{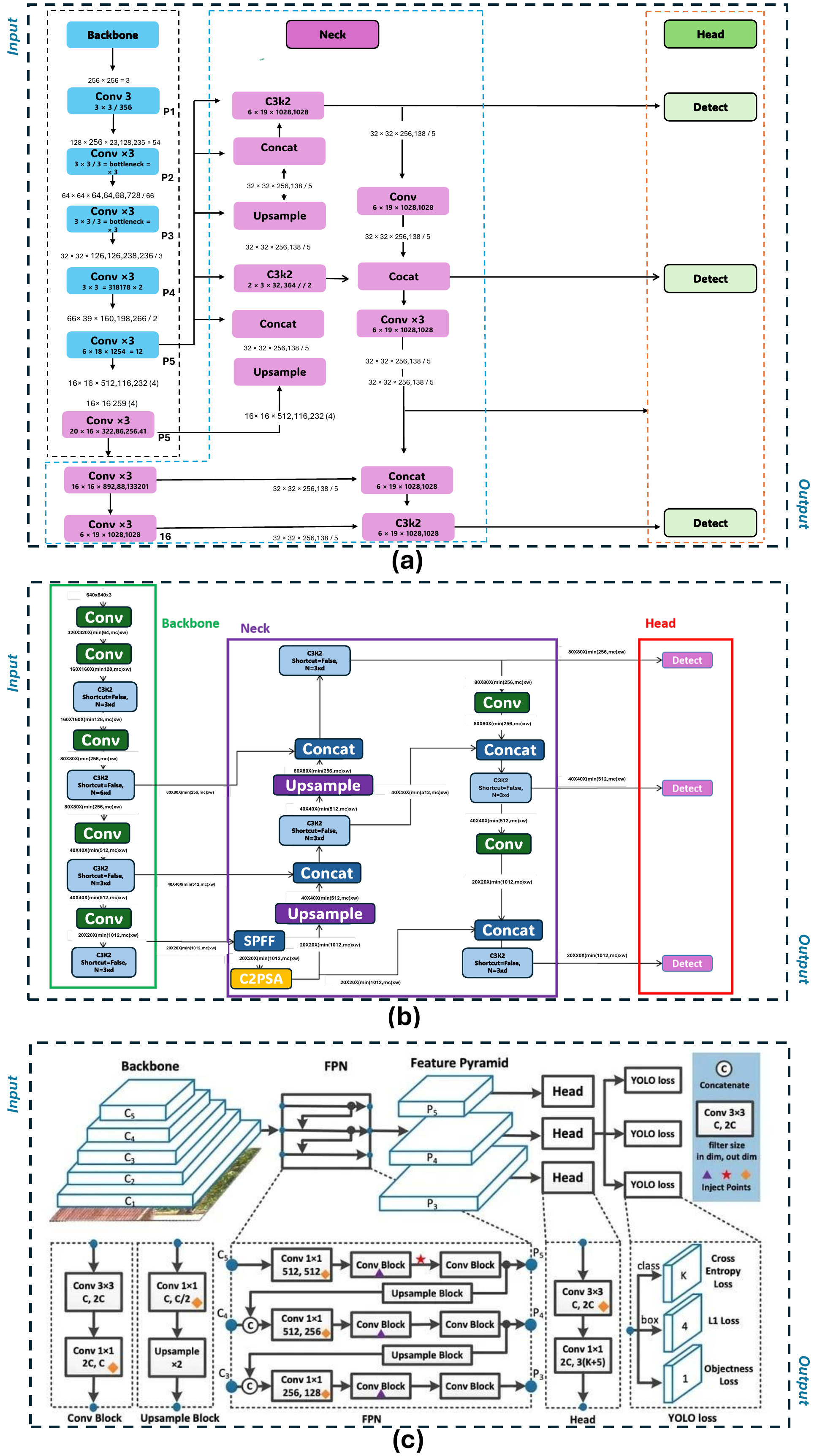}
    \caption{\textcolor{black}{Architectural overview of the model families and deployment concept considered in this study: (a) YOLO26, (b) YOLO11, and (c) YOLOv8}}
    \label{fig:architecture}
\end{figure}

YOLO26, illustrated in Fig.~\ref{fig:architecture}(a), represents the most recent Ultralytics generation evaluated in the cross-generation benchmark and places particular emphasis on deployment efficiency, simplified inference, and end-to-end prediction \cite{jocher2026ultralytics}. The architecture retains hierarchical multi-scale feature extraction, whereby progressively transformed representations encode complementary spatial and semantic information before being supplied to task-specific detection and segmentation components. Such multi-scale representation is particularly relevant to the present dataset because calyx, fruitlet, and peduncle instances differ substantially in apparent size and morphology; in particular, peduncles frequently appear as narrow structures occupying only a small fraction of the image.
\textcolor{black}{A distinguishing feature of YOLO26 is its end-to-end prediction design. In contrast to conventional detection pipelines that depend exclusively on a separate non-maximum suppression (NMS) stage for eliminating redundant detections, YOLO26 supports one-to-one prediction for NMS-free end-to-end deployment \cite{sapkota2025ultralytics, sapkota2026yoloe, sapkota2025yolo26}. The architecture additionally supports one-to-many prediction during optimization, providing denser supervision while retaining the deployment-oriented one-to-one prediction pathway \cite{jocher2026ultralytics}. For instance segmentation, object predictions are associated with learned mask representations, allowing the network to jointly perform localization, classification, and instance-level delineation \cite{jocher2026ultralytics}. This capability is important for the present application because successful perception requires not only identification of fruitlet anatomical components but also recovery of their spatial boundaries.}

\textcolor{black}{YOLO26 further introduces changes to its localization and optimization framework, including removal of Distribution Focal Loss (DFL) from the localization formulation and training-oriented developments such as ProgLoss, Spatial-Temporal Adaptive Learning (STAL), and MuSGD \cite{sapkota2025yolo26}. These modifications are intended to simplify deployment and improve optimization behavior across object scales. To determine how these architectural developments translate to fine-grained orchard perception, all five available segmentation scales---YOLO26n, YOLO26s, YOLO26m, YOLO26l, and YOLO26x---were included in the benchmark.}


\subsubsection{Ultralytics YOLO11}

\textcolor{black}{YOLO11, shown in Fig.~\ref{fig:architecture}(b), provides an intermediate architectural generation between YOLOv8 and YOLO26. It retains hierarchical backbone--neck--head processing while incorporating updated feature-extraction and feature-fusion components intended to improve representational efficiency. Spatial resolution is progressively reduced through the backbone as feature abstraction increases, after which representations from different depths are integrated to preserve both localization-sensitive and semantically informative features. This combination is relevant to green-on-green orchard imagery, where discrimination among fruitlets, calyxes, peduncles, leaves, and surrounding vegetation frequently depends on subtle differences in shape, boundaries, texture, and local context.}

\textcolor{black}{YOLO11 employs C3k2-based feature-processing blocks that provide hierarchical feature transformation while controlling computational complexity \cite{wang2025street, du2025yolo}. Its feature-processing pathway further incorporates mechanisms such as SPPF and C2PSA to increase effective receptive-field coverage and strengthen spatial feature representation \cite{el2025advanced, liu2026yolo}. Multi-scale feature fusion subsequently combines deeper semantic information with higher-resolution spatial representations before prediction. These characteristics make YOLO11 particularly informative as an intermediate reference for determining whether architectural progression translates into measurable improvements for small and visually ambiguous anatomical structures.}

\textcolor{black}{The YOLO11-seg models extend object detection with instance-mask prediction, enabling simultaneous localization, classification, and delineation of individual objects \cite{yao2025research, maskeliunas2025hybrid}. Accordingly, each predicted instance can be assigned to calyx, fruitlet, or peduncle while retaining its object-specific spatial mask. Five model capacities:YOLO11n, YOLO11s, YOLO11m, YOLO11l, and YOLO11x were evaluated using the same dataset organization and common training framework employed for the other YOLO generations, enabling model capacity and architectural generation to be examined within a unified experimental design.}

\subsubsection{Ultralytics YOLOv8}

\textcolor{black}{YOLOv8, illustrated in Fig.~\ref{fig:architecture}(c), serves as the earlier-generation Ultralytics reference in the cross-generation comparison. YOLOv8 employs an anchor-free detection formulation together with a decoupled prediction design, reducing dependence on predefined anchor boxes and allowing localization and classification to be handled through dedicated prediction pathways \cite{sohan2024review}. Its backbone is organized around convolutional feature extraction and C2f modules that facilitate feature reuse and gradient propagation while maintaining computational efficiency \cite{safaldin2024improved, khow2024improved}. As features propagate through the network, spatial resolution decreases while channel depth and semantic abstraction progressively increase.}

\textcolor{black}{The YOLOv8 neck integrates representations across multiple spatial scales through feature-pyramid-style operations involving upsampling, concatenation, and subsequent feature transformation \cite{luo2025sfw, zhou2024detection}. This multi-scale aggregation is particularly important for the present dataset because comparatively large fruitlet regions and substantially smaller calyx or peduncle structures may occur simultaneously within the same orchard image. Combining higher-resolution spatial features with deeper semantic representations enables the prediction head to reason over objects exhibiting markedly different pixel footprints \cite{wang2026enhanced, zenghao2026fast}.}

\textcolor{black}{YOLOv8-seg augments this detection framework with instance-segmentation functionality based on learned prototype masks and instance-specific mask coefficients \cite{bai2023automated, yue2023improved, sapkota2024comparing}. The resulting representation permits simultaneous bounding-box localization and pixel-level delineation of individual anatomical structures. YOLOv8 therefore provides an established reference against which the architectural and computational progression of YOLO11 and YOLO26 can be evaluated. Consistent with the other generations, YOLOv8n, YOLOv8s, YOLOv8m, YOLOv8l, and YOLOv8x were included, resulting in five YOLOv8 configurations within the benchmark.}


\subsubsection{Cross-Generation Training and Small-Object-Focused Optimization}

\textcolor{black}{The experimental design comprised three Ultralytics YOLO generations---YOLOv8, YOLO11, and YOLO26---with five model-capacity levels within each generation: nano (n), small (s), medium (m), large (l), and extra-large (x). This resulted in 15 distinct architecture--scale combinations for cross-generation evaluation. Let the YOLO generation be denoted by $g \in \{\mathrm{v8},\mathrm{v11},\mathrm{v26}\}$ and the model scale by $k \in \{n,s,m,l,x\}$. For an input orchard image $I$, the prediction process for a given segmentation model can be expressed generally as}

\begin{equation}
\textcolor{black}{
\mathcal{Y}^{(g,k)}
=
f_{\theta^{(g,k)}}(I),
}
\end{equation}

\textcolor{black}{where $f_{\theta^{(g,k)}}$ denotes the corresponding YOLO instance-segmentation network parameterized by learned weights $\theta^{(g,k)}$, and $\mathcal{Y}^{(g,k)}$ represents the resulting class predictions, bounding-box coordinates, confidence scores, and instance masks. This generalized formulation provides a common representation of the prediction task across the three YOLO generations despite differences in their internal architectural components. Each network was fine-tuned for the same three anatomical classes---calyx, fruitlet, and peduncle---using identical dataset partitions to support controlled cross-generation comparison.}

\textcolor{black}{During supervised training, the model parameters were optimized using the class, localization, and instance-mask annotations available in the training dataset. The optimization process can be represented generally as}

\begin{equation}
\textcolor{black}{
\theta^{*}_{g,k}
=
\arg\min_{\theta_{g,k}}
\mathcal{L}
\left(
f_{\theta_{g,k}}(I),Y
\right),
}
\end{equation}

\textcolor{black}{where $Y$ denotes the corresponding ground-truth class, bounding-box, and instance-mask annotations, and $\mathcal{L}$ represents the composite training objective implemented by the respective Ultralytics YOLO architecture. The internal loss formulations of YOLOv8, YOLO11, and YOLO26 were retained without manual modification. Accordingly, the experimental comparison focused on fine-tuning the publicly available pretrained segmentation architectures rather than introducing changes to their internal loss functions or prediction heads.}

\textcolor{black}{Two complementary training configurations were investigated for every architecture--scale combination. The conventional configuration used an input resolution of $640\times640$ pixels and served as the reference training condition. The small-object-focused configuration increased the input resolution to $960\times960$ pixels to preserve greater spatial information for compact and geometrically fine anatomical structures. Thus, the 15 architecture--scale combinations were examined under two training conditions, resulting in 30 model--resolution configurations. This factorial organization enabled the effects of YOLO generation, model capacity, and small-object-focused optimization to be examined systematically within a common experimental framework.}

\textcolor{black}{The motivation for the higher-resolution configuration arises from the inherent difficulty of representing small objects during deep feature extraction. Small anatomical structures can lose discriminative information during image resizing and successive feature-map downsampling, particularly when their original spatial footprint is limited \cite{nikouei2025small, nisa2025image, habash2025recent}. This limitation is especially relevant to the calyx and peduncle classes considered in this study. Peduncles are typically narrow and elongated, whereas calyx structures occupy relatively compact and irregular regions. Both classes can additionally exhibit weak visual contrast against green fruitlets, leaves, stems, and surrounding vegetation. Increasing the training resolution from $640\times640$ to $960\times960$ pixels was therefore investigated as a direct strategy for retaining more spatial information associated with these challenging anatomical targets.}

\textcolor{black}{Conceptually, the benefit of increasing image resolution can be interpreted through the pixel representation of an object within the network input. Let an anatomical target occupy a normalized spatial region $\Omega_o$ in the original image. When the same scene is represented at two square input resolutions $r_1$ and $r_2$, where $r_2>r_1$, the approximate number of input pixels representing the target increases proportionally to the square of the resolution ratio:}

\begin{equation}
\textcolor{black}{
\frac{N_o^{(r_2)}}{N_o^{(r_1)}}
\approx
\left(\frac{r_2}{r_1}\right)^2.
}
\end{equation}

\textcolor{black}{For the resolutions investigated in this study, increasing the network input from $640\times640$ to $960\times960$ pixels corresponds to a 1.5-fold increase along each spatial dimension and, before subsequent network downsampling, approximately a 2.25-fold increase in the number of input pixels representing an object of the same relative image area. The higher-resolution configuration was therefore intended to provide the network with a richer initial spatial representation of small structures, although the extent to which this additional information is preserved and translated into improved predictions depends on the architecture, feature hierarchy, model capacity, and optimization process.}

\textcolor{black}{Both training configurations retained the same underlying YOLO architectures and pretrained initialization. The conventional $640\times640$ configuration employed the standard training workflow, whereas the $960\times960$ configuration incorporated the small-object-focused training settings described in Table~\ref{tab:training_config}, including the explicitly configured scale, mosaic, and late-stage mosaic-closure parameters. These adjustments were intended to expose the models to spatial and scale variability while allowing optimization to stabilize on complete images during the final training epochs. Importantly, the higher-resolution configuration did not introduce new network layers, prediction heads, or loss functions; consequently, differences between the two configurations primarily reflect changes in the spatial representation and associated training configuration rather than modification of the fundamental YOLO architecture.}

\textcolor{black}{A maximum optimization schedule of 200 epochs was specified for the principal training runs, with validation performed throughout training and checkpoint selection based on validation performance. Where validation-based early stopping was activated through the framework configuration, training could terminate before reaching the maximum epoch budget when the monitored validation criterion ceased to improve for the specified patience interval. The best-performing checkpoint produced during each training run was retained for subsequent evaluation rather than relying exclusively on the weights from the final epoch. This procedure reduced the influence of late-stage overfitting and provided a consistent basis for comparing models with different convergence characteristics.}

\textcolor{black}{The resulting experimental framework therefore separates three factors that may influence fine-grained orchard perception: architectural generation, represented by YOLOv8, YOLO11, and YOLO26; model capacity, represented by the n, s, m, l, and x variants; and training configuration, represented by conventional $640\times640$ and small-object-focused $960\times960$ optimization. Detection accuracy, instance-segmentation performance, convergence behavior, training duration, model complexity, and processing efficiency were subsequently compared across these configurations. This design enables the study to determine whether increased input resolution consistently benefits small-object perception, whether such improvements depend on YOLO generation or model capacity, and whether the associated computational cost is justified by corresponding gains in the detection and segmentation of fine anatomical structures.}

\subsection{Training Environment and Cross-Generation Model Configuration}

\textcolor{black}{All YOLOv8, YOLO11, and YOLO26 instance-segmentation models were trained and evaluated on a common GPU-enabled computational platform to ensure consistency across model generations and capacity levels. The workstation was equipped with an AMD Ryzen Threadripper 7960 processor containing 24 physical cores and 48 processing threads, 256~GiB of system memory, and 9.7~TB of storage. GPU acceleration was provided by an NVIDIA RTX 5000 Ada Generation graphics processor, with 32,237~MiB of CUDA-accessible memory reported during model execution. This hardware configuration provided sufficient memory and computational throughput to train the nano (n), small (s), medium (m), large (l), and extra-large (x) segmentation variants of all three YOLO generations. The complete hardware and software environment is summarized in Table~\ref{tab:compute_env}.}

\textcolor{black}{The software stack consisted of Ubuntu 22.04.5 LTS, Python 3.10.12, PyTorch 2.1.0+cu121, and the Ultralytics framework, with model optimization executed through the PyTorch/CUDA backend on GPU device 0. Automatic Mixed Precision (AMP) was enabled during the principal GPU training runs to reduce memory demand and improve computational throughput. Image caching was disabled (\texttt{cache=False}), and four data-loading workers were used to parallelize image loading and preprocessing. These implementation settings, together with the workstation specifications listed in Table~\ref{tab:compute_env}, define the computational environment under which the cross-generation benchmark was reproduced.}

\begin{table}[ht!]
\centering
\caption{\textbf{\textcolor{black}{Hardware and software environment used for cross-generation YOLO training and evaluation.}}}
\label{tab:compute_env}
{\color{black}
\begin{tabular}{p{0.18\textwidth} p{0.26\textwidth}}
\toprule
\textbf{Component} & \textbf{Specification} \\
\midrule
Processor (CPU) & AMD Ryzen Threadripper 7960 \\
CPU cores / threads & 24 cores / 48 threads \\
System memory & 256~GiB RAM \\
Graphics processor & NVIDIA RTX 5000 Ada Generation \\
Reported GPU memory & 32,237~MiB CUDA memory ($\sim$33.8~GB device-reported capacity) \\
GPU device & CUDA device 0 \\
Operating system & Ubuntu 22.04.5 LTS (64-bit) \\
Desktop environment & GNOME 42.9 \\
Windowing system & X11 \\
Storage capacity & 9.7~TB \\
Python & 3.10.12 \\
PyTorch & 2.1.0+cu121 \\
Ultralytics & Version 8.4.43 \\
Deep-learning backend & PyTorch / CUDA \\
Training task & Multi-class instance segmentation \\
Target classes & Calyx, fruitlet, and peduncle \\
Automatic Mixed Precision & Enabled for principal GPU training runs \\
Image caching & Disabled (\texttt{cache=False}) \\
\bottomrule
\end{tabular}
}
\end{table}

\textcolor{black}{The experimental benchmark comprised 15 segmentation models: YOLOv8n/s/m/l/x, YOLO11n/s/m/l/x, and YOLO26n/s/m/l/x. Each architecture was initialized from its corresponding pretrained Ultralytics segmentation checkpoint (\texttt{pretrained=True}) and fine-tuned on the same three-class orchard dataset. The dataset definition, class ordering, and training, validation, and test partitions were held fixed across all model families so that differences in predictive and computational behavior could be examined under comparable data exposure.}

\textcolor{black}{Each of the 15 architecture–scale combinations was independently trained under 640×640 and 960×960 input-resolution configurations, resulting in 30 model-training runs. Validation was enabled throughout optimization (\texttt{val=True}) to monitor generalization performance, while model checkpoints were saved using \texttt{save=True}. Intermediate checkpoints were retained at 10-epoch intervals (\texttt{save\_period=10}), and training curves were generated using \texttt{plots=True}. A patience value of 100 epochs was used for early stopping; consequently, some model configurations completed the full 200-epoch schedule, whereas others terminated earlier when validation performance failed to improve over the specified patience interval. In all cases, the best validation checkpoint rather than the final training epoch was retained for subsequent analysis.}

\textcolor{black}{Optimizer selection was specified using \texttt{optimizer="auto"}, allowing the Ultralytics framework to select the optimization algorithm and associated effective parameters at runtime. In the recorded YOLOv8 and YOLO11 training logs, this procedure resolved to AdamW with an initial learning rate of 0.001429 and momentum of 0.9. A weight decay of 0.0005 was applied to the corresponding weight parameter group, together with a final learning-rate factor of 0.01, three warm-up epochs, a warm-up momentum of 0.8, and a warm-up bias learning rate of 0.1. Because the optimizer was framework-resolved, the values recorded during execution provide the appropriate description of the effective optimization procedure. The complete training configuration is reported in Table~\ref{tab:training_config}.}

\textcolor{black}{Data augmentation was applied consistently to improve robustness to the scale variability, illumination differences, and background clutter characteristic of orchard imagery. The recorded configuration employed mosaic augmentation with probability 1.0, scale augmentation of 0.5, translation of 0.1, and horizontal flipping with probability 0.5. HSV hue, saturation, and value perturbations were set to 0.015, 0.7, and 0.4, respectively. Vertical flipping, MixUp, copy-paste augmentation, rotation, shear, and perspective transformations were disabled. Mosaic augmentation was discontinued during the final 10 epochs (\texttt{close\_mosaic=10}), allowing late-stage optimization to proceed using complete, non-mosaicked images. For instance segmentation, overlapping masks were enabled (\texttt{overlap\_mask=True}) with a mask ratio of 4. These augmentation and segmentation-specific parameters are also consolidated in Table~\ref{tab:training_config}.}

\textcolor{black}{Additional reproducibility controls included a fixed random seed of 42, \texttt{deterministic=False}, a maximum detection count of 300, disabled image caching, and AMP-enabled GPU training. Collectively, the common input resolution, fixed dataset partitions, pretrained initialization, shared optimization settings, standardized augmentation pipeline, validation-based checkpoint selection, and consistent computational environment provided a standardized experimental framework for comparing the effects of YOLO generation and model capacity. Table~\ref{tab:compute_env} documents the execution environment, whereas Table~\ref{tab:training_config} summarizes the model-training and augmentation parameters.}

\begin{table*}[ht!]
\centering
\caption{\textbf{\textcolor{black}{Training and augmentation configurations used for the cross-generation YOLO experiments. The conventional configuration used $640\times640$-pixel inputs, whereas the small-object-focused configuration used $960\times960$-pixel inputs with additional augmentation controls. Parameters identified as framework-resolved were obtained from the recorded Ultralytics training configuration.}}}
\label{tab:training_config}

{\color{black}
\resizebox{0.98\textwidth}{!}{
\begin{tabular}{llll}
\toprule
\textbf{Training Component} &
\textbf{Conventional Configuration} &
\textbf{Small-Object Configuration} &
\textbf{Implementation Detail} \\
\midrule

YOLO families &
YOLOv8 / YOLO11 / YOLO26 &
YOLOv8 / YOLO11 / YOLO26 &
Instance-segmentation models \\

Model scales &
n, s, m, l, x &
n, s, m, l, x &
Five scales per generation \\

Total architectures &
15 &
15 &
Three generations $\times$ five scales \\

Model initialization &
Pretrained segmentation weights &
Pretrained segmentation weights &
\texttt{pretrained=True} \\

Input image size &
$640\times640$ &
$960\times960$ &
Square model-training input \\

Maximum training epochs &
200 &
200 &
Nominal principal training schedule \\

Early-stopping patience$^{a}$ &
100 epochs &
100 epochs &
Ultralytics default if not overridden \\

Batch size &
8 &
8 &
Principal GPU training \\

Data-loader workers &
4 &
2 &
Explicitly defined in training scripts \\

Optimizer specification &
Auto &
Auto &
\texttt{optimizer="auto"} \\

Resolved optimizer$^{b}$ &
AdamW &
AdamW &
Automatically selected by Ultralytics \\

Resolved initial learning rate$^{b}$ &
0.001429 &
0.001429 &
Recorded in inspected logs \\

Resolved momentum$^{b}$ &
0.9 &
0.9 &
Recorded in inspected logs \\

Final learning-rate factor &
0.01 &
0.01 &
Framework/default configuration \\

Weight decay &
0.0005 &
0.0005 &
Applicable weight parameter group \\

Warm-up epochs &
3.0 &
3.0 &
Framework-resolved setting \\

Warm-up momentum &
0.8 &
0.8 &
Framework-resolved setting \\

Warm-up bias learning rate &
0.1 &
0.1 &
Framework-resolved setting \\

Mosaic augmentation &
1.0 &
1.0 &
Enabled during training \\

Close mosaic &
10 epochs &
10 epochs &
Mosaic disabled during final 10 epochs \\

Scale augmentation &
0.5 &
0.5 &
Explicitly specified for small-object configuration$^{c}$ \\

Translation &
0.1 &
0.1 &
Framework/default augmentation \\

Horizontal flip &
0.5 &
0.5 &
\texttt{fliplr=0.5} \\

Vertical flip &
0.0 &
0.0 &
Disabled \\

HSV hue &
0.015 &
0.015 &
Color-space augmentation \\

HSV saturation &
0.7 &
0.7 &
Color-space augmentation \\

HSV value &
0.4 &
0.4 &
Color-space augmentation \\

Rotation &
0.0 &
0.0 &
Disabled \\

Shear &
0.0 &
0.0 &
Disabled \\

Perspective &
0.0 &
0.0 &
Disabled \\

MixUp &
0.0 &
0.0 &
Disabled \\

Copy-paste &
0.0 &
0.0 &
Disabled \\

Dropout &
0.0 &
0.0 &
Framework configuration \\

Mask ratio &
4 &
4 &
Instance-segmentation mask setting \\

Overlapping masks &
Enabled &
Enabled &
\texttt{overlap\_mask=True} \\

Maximum detections &
300 &
300 &
Framework evaluation setting \\

Validation during training &
Enabled &
Enabled &
\texttt{val=True} \\

Checkpoint saving &
Enabled &
Enabled &
\texttt{save=True} \\

Checkpoint interval &
10 epochs &
10 epochs &
\texttt{save\_period=10} \\

Automatic Mixed Precision &
Enabled &
Enabled &
AMP-enabled principal GPU runs \\

Image caching &
Disabled &
Disabled &
\texttt{cache=False} \\

Training plots &
Enabled &
Enabled &
\texttt{plots=True} \\

Random seed$^{d}$ &
42 &
Framework/default &
Explicitly specified only in conventional script \\

Deterministic execution$^{d}$ &
Disabled &
Framework/default &
Explicitly specified only in conventional script \\

Fallback batch size &
4 &
4 &
Used only if principal training failed \\

Fallback workers &
0 &
0 &
Safe-mode configuration \\

Fallback AMP &
Disabled &
Disabled &
Safe-mode configuration \\

Fallback epochs &
200 &
100 &
Used only if fallback routine was triggered \\

\bottomrule
\end{tabular}
}
}

\vspace{1mm}
{\color{black}\footnotesize
$^{a}$ Early stopping followed the Ultralytics training configuration used during model optimization.

$^{b}$ With \texttt{optimizer="auto"}, the recorded YOLOv8, YOLO11, and YOLO26 runs resolved to AdamW with an initial learning rate of 0.001429 and momentum of 0.9.

$^{c}$ The small-object-focused configuration explicitly employed \texttt{scale=0.5}, \texttt{mosaic=1.0}, and \texttt{close\_mosaic=10} to support robust learning of fine-scale targets; corresponding conventional settings followed the Ultralytics training configuration.

$^{d}$ Reproducibility settings were managed through the training configuration, with \texttt{seed=42} and \texttt{deterministic=False} explicitly specified for the conventional runs.
}

\end{table*}

\subsection{Implementation Workflow for Cross-Generation YOLO Benchmarking}

\textcolor{black}{The experimental workflow was designed to provide a controlled comparison of Ultralytics YOLOv8, YOLO11, and YOLO26 across five model scales: nano (n), small (s), medium (m), large (l), and extra-large (x) as illustrated in Fig.~\ref{fig:workflow}. A fixed orchard dataset containing instance-level annotations of calyx, fruitlet, and peduncle was used with identical training, validation, and test partitions for all models. The resulting 15 architecture--scale combinations were initialized from their corresponding pretrained segmentation checkpoints and evaluated under two training configurations: a conventional $640\times640$-pixel setting and a small-object-focused $960\times960$-pixel setting, yielding 30 model--resolution configurations.}

\textcolor{black}{Both training configurations followed the common optimization framework summarized in Table~\ref{tab:training_config}. The principal schedule used a maximum of 200 epochs, batch size 8, validation-based monitoring, pretrained initialization, automatic optimizer selection, AMP-enabled GPU execution, and checkpoint saving at 10-epoch intervals. Early stopping with a patience of 100 epochs was used, and the best validation checkpoint from each run was retained for subsequent evaluation. Training duration and completed epoch count therefore varied according to model convergence behavior.}

\textcolor{black}{All retained models were assessed using the same quantitative framework. Predictive performance was characterized using precision, recall, F1-score, box mAP$_{50}$, box mAP$_{50:95}$, mask mAP$_{50}$, and mask mAP$_{50:95}$. Computational and training characteristics included layer count, trainable parameters, GFLOPs, checkpoint size, preprocessing time, inference time, postprocessing time, completed epochs, best epoch, and wall-clock training duration. This common evaluation protocol enabled direct comparison across YOLO generation, model scale, and training configuration.}

\textcolor{black}{The analysis therefore examined three complementary factors: architectural progression from YOLOv8 to YOLO11 and YOLO26, model-capacity scaling from nano to extra-large, and the effect of conventional versus small-object-focused input-resolution training. Predictive gains were interpreted jointly with computational and training costs to identify configurations offering favorable accuracy--efficiency trade-offs for fine-grained orchard perception. Contingency routines for GPU-memory or AMP-related failures were included only as execution safeguards and were not considered part of the principal experimental protocol.}

\subsection{Performance Evaluation Across YOLO Generations and Model Scales}

\textcolor{black}{The performance of the 15 Ultralytics instance-segmentation architectures was evaluated using a common quantitative framework to enable direct comparison across YOLO generations and model capacities. The benchmark comprised the nano (n), small (s), medium (m), large (l), and extra-large (x) variants of YOLOv8, YOLO11, and YOLO26, each investigated under the conventional $640\times640$-pixel and small-object-focused $960\times960$-pixel training configurations, resulting in 30 model--resolution experiments. The dataset partitions were kept fixed across all experiments. Specifically, the 49-image validation set was used exclusively during model development for convergence monitoring, early stopping, and selection of the best-performing checkpoint, whereas all final quantitative results reported in Section~3 and Tables~3 onward were calculated on the independent held-out 48-image test set. This strict separation ensured that the comparative results represent performance on images not used for model optimization or checkpoint selection. The test set contained the three anatomical classes: \textit{calyx}, \textit{fruitlet}, and \textit{peduncle} which present substantially different perceptual challenges. Fruitlets generally occupy larger and more visually distinguishable regions, whereas calyxes and particularly peduncles are smaller, geometrically complex, frequently occluded, and characterized by weak foreground--background contrast within dense green canopy imagery. Performance was therefore evaluated both at the aggregate model level and independently for each anatomical class to characterize generation-, scale-, resolution-, and class-dependent behavior. Owing to the substantial computational requirements associated with training and evaluating the complete set of 30 model--resolution configurations, the benchmark was conducted using a single training run per configuration. Consequently, the reported results characterize the observed performance under the standardized experimental setting rather than across-run variability; repeated experiments with multiple random seeds would be valuable in future work for quantifying performance variance and statistical robustness.Precision and recall were calculated as}

\begin{equation}
\textcolor{black}{
P=\frac{TP}{TP+FP},
\qquad
R=\frac{TP}{TP+FN},
}
\end{equation}

\textcolor{black}{where $TP$, $FP$, and $FN$ denote true-positive, false-positive, and false-negative object predictions, respectively. Precision measures the reliability of positive predictions, whereas recall measures the proportion of ground-truth instances successfully recovered by a model. This distinction is particularly important for the present orchard environment because false positives may arise from visually similar leaves, shoots, stems, or other background structures, while false negatives are especially consequential for small or partially occluded calyx and peduncle instances.}

\textcolor{black}{The F1-score was used to summarize the balance between precision and recall and was computed as}

\begin{equation}
\textcolor{black}{
F1=2\frac{PR}{P+R}.
}
\end{equation}

\textcolor{black}{This metric provides a complementary measure of model behavior when gains in recall are accompanied by reduced precision or vice versa. Such trade-offs are relevant for robotic perception because a configuration producing high recall but excessive false detections may be unsuitable for downstream manipulation, whereas a highly conservative model may fail to identify actionable anatomical targets.}

\textcolor{black}{Spatial agreement between a predicted region $A$ and its corresponding ground-truth region $B$ was quantified using Intersection over Union (IoU):}

\begin{equation}
\textcolor{black}{
IoU(A,B)=
\frac{|A\cap B|}
{|A\cup B|}.
}
\end{equation}

\textcolor{black}{For bounding-box evaluation, $A$ and $B$ represent predicted and ground-truth boxes, whereas for instance segmentation they represent the corresponding predicted and reference masks. IoU is particularly informative for narrow anatomical structures such as peduncles because comparatively small boundary deviations can produce substantial changes in overlap. Object-level detection statistics and pixel-level mask overlap were treated separately to avoid conflating localization performance with segmentation-boundary quality.}

\textcolor{black}{Average Precision was calculated from the precision--recall relationship for each anatomical class, and mean Average Precision was obtained by averaging AP across the three target classes:}

\begin{equation}
\textcolor{black}{
mAP=\frac{1}{C}\sum_{c=1}^{C}AP_c,
\qquad C=3.
}
\end{equation}

\textcolor{black}{Performance was reported using both $mAP_{50}$ and $mAP_{50:95}$. The former evaluates predictions at an IoU threshold of 0.50, whereas $mAP_{50:95}$ averages AP over ten IoU thresholds from 0.50 to 0.95 at increments of 0.05:}

\begin{equation}
\textcolor{black}{
mAP_{50:95}
=
\frac{1}{10}
\sum_{\tau \in
\{0.50,0.55,\ldots,0.95\}}
mAP_{\tau}.
}
\end{equation}

\textcolor{black}{Accordingly, four principal accuracy measures were reported for each model: box $mAP_{50}$, box $mAP_{50:95}$, mask $mAP_{50}$, and mask $mAP_{50:95}$. The $mAP_{50}$ metric provides a comparatively permissive measure of successful localization, whereas $mAP_{50:95}$ imposes progressively stricter spatial-overlap requirements and therefore provides a more rigorous measure of bounding-box localization and mask quality. Class-wise AP and recall were additionally examined for calyx, fruitlet, and peduncle to determine whether architectural generation and model capacity affected the three anatomical structures differently.}

\textcolor{black}{Beyond predictive accuracy, model complexity and computational efficiency were quantified to support deployment-oriented comparison. Architectural complexity was characterized using the number of layers, trainable parameter count, and GFLOPs reported by the fused Ultralytics inference models. The total number of trainable parameters was represented as}

\begin{equation}
\textcolor{black}{
N_{\mathrm{param}}
=
\sum_{j=1}^{L}N_j,
}
\end{equation}

\textcolor{black}{where $N_j$ denotes the number of trainable weights and biases associated with parameterized layer $j$. Computational complexity was expressed in giga floating-point operations as}

\begin{equation}
\textcolor{black}{
GFLOPs=\frac{FLOPs}{10^{9}}.
}
\end{equation}

\textcolor{black}{These measures characterize intrinsic network capacity and computational demand and therefore facilitate comparison among the n, s, m, l, and x variants both within and across YOLO generations. Stored checkpoint size was additionally recorded because memory footprint represents an important practical consideration for deployment on robotic and edge-computing platforms.}

\textcolor{black}{Processing efficiency was evaluated using the preprocessing, inference, and postprocessing times reported during model validation. These quantities were treated separately because they represent distinct stages of the prediction pipeline: preprocessing includes image preparation before network execution, inference represents the forward computational pass through the model, and postprocessing includes the operations required to convert network outputs into final predictions. Recorded wall-clock training time, completed epoch count, early-stopping behavior, and best validation epoch were also analyzed to characterize optimization efficiency and convergence behavior across model generations and scales.}

\textcolor{black}{The resulting comparative analysis was organized along two principal dimensions. First, cross-generation evaluation compared YOLOv8, YOLO11, and YOLO26 to determine whether architectural progression produced consistent improvements in fine-grained orchard perception and computational efficiency. Second, scale-dependent evaluation compared the n, s, m, l, and x configurations within each generation to determine whether increasing model capacity produced proportional gains in detection and segmentation performance or whether performance saturation occurred at higher computational complexity. Particular emphasis was placed on calyx and peduncle performance because their relatively small spatial footprint and morphological complexity provide a stringent test of fine-grained perception capability.}

\textcolor{black}{SAHI was considered separately as an external deployment-oriented strategy for increasing the effective representation of small objects, but it was not incorporated into the quantitative training framework described above and did not alter the learned parameters, layer structure, GFLOPs, or checkpoint size of the evaluated YOLO models. Consequently, the principal comparative results are interpreted in terms of YOLO generation, model scale, predictive accuracy, convergence behavior, and computational efficiency, while slicing-based deployment is discussed as a complementary inference consideration rather than as an additional trainable architecture. This distinction preserves consistency between the implemented experiments and the conclusions drawn from the cross-generation benchmark.}

\section{Results and Discussion}

\begin{figure*}[h!]
    \centering
    \includegraphics[width=0.99\linewidth]{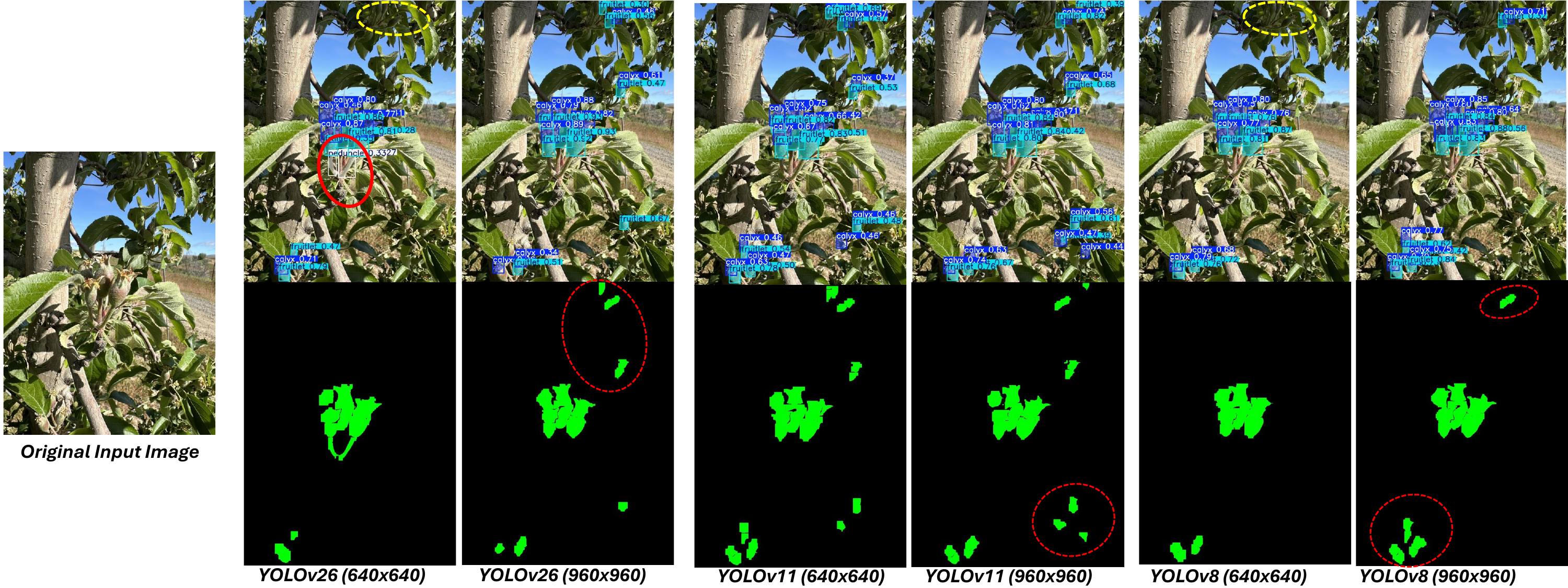}
    \caption{\textcolor{black}{\textbf{Qualitative comparison of cross-generation YOLO detection and instance segmentation under conventional and small-object-focused training configurations.} The leftmost panel shows the representative original orchard image, followed by prediction and corresponding heatmap visualizations obtained using YOLO26, YOLO11, and YOLOv8 at $640\times640$ and $960\times960$ input resolutions. The examples illustrate differences in the localization and segmentation of calyx, fruitlet, and peduncle structures under challenging green-on-green conditions. Overall, the $960\times960$ configurations recover additional small fruitlet instances in several regions, although model-specific differences remain evident, particularly for the substantially smaller and less represented peduncle class.}}
    
    \vspace{1mm}
    \begin{minipage}{0.98\linewidth}
    \footnotesize
    \textcolor{black}{\textit{Note:} Yellow dotted circles indicate representative fruitlets that were missed by the corresponding model. Red dotted circles highlight regions where the $960\times960$ small-object-focused configuration produced additional detections or segmentation masks relative to its $640\times640$ counterpart. The solid red circle identifies a particularly challenging peduncle that was successfully detected and segmented by YOLO26 at $640\times640$ but was not recovered by the other illustrated model--resolution configurations. Heatmaps provide complementary visualization of the spatial regions contributing to the corresponding model predictions.}
    \end{minipage}
    
    \label{fig:exampleImage}
\end{figure*}

\textcolor{black}{The qualitative results provide an initial visual assessment of how YOLO generation and input-resolution optimization influenced the detection and instance segmentation of fine-grained fruitlet anatomy before examining the quantitative performance metrics. Figure~\ref{fig:exampleImage} presents a representative orchard image in the leftmost panel, followed by the corresponding detection and segmentation outputs generated by YOLO26, YOLO11, and YOLOv8 under the $640\times640$ and small-object-focused $960\times960$ configurations, together with their associated heatmap visualizations. Across the examples, all three YOLO generations successfully identified substantial portions of the fruitlet clusters and produced instance masks for the calyx, fruitlet, and peduncle classes despite severe green-on-green similarity, dense clustering, partial occlusion, and considerable differences in anatomical scale. Nevertheless, the highlighted regions reveal important differences among model generations and training resolutions. The yellow dotted circles identify representative fruitlets that remained undetected or unsegmented in particular model outputs, illustrating that missed detections persisted when small fruitlets were partially occluded or visually embedded within surrounding foliage and neighboring fruit. In contrast, the red dotted circles in the corresponding visualization maps highlight regions where the $960\times960$ configuration produced additional object responses and segmentation masks compared with the $640\times640$ configuration. This behavior is observed at several locations in Fig.~\ref{fig:exampleImage}, qualitatively supporting the premise that increasing the training input resolution can preserve additional spatial information for small targets and improve the representation of fruitlets that may become less distinguishable at the lower resolution. The improvement, however, was not uniform across every anatomical structure or model generation. A particularly informative exception is highlighted by the solid red circle, where a very small peduncle was successfully detected and segmented by YOLO26 at $640\times640$ resolution, whereas the corresponding structure was not identified by the other illustrated model--resolution configurations. This observation is noteworthy because the peduncle represents the smallest and geometrically most difficult target class in the dataset, characterized by a narrow elongated structure, limited pixel footprint, and strong visual similarity to surrounding stems and vegetation. Moreover, peduncle annotations were less represented in the training dataset than the fruitlet and calyx classes, which may have contributed to greater variability in class-specific learning and detection performance. The ability of the $640\times640$ YOLO26 configuration to recover this particular peduncle nevertheless indicates that higher input resolution alone does not determine successful small-object perception; architectural generation, learned feature representation, object context, and model-specific optimization can also influence individual predictions. Taken collectively, however, the broader qualitative pattern in Fig.~\ref{fig:exampleImage}, particularly the additional responses highlighted by the red dotted circles and the missed fruitlets identified by the yellow dotted circles, suggests that the $960\times960$ small-object-focused configuration generally provides more favorable visual representation and segmentation of small fruitlets than the conventional $640\times640$ configuration. These qualitative observations motivate the subsequent quantitative analysis, in which the effects of YOLO generation, model scale, and input-resolution optimization are systematically examined using class-wise and aggregate detection and instance-segmentation metrics.}

\subsection{Cross-Generation Detection and Instance Segmentation Performance}
\subsubsection{YOLO27 Performance}
\label{subsubsec:yolo27_performance}

Experimental performance results for YOLO27 will be incorporated in the next manuscript version once the corresponding public trainable models become available.\footnote{This subsection will be updated with complete YOLO27 experimental results in the next version of the manuscript.}

\subsubsection{YOLO26 Performance}
\label{subsubsec:yolo26_performance}

The aggregate ``All''-class performance of the five YOLO26 instance-segmentation scales under the conventional 640$\times$640-pixel and small-object-focused 960$\times$960-pixel configurations is summarized in Table~\ref{tab:yolo26_overall_performance}. All models were evaluated using the same validation split, comprising 49 images and 694 annotated instances, thereby permitting direct comparison across model capacities and input resolutions. The results demonstrate that neither increasing model scale nor increasing input resolution produces a universally monotonic improvement in detection and segmentation performance. Instead, different configurations optimize different aspects of the precision--recall and localization--segmentation trade-off.

\begin{table*}[ht!]
\centering
\caption{\textbf{Aggregate detection and instance-segmentation performance of YOLO26 segmentation models under 640$\times$640 and 960$\times$960 input resolutions.}
Results correspond to the aggregate ``All'' category across calyx, fruitlet, and peduncle classes. Precision (P) and recall (R) correspond to bounding-box detection, while box and mask mAP are reported at IoU~=~0.50 and over IoU~=~0.50--0.95. Bold values indicate the highest value obtained for each metric across all evaluated YOLO26 configurations.}
\label{tab:yolo26_overall_performance}

\setlength{\tabcolsep}{4.6pt}
\renewcommand{\arraystretch}{1.15}
\small

\begin{tabular}{llcccccc}
\toprule
\textbf{Model} &
\textbf{Resolution} &
\textbf{P} &
\textbf{R} &
\textbf{Box mAP$_{50}$} &
\textbf{Box mAP$_{50:95}$} &
\textbf{Mask mAP$_{50}$} &
\textbf{Mask mAP$_{50:95}$} \\
\midrule

YOLO26n-seg & 640 & 0.568 & 0.704 & 0.661 & 0.363 & 0.644 & 0.335 \\
YOLO26n-seg & 960 & 0.661 & 0.697 & 0.682 & 0.414 & 0.676 & 0.384 \\
\midrule

YOLO26s-seg & 640 & 0.565 & 0.646 & 0.627 & 0.377 & 0.607 & 0.341 \\
YOLO26s-seg & 960 & 0.618 & 0.751 & 0.687 & 0.425 & \textbf{0.686} & \textbf{0.397} \\
\midrule

YOLO26m-seg & 640 & 0.648 & 0.641 & 0.654 & 0.403 & 0.629 & 0.361 \\
YOLO26m-seg & 960 & 0.588 & 0.728 & 0.665 & 0.419 & 0.665 & 0.395 \\
\midrule

YOLO26l-seg & 640 & 0.676 & 0.727 & \textbf{0.696} & 0.422 & 0.667 & 0.373 \\
YOLO26l-seg & 960 & 0.639 & 0.698 & 0.668 & \textbf{0.428} & 0.678 & 0.393 \\
\midrule

YOLO26x-seg & 640 & 0.578 & \textbf{0.765} & 0.669 & 0.405 & 0.675 & 0.385 \\
YOLO26x-seg & 960 & \textbf{0.689} & 0.631 & 0.660 & 0.414 & 0.658 & 0.382 \\

\bottomrule
\end{tabular}

\vspace{1mm}
\begin{minipage}{0.97\textwidth}
\footnotesize
\textit{Note:} Higher precision indicates a lower proportion of false-positive detections, whereas higher recall indicates that a larger proportion of annotated objects are recovered. Higher mAP values indicate stronger localization or segmentation performance. mAP$_{50:95}$ is the more stringent metric because prediction quality is averaged across IoU thresholds from 0.50 to 0.95.
\end{minipage}
\end{table*}

The results in Table~\ref{tab:yolo26_overall_performance} reveal that the highest value for a given metric does not consistently correspond to the largest network. The highest aggregate precision was obtained by YOLO26x-seg at 960 pixels ($P=0.689$), indicating that this configuration produced the strongest false-positive suppression among the evaluated models. In contrast, the highest aggregate recall was achieved by YOLO26x-seg at 640 pixels ($R=0.765$), meaning that this configuration recovered the largest fraction of annotated objects. These two results also illustrate why precision and recall should be interpreted jointly: the same extra-large architecture moves from a recall-dominant operating point at 640 pixels to a precision-dominant operating point at 960 pixels.

For bounding-box localization, YOLO26l-seg at 640 pixels achieved the highest box mAP$_{50}$ of 0.696, whereas YOLO26l-seg at 960 pixels achieved the highest box mAP$_{50:95}$ of 0.428. The difference between these two metrics is informative. mAP$_{50}$ represents successful localization at a relatively permissive overlap threshold and therefore reflects whether objects are detected in approximately the correct spatial region. In contrast, mAP$_{50:95}$ progressively penalizes localization errors as the required IoU becomes more stringent. Consequently, the slightly higher mAP$_{50:95}$ obtained at 960 pixels suggests that the higher-resolution large model more accurately localized object boundaries even though its mAP$_{50}$ was lower than the 640-pixel counterpart.

Instance-segmentation performance shows an especially important result. YOLO26s-seg at 960 pixels achieved both the highest mask mAP$_{50}$ (0.686) and the highest mask mAP$_{50:95}$ (0.397) among all ten configurations. The corresponding YOLO26m-960 and YOLO26l-960 configurations were very close, with mask mAP$_{50:95}$ values of 0.395 and 0.393, respectively. Therefore, approximately three- to six-fold increases in parameter count beyond YOLO26s did not translate into higher aggregate mask accuracy. This observation indicates that increasing input-level spatial representation can be more beneficial to fine-grained segmentation than simply increasing network capacity.

The effect of increasing resolution from 640 to 960 pixels is further quantified in Table~\ref{tab:yolo26_resolution_change}. Positive values represent improvement under the small-object-focused configuration, whereas negative values indicate reduced performance.

\begin{table*}[ht!]
\centering
\caption{\textbf{Absolute change in aggregate YOLO26 performance when increasing input resolution from 640$\times$640 to 960$\times$960 pixels.}
Changes were calculated as $\Delta=\mathrm{Metric}_{960}-\mathrm{Metric}_{640}$.}
\label{tab:yolo26_resolution_change}

\setlength{\tabcolsep}{5.0pt}
\renewcommand{\arraystretch}{1.15}
\small

\begin{tabular}{lrrrrrr}
\toprule
\textbf{Model} &
$\Delta$\textbf{P} &
$\Delta$\textbf{R} &
$\Delta$\textbf{Box mAP$_{50}$} &
$\Delta$\textbf{Box mAP$_{50:95}$} &
$\Delta$\textbf{Mask mAP$_{50}$} &
$\Delta$\textbf{Mask mAP$_{50:95}$} \\
\midrule

YOLO26n-seg & +0.093 & -0.007 & +0.021 & +0.051 & +0.032 & +0.049 \\
YOLO26s-seg & +0.053 & +0.105 & +0.060 & +0.048 & +0.079 & +0.056 \\
YOLO26m-seg & -0.060 & +0.087 & +0.011 & +0.016 & +0.036 & +0.034 \\
YOLO26l-seg & -0.037 & -0.029 & -0.028 & +0.006 & +0.011 & +0.020 \\
YOLO26x-seg & +0.111 & -0.134 & -0.009 & +0.009 & -0.017 & -0.003 \\

\bottomrule
\end{tabular}
\end{table*}

The paired comparison demonstrates that higher resolution does not improve all models uniformly. YOLO26s-seg exhibited the clearest and most consistent benefit: all six aggregate metrics increased at 960 pixels. Precision increased by 0.053, recall by 0.105, box mAP$_{50}$ by 0.060, box mAP$_{50:95}$ by 0.048, mask mAP$_{50}$ by 0.079, and mask mAP$_{50:95}$ by 0.056. This model therefore represents the strongest evidence that increasing spatial input resolution can simultaneously improve object recovery, localization, and pixel-level segmentation for the present fine-grained orchard-perception task.

YOLO26n-seg also benefited substantially from the 960-pixel configuration. Precision increased from 0.568 to 0.661, corresponding to an absolute gain of 0.093, while box mAP$_{50:95}$ improved by 0.051 and mask mAP$_{50:95}$ by 0.049. Recall remained essentially unchanged, decreasing only from 0.704 to 0.697. Thus, the primary effect of higher resolution for the nano model was improved localization and segmentation quality rather than increased object recovery.

YOLO26m-seg exhibited a different response. Recall increased substantially from 0.641 to 0.728 (+0.087), while box mAP$_{50}$, box mAP$_{50:95}$, mask mAP$_{50}$, and mask mAP$_{50:95}$ improved by 0.011, 0.016, 0.036, and 0.034, respectively. However, precision decreased by 0.060. This indicates that the 960-pixel medium model became more sensitive to object instances but accepted more false-positive detections, illustrating a recall-oriented shift in the operating point.

For YOLO26l-seg, the effect of resolution was mixed. Precision and recall decreased by 0.037 and 0.029, respectively, and box mAP$_{50}$ declined by 0.028. Nevertheless, stricter localization and segmentation metrics improved: box mAP$_{50:95}$ increased by 0.006, mask mAP$_{50}$ by 0.011, and mask mAP$_{50:95}$ by 0.020. Thus, the higher-resolution large model did not detect more objects overall, but the instances it successfully represented were segmented more accurately under stricter overlap requirements.

The extra-large configuration exhibited the strongest evidence of diminishing returns. Increasing resolution raised precision substantially from 0.578 to 0.689 (+0.111), but recall declined from 0.765 to 0.631 (-0.134). Box mAP$_{50}$ decreased slightly by 0.009 and mask mAP$_{50}$ by 0.017, while box mAP$_{50:95}$ increased by only 0.009 and mask mAP$_{50:95}$ decreased marginally by 0.003. Thus, higher resolution altered the precision--recall balance of YOLO26x rather than producing a meaningful net improvement in AP-based accuracy.

These findings also demonstrate that increasing model scale does not consistently improve accuracy. At 640 pixels, box mAP$_{50}$ increases from 0.661 for YOLO26n to 0.696 for YOLO26l but then decreases to 0.669 for YOLO26x. Mask mAP$_{50:95}$ similarly progresses from 0.335 to 0.341, 0.361, 0.373, and 0.385 from n through x, indicating a more gradual improvement in strict mask quality. However, the relationship changes substantially at 960 pixels: YOLO26s achieves the highest mask mAP$_{50:95}$, YOLO26l provides the highest box mAP$_{50:95}$, and YOLO26x provides the highest precision. Model scaling therefore improves representational capacity but does not produce proportional or universal gains in predictive accuracy.

The computational implications of these trends are summarized in Table~\ref{tab:yolo26_complexity_speed}. The number of layers, parameter count, and GFLOPs describe intrinsic architectural complexity and remain identical for corresponding 640- and 960-pixel variants because changing input resolution does not modify the underlying network architecture. In contrast, preprocessing, inference, and postprocessing times reflect the computational cost of processing each image and therefore vary with both model scale and input resolution.

\begin{table*}[ht!]
\centering
\caption{\textbf{Architectural complexity and validation-time processing speed of YOLO26 segmentation models under 640$\times$640 and 960$\times$960 input resolutions.}
Lower preprocessing, inference, and postprocessing times indicate greater computational efficiency, whereas higher parameter counts and GFLOPs indicate greater model capacity and computational demand.}
\label{tab:yolo26_complexity_speed}

\setlength{\tabcolsep}{4.2pt}
\renewcommand{\arraystretch}{1.15}
\small

\begin{tabular}{llrrrrrr}
\toprule
\textbf{Model} &
\textbf{Resolution} &
\textbf{Layers} &
\textbf{Parameters} &
\textbf{GFLOPs} &
\textbf{Preprocess} &
\textbf{Inference} &
\textbf{Postprocess} \\
& & & & & \textbf{(ms)} & \textbf{(ms)} & \textbf{(ms)} \\
\midrule

YOLO26n-seg & 640 & 139 & 2,689,469  & 9.0   & 0.1 & 1.9  & 0.3 \\
YOLO26n-seg & 960 & 139 & 2,689,469  & 9.0   & 0.2 & 1.8  & 0.6 \\
\midrule

YOLO26s-seg & 640 & 139 & 10,366,501 & 34.1  & 0.3 & 1.8  & 0.4 \\
YOLO26s-seg & 960 & 139 & 10,366,501 & 34.1  & 0.2 & 3.0  & 0.8 \\
\midrule

YOLO26m-seg & 640 & 149 & 23,509,781 & 121.2 & 0.1 & 3.0  & 0.4 \\
YOLO26m-seg & 960 & 149 & 23,509,781 & 121.2 & 0.2 & 5.9  & 0.8 \\
\midrule

YOLO26l-seg & 640 & 207 & 27,906,069 & 139.4 & 0.2 & 3.9  & 0.3 \\
YOLO26l-seg & 960 & 207 & 27,906,069 & 139.4 & 0.2 & 7.7  & 0.6 \\
\midrule

YOLO26x-seg & 640 & 207 & 62,730,197 & 313.0 & 0.1 & 6.3  & 0.9 \\
YOLO26x-seg & 960 & 207 & 62,730,197 & 313.0 & 0.2 & 12.8 & 0.7 \\

\bottomrule
\end{tabular}

\vspace{1mm}
\begin{minipage}{0.97\textwidth}
\footnotesize
\textit{Note:} Layers, parameters, and GFLOPs correspond to fused YOLO26 inference models. Higher values for these quantities represent greater architectural and computational complexity rather than inherently superior predictive performance. Processing times correspond to the validation output and are reported per image.
\end{minipage}
\end{table*}

The architectural progression from YOLO26n to YOLO26x results in a substantial increase in model complexity. YOLO26n contains only 2.69 million parameters and requires 9.0 GFLOPs, whereas YOLO26x contains 62.73 million parameters and requires 313.0 GFLOPs. The extra-large model therefore contains approximately 23.3 times more parameters and requires approximately 34.8 times more floating-point operations than the nano model. The large and extra-large models also contain 207 fused layers compared with 139 layers for the nano and small configurations. The highest parameter count, layer count, and GFLOPs therefore identify the computationally largest architecture, not necessarily the most accurate one. Conversely, the lowest values identify models with the smallest memory and computation requirements, which are desirable for resource-constrained robotic deployment.

Processing time exhibits a similarly strong scale dependence. At 640 pixels, inference latency ranged from 1.8--1.9~ms for the nano and small models to 3.0~ms for YOLO26m, 3.9~ms for YOLO26l, and 6.3~ms for YOLO26x. At 960 pixels, inference time increased to 3.0~ms for YOLO26s, 5.9~ms for YOLO26m, 7.7~ms for YOLO26l, and 12.8~ms for YOLO26x. The extra-large 960-pixel model therefore required more than four times the inference time of the small 960-pixel model while failing to exceed its mask mAP$_{50}$ or mask mAP$_{50:95}$. This difference is central to model selection for robotic perception, because additional network capacity is only advantageous when the corresponding accuracy improvement justifies its computational cost.

Preprocessing remained very small across all configurations, ranging from approximately 0.1 to 0.3~ms per image, while postprocessing ranged from approximately 0.3 to 0.9~ms. Consequently, the forward network pass constituted the dominant component of the reported processing latency. The lowest inference-time values therefore indicate the configurations most suitable for high-frequency perception, whereas the highest inference-time values represent the largest computational burden. Importantly, processing latency should be minimized, whereas precision, recall, and AP metrics should be maximized; therefore, the meaning of the ``best'' numerical value is direction-dependent across the two classes of metrics.

A particularly important result emerges when predictive accuracy and computational complexity are considered jointly. YOLO26s-seg at 960 pixels achieved the highest mask mAP$_{50}$ (0.686) and mask mAP$_{50:95}$ (0.397), together with box mAP$_{50}=0.687$, box mAP$_{50:95}=0.425$, and recall of 0.751. Yet this model contains only 10.37 million parameters, requires 34.1 GFLOPs, and exhibits a measured inference latency of 3.0~ms per image. In comparison, YOLO26x-seg at 960 pixels contains 62.73 million parameters, requires 313.0 GFLOPs, and exhibits 12.8~ms inference latency, while producing lower box mAP$_{50}$ (0.660), lower mask mAP$_{50}$ (0.658), and lower mask mAP$_{50:95}$ (0.382). Thus, increasing computational complexity by approximately six times in parameter count and more than nine times in GFLOPs did not improve aggregate segmentation accuracy.

From an overall performance perspective, the results therefore do not support designating the largest YOLO26 variant as universally superior. Instead, different configurations dominate different metrics. YOLO26l-640 provides the highest box mAP$_{50}$, YOLO26l-960 provides the highest box mAP$_{50:95}$, YOLO26x-640 provides the highest recall, YOLO26x-960 provides the highest precision, and YOLO26s-960 provides the strongest overall instance-segmentation performance. For a study focused jointly on object detection and anatomically fine-grained instance segmentation, YOLO26s-seg at 960 pixels therefore represents the most balanced YOLO26 configuration because it combines near-leading box localization with the highest mask accuracy while maintaining substantially lower computational complexity than the large and extra-large variants.

\begin{figure}[h!]
    \centering
    \includegraphics[width=0.99\linewidth]{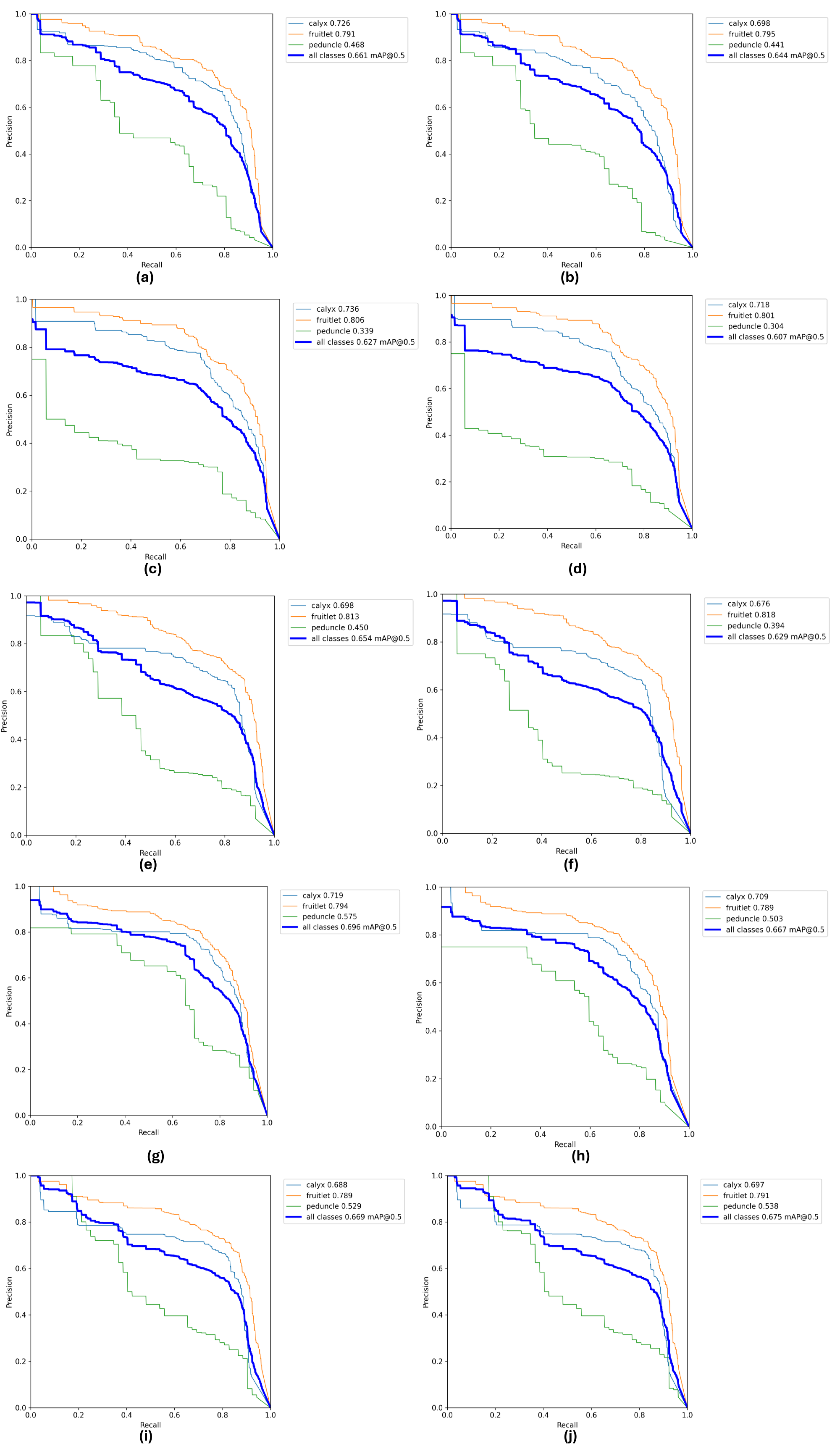}
  \caption{\textcolor{black}{Training--validation precision--recall (PR) curves for YOLO26 instance-segmentation models trained using the conventional $640\times640$-pixel input-resolution configuration: (a,b) YOLO26n-seg, (c,d) YOLO26s-seg, (e,f) YOLO26m-seg, (g,h) YOLO26l-seg, and (i,j) YOLO26x-seg. Panels (a,c,e,g,i) present bounding-box PR curves, whereas panels (b,d,f,h,j) present instance-mask PR curves.}}
    \label{fig:64026}
\end{figure}

\begin{figure}[h!]
    \centering
    \includegraphics[width=0.99\linewidth]{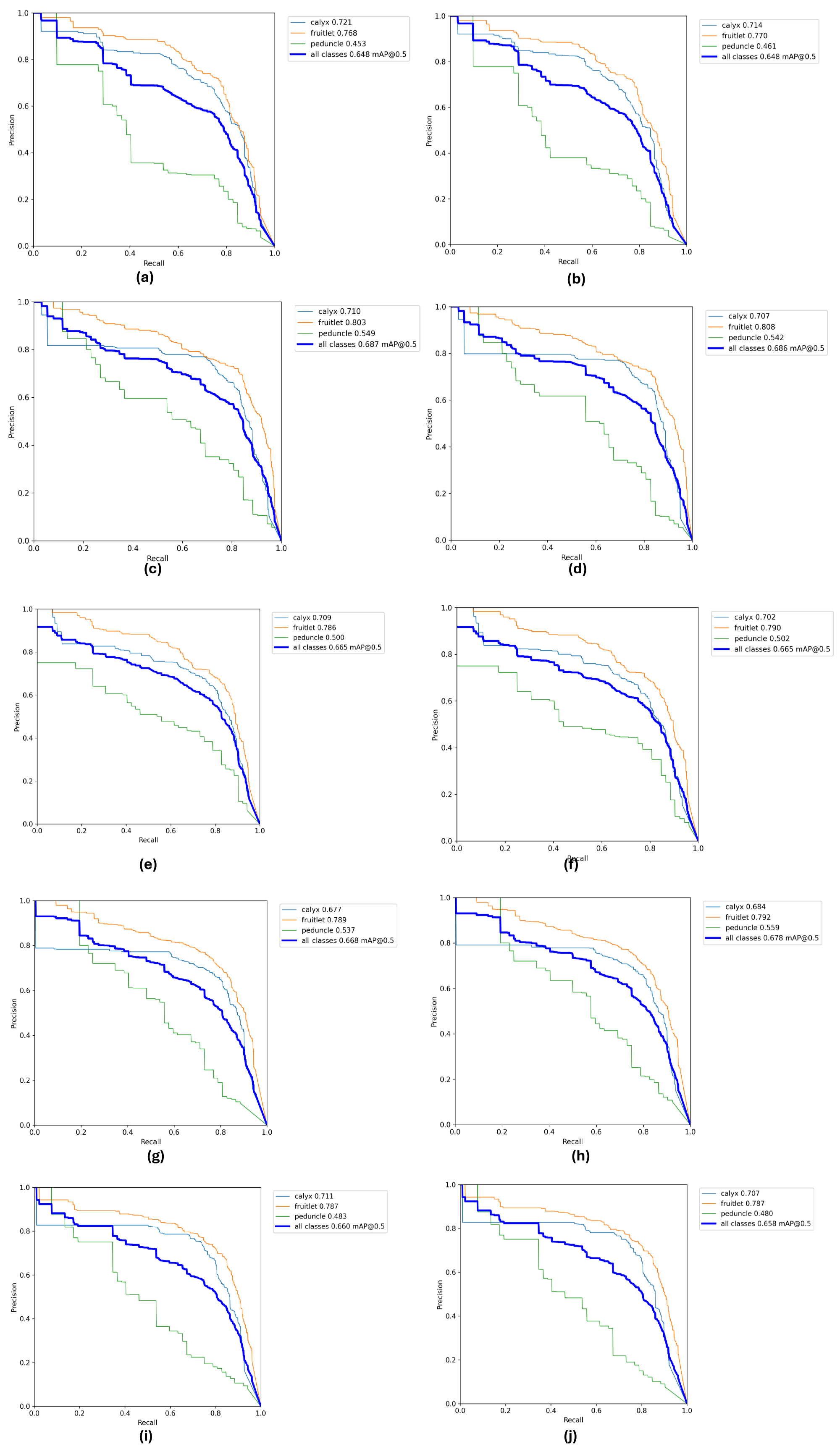}
\caption{\textcolor{black}{Training--validation precision--recall (PR) curves for YOLO26 instance-segmentation models trained using the small-object-focused $960\times960$-pixel input-resolution configuration: (a,b) YOLO26n-seg, (c,d) YOLO26s-seg, (e,f) YOLO26m-seg, (g,h) YOLO26l-seg, and (i,j) YOLO26x-seg. Panels (a,c,e,g,i) present bounding-box PR curves, whereas panels (b,d,f,h,j) present instance-mask PR curves.}}
    \label{fig:96026}
\end{figure}

\textcolor{black}{Collectively, these results establish three important observations regarding the interaction between model capacity and input-resolution optimization. First, predictive performance does not increase monotonically from the nano to extra-large variants, indicating that greater network capacity alone does not necessarily provide superior detection or instance-segmentation performance for fine-grained orchard structures. This behavior is also evident from the bounding-box and mask precision--recall (PR) characteristics of the five YOLO26 variants under the conventional $640\times640$ training configuration (Fig.~\ref{fig:64026}), where increasing model scale does not consistently produce corresponding improvements in the PR relationships. Second, increasing the input resolution from $640\times640$ to $960\times960$ produces the strongest and most consistent benefits for the nano, small, and medium configurations, whereas the gains become less uniform for the large and extra-large variants. Comparison of the PR curves obtained at $640\times640$ (Fig.~\ref{fig:64026}) with those obtained using the small-object-focused $960\times960$ configuration (Fig.~\ref{fig:96026}) further illustrates how increased spatial resolution influences both bounding-box localization and instance-mask prediction across model capacities. In particular, the $960\times960$ PR profiles in Fig.~\ref{fig:96026} complement the quantitative results by demonstrating that higher input resolution can provide favorable precision--recall behavior without requiring progression toward the highest-capacity models. Third, moderately sized architectures trained with greater spatial resolution can equal or outperform substantially larger networks for fine-grained instance segmentation, demonstrating that increasing network capacity and increasing spatial representation are not interchangeable optimization strategies. Taken together, the quantitative results and the PR-curve comparisons in Figs.~\ref{fig:64026} and~\ref{fig:96026} indicate that effective preservation of fine spatial information can be at least as important as increasing model capacity for the present green-on-green orchard perception problem. The subsequent class-wise analysis therefore examines whether these aggregate performance differences are concentrated within the most spatially challenging anatomical structures, particularly the calyx and peduncle.}

\subsubsection{YOLO11 Performance}
\label{subsubsec:yolo11_performance}

\textcolor{black}{The aggregate ``All''-class performance of the five YOLO11 segmentation scales under the conventional 640$\times$640-pixel and small-object-focused 960$\times$960-pixel configurations is summarized in Table~\ref{tab:yolo11_overall_performance}. All configurations were evaluated using the same validation split of 49 images containing 694 annotated instances. Similar to the behavior observed for YOLO26, the YOLO11 results demonstrate that increasing either network capacity or input resolution does not produce a uniformly monotonic response across precision, recall, and average-precision metrics. Instead, the strongest operating point depends on whether the objective emphasizes false-positive suppression, object recovery, strict spatial localization, or mask quality.}

\begin{table*}[ht!]
\centering
\caption{\textbf{\textcolor{black}{Aggregate detection and instance-segmentation performance of YOLO11 segmentation models under 640$\times$640 and 960$\times$960 input resolutions.}
Results correspond to the aggregate ``All'' category across calyx, fruitlet, and peduncle classes. Precision (P) and recall (R) correspond to bounding-box detection, while box and mask mAP are reported at IoU~=~0.50 and over IoU~=~0.50--0.95. Bold values indicate the highest value obtained for each metric across all YOLO11 configurations.}}
\label{tab:yolo11_overall_performance}

\setlength{\tabcolsep}{4.6pt}
\renewcommand{\arraystretch}{1.15}
\small

\begin{tabular}{llcccccc}
\toprule
\textbf{Model} &
\textbf{Resolution} &
\textbf{P} &
\textbf{R} &
\textbf{Box mAP$_{50}$} &
\textbf{Box mAP$_{50:95}$} &
\textbf{Mask mAP$_{50}$} &
\textbf{Mask mAP$_{50:95}$} \\
\midrule

YOLO11n-seg & 640 & 0.546 & 0.690 & 0.637 & 0.357 & 0.622 & 0.318 \\
YOLO11n-seg & 960 & 0.595 & 0.705 & 0.666 & 0.396 & 0.669 & 0.367 \\
\midrule

YOLO11s-seg & 640 & 0.553 & 0.668 & 0.623 & 0.364 & 0.621 & 0.334 \\
YOLO11s-seg & 960 & \textbf{0.668} & 0.641 & \textbf{0.694} & \textbf{0.426} & \textbf{0.692} & \textbf{0.402} \\
\midrule

YOLO11m-seg & 640 & 0.612 & 0.699 & 0.680 & 0.404 & 0.678 & 0.361 \\
YOLO11m-seg & 960 & 0.651 & 0.642 & 0.662 & 0.414 & 0.661 & 0.382 \\
\midrule

YOLO11l-seg & 640 & 0.637 & 0.679 & 0.645 & 0.389 & 0.643 & 0.343 \\
YOLO11l-seg & 960 & 0.558 & 0.756 & 0.652 & 0.414 & 0.654 & 0.384 \\
\midrule

YOLO11x-seg & 640 & 0.625 & 0.687 & 0.645 & 0.384 & 0.645 & 0.339 \\
YOLO11x-seg & 960 & 0.570 & \textbf{0.762} & 0.665 & 0.417 & 0.669 & 0.387 \\

\bottomrule
\end{tabular}

\vspace{1mm}
\begin{minipage}{0.97\textwidth}
\footnotesize
\textit{Note:} Higher precision indicates stronger suppression of false-positive detections, whereas higher recall indicates recovery of a larger proportion of annotated objects. Higher mAP values indicate stronger localization or segmentation performance. mAP$_{50:95}$ is the more stringent metric because prediction quality is averaged across IoU thresholds from 0.50 to 0.95.
\end{minipage}
\end{table*}

\textcolor{black}{A particularly notable result is that YOLO11s-seg at 960 pixels provides the strongest overall accuracy profile despite being only the second-smallest model in the YOLO11 family. It achieves the highest precision ($P=0.668$), box mAP$_{50}$ (0.694), box mAP$_{50:95}$ (0.426), mask mAP$_{50}$ (0.692), and mask mAP$_{50:95}$ (0.402). Thus, five of the six principal aggregate metrics are maximized by the small 960-pixel model. The only exception is recall, for which YOLO11x-seg at 960 pixels achieves the highest value ($R=0.762$). This separation is scientifically important because it demonstrates that the largest network primarily favors object recovery, whereas the smaller 960-pixel configuration provides a more favorable balance of localization and segmentation quality.}

\textcolor{black}{The highest box mAP$_{50:95}$ and mask mAP$_{50:95}$ values are especially informative because these metrics incorporate increasingly stringent overlap requirements. YOLO11s-960 therefore does not merely identify objects under a permissive localization criterion; it also provides the strongest aggregate spatial agreement with the ground-truth boxes and masks. In comparison, YOLO11x-960 reaches high recall but lower box mAP$_{50:95}$ (0.417) and mask mAP$_{50:95}$ (0.387). Consequently, increasing model capacity by more than six-fold in parameter count from the small to extra-large architecture does not improve the principal AP-based accuracy measures.}

\textcolor{black}{Table~\ref{tab:yolo11_resolution_change} quantifies the paired effect of increasing the input resolution from 640 to 960 pixels for each YOLO11 scale.}

\begin{table*}[ht!]
\centering
\caption{\textbf{Absolute change in aggregate YOLO11 performance when increasing input resolution from 640$\times$640 to 960$\times$960 pixels.}
Changes were calculated as $\Delta=\mathrm{Metric}_{960}-\mathrm{Metric}_{640}$. Positive values indicate improvement under the higher-resolution configuration.}
\label{tab:yolo11_resolution_change}

\setlength{\tabcolsep}{5.0pt}
\renewcommand{\arraystretch}{1.15}
\small

\begin{tabular}{lrrrrrr}
\toprule
\textbf{Model} &
$\Delta$\textbf{P} &
$\Delta$\textbf{R} &
$\Delta$\textbf{Box mAP$_{50}$} &
$\Delta$\textbf{Box mAP$_{50:95}$} &
$\Delta$\textbf{Mask mAP$_{50}$} &
$\Delta$\textbf{Mask mAP$_{50:95}$} \\
\midrule

YOLO11n-seg & +0.049 & +0.015 & +0.029 & +0.039 & +0.047 & +0.049 \\
YOLO11s-seg & +0.115 & -0.027 & +0.071 & +0.062 & +0.071 & +0.068 \\
YOLO11m-seg & +0.039 & -0.057 & -0.018 & +0.010 & -0.017 & +0.021 \\
YOLO11l-seg & -0.079 & +0.077 & +0.007 & +0.025 & +0.011 & +0.041 \\
YOLO11x-seg & -0.055 & +0.075 & +0.020 & +0.033 & +0.024 & +0.048 \\

\bottomrule
\end{tabular}
\end{table*}

\textcolor{black}{The nano configuration exhibits one of the most uniformly positive responses to increased spatial resolution. Moving from 640 to 960 pixels improves precision by 0.049, recall by 0.015, box mAP$_{50}$ by 0.029, box mAP$_{50:95}$ by 0.039, mask mAP$_{50}$ by 0.047, and mask mAP$_{50:95}$ by 0.049. Thus, every principal metric improves without an evident precision--recall trade-off. This result suggests that the additional image detail available at 960 pixels compensates effectively for the relatively limited representational capacity of the nano architecture.}

\textcolor{black}{YOLO11s shows an even larger AP-based improvement. Precision increases by 0.115, box mAP$_{50}$ by 0.071, box mAP$_{50:95}$ by 0.062, mask mAP$_{50}$ by 0.071, and mask mAP$_{50:95}$ by 0.068. Recall decreases modestly by 0.027, indicating that the higher-resolution model becomes more selective while substantially improving the spatial quality of its accepted predictions. The resulting configuration nevertheless provides the highest overall AP-based performance in the YOLO11 family.}

\textcolor{black}{The effect becomes more heterogeneous at medium and larger capacities. YOLO11m-960 improves strict box and mask quality by +0.010 and +0.021 in mAP$_{50:95}$, respectively, but box and mask mAP$_{50}$ decrease slightly and recall declines by 0.057. Conversely, YOLO11l-960 shifts strongly toward recall, increasing it by 0.077 while precision decreases by 0.079. Despite this change, box mAP$_{50:95}$ and mask mAP$_{50:95}$ still improve by 0.025 and 0.041. YOLO11x behaves similarly, with recall increasing by 0.075 and strict box and mask AP increasing by 0.033 and 0.048, respectively, while precision decreases by 0.055.}

\textcolor{black}{An important overall pattern therefore emerges: although the direction of precision and recall changes varies by model scale, \emph{mAP$_{50:95}$ improves at 960 pixels for every YOLO11 variant for both boxes and masks}. This consistency is particularly relevant for the present fine-grained segmentation problem because mAP$_{50:95}$ is more sensitive to spatial localization and mask-boundary quality than mAP$_{50}$. Higher-resolution training consequently provides its most systematic benefit in the stricter spatial-accuracy metrics.}

\textcolor{black}{Increasing model scale itself does not produce a monotonic improvement. At 640 pixels, box mAP$_{50}$ rises from 0.637 for YOLO11n to a maximum of 0.680 for YOLO11m but then decreases to 0.645 for both l and x. Mask mAP$_{50:95}$ similarly peaks at 0.361 for YOLO11m rather than the largest configurations. At 960 pixels, YOLO11s becomes the strongest AP-based model, while larger m, l, and x configurations do not exceed its localization or segmentation performance. Thus, increased model capacity alone is insufficient to guarantee improved accuracy on fine-grained green-on-green orchard targets.}

\textcolor{black}{The architectural complexity and measured processing characteristics are summarized in Table~\ref{tab:yolo11_complexity_speed}.}

\begin{table*}[ht!]
\centering
\caption{\textbf{Architectural complexity and validation-time processing speed of YOLO11 segmentation models under 640$\times$640 and 960$\times$960 input resolutions.}
Lower preprocessing, inference, and postprocessing times indicate greater computational efficiency, whereas larger parameter counts and GFLOPs indicate greater architectural capacity and computational demand.}
\label{tab:yolo11_complexity_speed}

\setlength{\tabcolsep}{4.2pt}
\renewcommand{\arraystretch}{1.15}
\small

\begin{tabular}{llrrrrrr}
\toprule
\textbf{Model} &
\textbf{Resolution} &
\textbf{Layers} &
\textbf{Parameters} &
\textbf{GFLOPs} &
\textbf{Preprocess} &
\textbf{Inference} &
\textbf{Postprocess} \\
& & & & & \textbf{(ms)} & \textbf{(ms)} & \textbf{(ms)} \\
\midrule

YOLO11n-seg & 640 & 114 & 2,835,153  & 9.6   & 0.8 & 1.0  & 2.4 \\
YOLO11n-seg & 960 & 114 & 2,835,153  & 9.6   & 0.5 & 2.8  & 1.9 \\
\midrule

YOLO11s-seg & 640 & 114 & 10,067,977 & 32.8  & 0.8 & 2.3  & 3.5 \\
YOLO11s-seg & 960 & 114 & 10,067,977 & 32.8  & 0.3 & 3.8  & 1.6 \\
\midrule

YOLO11m-seg & 640 & 139 & 22,337,625 & 112.9 & 0.8 & 3.8  & 2.4 \\
YOLO11m-seg & 960 & 139 & 22,337,625 & 112.9 & 0.3 & 5.6  & 0.9 \\
\midrule

YOLO11l-seg & 640 & 204 & 27,586,905 & 131.8 & 1.5 & 5.8  & 2.5 \\
YOLO11l-seg & 960 & 204 & 27,586,905 & 131.8 & 0.2 & 8.4  & 1.2 \\
\midrule

YOLO11x-seg & 640 & 204 & 62,005,593 & 295.9 & 1.9 & 6.6  & 2.9 \\
YOLO11x-seg & 960 & 204 & 62,005,593 & 295.9 & 0.3 & 12.9 & 1.9 \\

\bottomrule
\end{tabular}
\end{table*}

\textcolor{black}{Architectural complexity increases markedly with model scale. YOLO11n contains 2.84 million parameters and requires 9.6 GFLOPs, whereas YOLO11x contains 62.01 million parameters and requires 295.9 GFLOPs. The extra-large architecture therefore contains approximately 21.9 times more parameters and requires approximately 30.8 times more floating-point operations than the nano model. The fused layer count also increases from 114 for n and s to 139 for m and 204 for l and x. As with YOLO26, however, a higher layer count, parameter count, or GFLOP value represents greater representational and computational capacity rather than intrinsically superior accuracy.}

\textcolor{black}{Inference time increases with both model capacity and, generally, input resolution. At 640 pixels, inference latency ranges from 1.0~ms for YOLO11n to 2.3~ms for YOLO11s, 3.8~ms for YOLO11m, 5.8~ms for YOLO11l, and 6.6~ms for YOLO11x. At 960 pixels, the corresponding values increase to 2.8, 3.8, 5.6, 8.4, and 12.9~ms. The computational penalty of higher spatial resolution therefore becomes progressively more consequential for larger networks. In particular, YOLO11x-960 requires 12.9~ms per image for inference compared with only 3.8~ms for YOLO11s-960.}

\textcolor{black}{The accuracy--complexity relationship strongly favors YOLO11s-960. Despite using only 10.07 million parameters and 32.8 GFLOPs, it achieves the highest precision and all four AP metrics in the YOLO11 family, while requiring only 3.8~ms of measured inference time. YOLO11x-960, in contrast, requires 62.01 million parameters, 295.9 GFLOPs, and 12.9~ms inference time, yet provides lower box and mask AP values. Its principal advantage is higher recall. Consequently, YOLO11s-960 represents the strongest balanced YOLO11 configuration when both fine-grained segmentation accuracy and computational efficiency are considered.}

\begin{figure}[h!]
    \centering
    \includegraphics[width=0.99\linewidth]{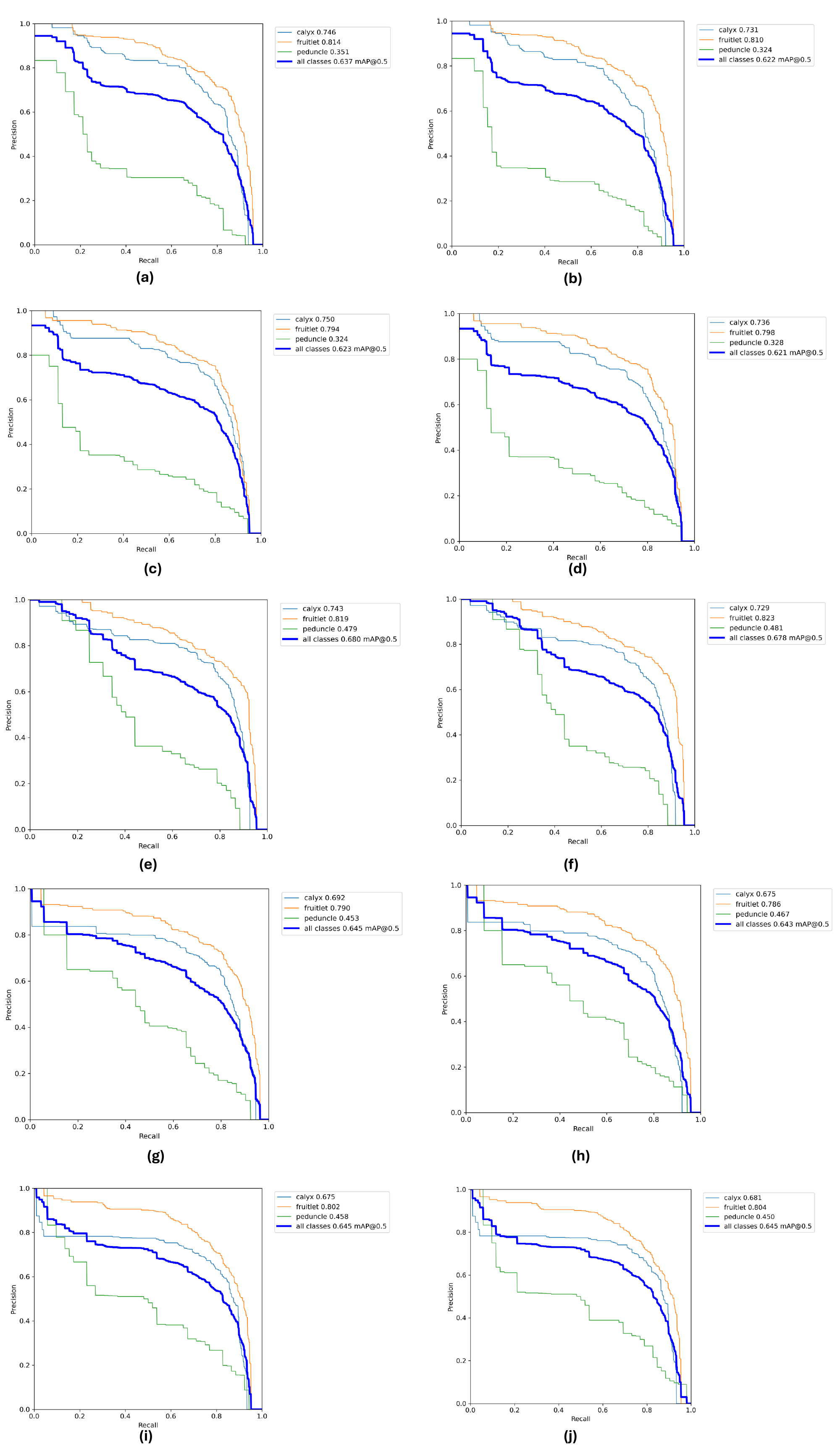}
\caption{\textcolor{black}{Training--validation precision--recall (PR) curves for YOLO11 instance-segmentation models trained using the conventional $640\times640$-pixel input-resolution configuration: (a,b) YOLO11n-seg, (c,d) YOLO11s-seg, (e,f) YOLO11m-seg, (g,h) YOLO11l-seg, and (i,j) YOLO11x-seg. Panels (a,c,e,g,i) present bounding-box PR curves, whereas panels (b,d,f,h,j) present instance-mask PR curves.}}
    \label{fig:11640}
\end{figure}

\begin{figure}[h!]
    \centering
    \includegraphics[width=0.99\linewidth]{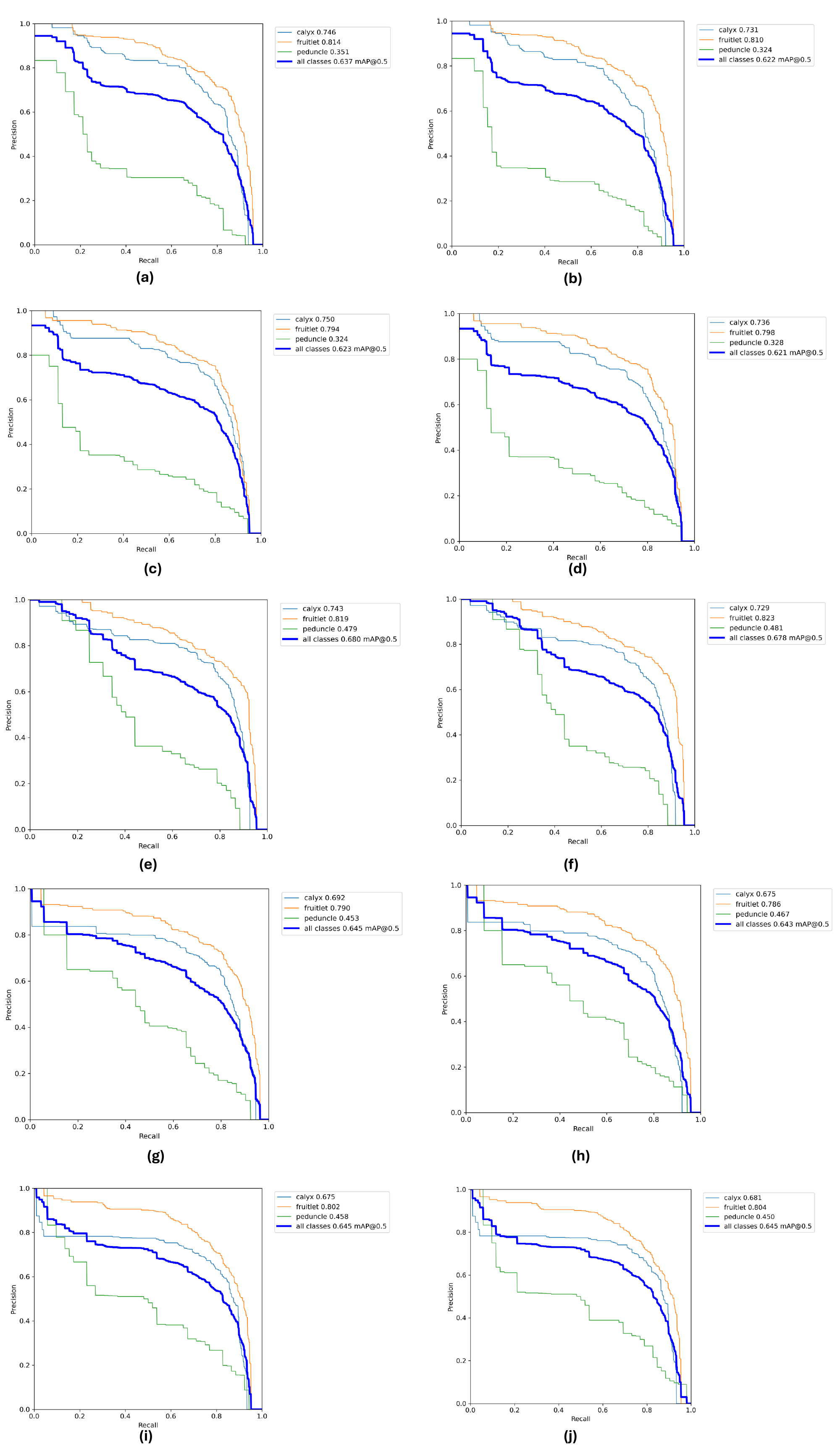}
\caption{\textcolor{black}{\textcolor{black}{Training--validation precision--recall (PR) curves for YOLO11 instance-segmentation models trained using the small-object-focused $960\times960$-pixel input-resolution configuration: (a,b) YOLO11n-seg, (c,d) YOLO11s-seg, (e,f) YOLO11m-seg, (g,h) YOLO11l-seg, and (i,j) YOLO11x-seg. Panels (a,c,e,g,i) present bounding-box PR curves, whereas panels (b,d,f,h,j) present instance-mask PR curves.}}}
    \label{fig:11960}
\end{figure}

\textcolor{black}{Collectively, the YOLO11 results reinforce three important observations regarding the effects of model capacity and input-resolution optimization. First, increasing model scale from nano to extra-large does not consistently translate into improved detection or instance-segmentation performance. This behavior is also reflected in the bounding-box and mask precision--recall (PR) curves obtained under the conventional $640\times640$-pixel configuration (Fig.~\ref{fig:11640}), where progression toward larger YOLO11 variants does not produce a uniformly superior precision--recall relationship. Second, increasing the input resolution to $960\times960$ pixels consistently improves the more stringent box and mask mAP$_{50:95}$ measures across all five YOLO11 scales, despite model-specific variations in individual precision and recall values. Comparison of the PR characteristics at $640\times640$ (Fig.~\ref{fig:11640}) and $960\times960$ (Fig.~\ref{fig:11960}) further illustrates how the small-object-focused configuration alters the balance between precision and recall across both bounding-box localization and instance-mask prediction. The PR curves in Fig.~\ref{fig:11960}, together with the corresponding AP-based metrics, indicate that the benefit of increased spatial resolution should therefore be interpreted through the overall precision--recall behavior rather than changes in either precision or recall alone. Third, the strongest AP-based YOLO11 configuration is achieved by the relatively compact small-scale model rather than the largest architecture, demonstrating that increased network capacity is not a prerequisite for improved fine-grained perception. Taken together, the quantitative metrics and PR-curve comparisons in Figs.~\ref{fig:11640} and~\ref{fig:11960} support the broader finding that preserving greater spatial information through higher-resolution training can be more advantageous for small-object detection and segmentation than simply increasing network capacity. This observation is particularly relevant to the present orchard dataset, where the smallest anatomical structures occupy limited image regions and can progressively lose discriminative information during feature extraction and spatial downsampling.}

\subsubsection{YOLOv8 Performance}
\label{subsubsec:yolov8_performance}

\textcolor{black}{The aggregate ``All''-class results for YOLOv8 are presented in Table~\ref{tab:yolov8_overall_performance}. As the earlier-generation reference architecture in the present benchmark, YOLOv8 provides an important basis for determining whether the effects of model scaling and increased input resolution observed for YOLO11 and YOLO26 are specific to recent architectures or represent a more general characteristic of fine-grained small-object segmentation. The results show that YOLOv8 remains competitive across several configurations, but, as with the newer generations, neither increasing network size nor increasing input resolution produces a universally monotonic improvement across all metrics.}

\begin{table*}[ht!]
\centering
\caption{\textbf{Aggregate detection and instance-segmentation performance of YOLOv8 segmentation models under 640$\times$640 and 960$\times$960 input resolutions.}
Results correspond to the aggregate ``All'' category across calyx, fruitlet, and peduncle classes. Precision (P) and recall (R) correspond to bounding-box detection. Bold values indicate the highest value obtained for each metric across all YOLOv8 configurations.}
\label{tab:yolov8_overall_performance}

\setlength{\tabcolsep}{4.6pt}
\renewcommand{\arraystretch}{1.15}
\small

\begin{tabular}{llcccccc}
\toprule
\textbf{Model} &
\textbf{Resolution} &
\textbf{P} &
\textbf{R} &
\textbf{Box mAP$_{50}$} &
\textbf{Box mAP$_{50:95}$} &
\textbf{Mask mAP$_{50}$} &
\textbf{Mask mAP$_{50:95}$} \\
\midrule

YOLOv8n-seg & 640 & 0.628 & 0.730 & 0.660 & 0.356 & 0.649 & 0.325 \\
YOLOv8n-seg & 960 & 0.619 & 0.628 & 0.646 & 0.390 & 0.652 & 0.371 \\
\midrule

YOLOv8s-seg & 640 & \textbf{0.656} & 0.659 & 0.666 & 0.390 & 0.648 & 0.356 \\
YOLOv8s-seg & 960 & 0.605 & \textbf{0.774} & \textbf{0.692} & \textbf{0.419} & \textbf{0.690} & \textbf{0.390} \\
\midrule

YOLOv8m-seg & 640 & 0.628 & 0.735 & 0.626 & 0.376 & 0.609 & 0.337 \\
YOLOv8m-seg & 960 & 0.640 & 0.636 & 0.660 & 0.395 & 0.654 & 0.371 \\
\midrule

YOLOv8l-seg & 640 & 0.495 & 0.758 & 0.636 & 0.360 & 0.632 & 0.327 \\
YOLOv8l-seg & 960 & 0.595 & 0.697 & 0.669 & 0.394 & 0.671 & 0.377 \\
\midrule

YOLOv8x-seg & 640 & 0.549 & 0.660 & 0.624 & 0.365 & 0.619 & 0.335 \\
YOLOv8x-seg & 960 & 0.588 & 0.719 & 0.647 & 0.404 & 0.651 & 0.380 \\

\bottomrule
\end{tabular}
\end{table*}

\textcolor{black}{The strongest aggregate YOLOv8 configuration is again the small model at 960 pixels. YOLOv8s-960 achieves the highest recall ($R=0.774$), box mAP$_{50}$ (0.692), box mAP$_{50:95}$ (0.419), mask mAP$_{50}$ (0.690), and mask mAP$_{50:95}$ (0.390). The only metric not maximized by this configuration is precision, for which YOLOv8s at 640 pixels achieves $P=0.656$. Thus, as observed for YOLO11 and YOLO26, a relatively lightweight small-scale model provides the strongest overall detection--segmentation operating point when trained at higher spatial resolution.}

\textcolor{black}{The distinction between YOLOv8s-640 and YOLOv8s-960 illustrates a pronounced precision--recall shift. Precision decreases from 0.656 to 0.605, but recall increases from 0.659 to 0.774. More importantly, every AP-based metric improves, including box mAP$_{50:95}$ from 0.390 to 0.419 and mask mAP$_{50:95}$ from 0.356 to 0.390. The higher-resolution model therefore recovers substantially more objects while simultaneously improving aggregate localization and mask quality despite accepting additional false positives.}

\textcolor{black}{The paired resolution effects across all model scales are quantified in Table~\ref{tab:yolov8_resolution_change}.}

\begin{table*}[ht!]
\centering
\caption{\textbf{Absolute change in aggregate YOLOv8 performance when increasing input resolution from 640$\times$640 to 960$\times$960 pixels.}
Changes were calculated as $\Delta=\mathrm{Metric}_{960}-\mathrm{Metric}_{640}$.}
\label{tab:yolov8_resolution_change}

\setlength{\tabcolsep}{5.0pt}
\renewcommand{\arraystretch}{1.15}
\small

\begin{tabular}{lrrrrrr}
\toprule
\textbf{Model} &
$\Delta$\textbf{P} &
$\Delta$\textbf{R} &
$\Delta$\textbf{Box mAP$_{50}$} &
$\Delta$\textbf{Box mAP$_{50:95}$} &
$\Delta$\textbf{Mask mAP$_{50}$} &
$\Delta$\textbf{Mask mAP$_{50:95}$} \\
\midrule

YOLOv8n-seg & -0.009 & -0.102 & -0.014 & +0.034 & +0.003 & +0.046 \\
YOLOv8s-seg & -0.051 & +0.115 & +0.026 & +0.029 & +0.042 & +0.034 \\
YOLOv8m-seg & +0.012 & -0.099 & +0.034 & +0.019 & +0.045 & +0.034 \\
YOLOv8l-seg & +0.100 & -0.061 & +0.033 & +0.034 & +0.039 & +0.050 \\
YOLOv8x-seg & +0.039 & +0.059 & +0.023 & +0.039 & +0.032 & +0.045 \\

\bottomrule
\end{tabular}
\end{table*}

\textcolor{black}{The most striking result in Table~\ref{tab:yolov8_resolution_change} is the consistency of the stringent AP metrics. Although precision and recall fluctuate substantially among model capacities, \emph{box mAP$_{50:95}$ and mask mAP$_{50:95}$ improve for every YOLOv8 scale at 960 pixels}. The gains in box mAP$_{50:95}$ range from +0.019 for YOLOv8m to +0.039 for YOLOv8x, while mask mAP$_{50:95}$ improves by +0.034 to +0.050 across the family. Thus, higher input resolution systematically improves spatial agreement under stricter overlap thresholds even when the number or confidence of recovered objects changes.}

\textcolor{black}{YOLOv8n demonstrates this distinction particularly clearly. Increasing resolution decreases precision by 0.009, recall by 0.102, and box mAP$_{50}$ by 0.014, yet box mAP$_{50:95}$ improves by 0.034 and mask mAP$_{50:95}$ by 0.046. This indicates that the higher-resolution nano model produces fewer successful detections overall but provides more spatially accurate localization and mask delineation for the predictions it retains. The effect therefore cannot be adequately characterized by recall or mAP$_{50}$ alone.}

\textcolor{black}{YOLOv8s exhibits a different response, with recall increasing substantially by 0.115 and all four AP measures improving. Its mask mAP$_{50}$ increases by 0.042 and mask mAP$_{50:95}$ by 0.034, establishing YOLOv8s-960 as the strongest overall configuration. YOLOv8m similarly gains in all AP measures despite a 0.099 reduction in recall. YOLOv8l gains 0.100 in precision and records the largest mask mAP$_{50:95}$ improvement (+0.050), although recall falls by 0.061. YOLOv8x provides the most uniformly positive response among the larger networks, with precision (+0.039), recall (+0.059), box mAP$_{50}$ (+0.023), box mAP$_{50:95}$ (+0.039), mask mAP$_{50}$ (+0.032), and mask mAP$_{50:95}$ (+0.045) all improving.}

\textcolor{black}{Model scaling again fails to produce monotonic accuracy gains. At 640 pixels, box mAP$_{50}$ is highest for YOLOv8s (0.666), while the medium, large, and extra-large configurations produce lower values of 0.626, 0.636, and 0.624, respectively. Strict mask accuracy follows a similar pattern, with YOLOv8s-640 achieving mask mAP$_{50:95}=0.356$, exceeding the larger m, l, and x variants. At 960 pixels, the small model remains strongest, with mask mAP$_{50:95}=0.390$. Consequently, the large increases in model capacity associated with the m, l, and x variants do not translate into proportional improvements for the present fine-grained perception task.}

\textcolor{black}{Table~\ref{tab:yolov8_complexity_speed} summarizes the architectural and processing characteristics underlying these accuracy differences.}

\begin{table*}[ht!]
\centering
\caption{\textbf{Architectural complexity and validation-time processing speed of YOLOv8 segmentation models under 640$\times$640 and 960$\times$960 input resolutions.}
Lower processing times represent greater computational efficiency, whereas higher parameter counts and GFLOPs represent greater model capacity and computational demand.}
\label{tab:yolov8_complexity_speed}

\setlength{\tabcolsep}{4.2pt}
\renewcommand{\arraystretch}{1.15}
\small

\begin{tabular}{llrrrrrr}
\toprule
\textbf{Model} &
\textbf{Resolution} &
\textbf{Layers} &
\textbf{Parameters} &
\textbf{GFLOPs} &
\textbf{Preprocess} &
\textbf{Inference} &
\textbf{Postprocess} \\
& & & & & \textbf{(ms)} & \textbf{(ms)} & \textbf{(ms)} \\
\midrule

YOLOv8n-seg & 640 & 86  & 3,258,649  & 11.3  & 0.4 & 1.2  & 2.2 \\
YOLOv8n-seg & 960 & 86  & 3,258,649  & 11.3  & 0.2 & 1.7  & 2.3 \\
\midrule

YOLOv8s-seg & 640 & 86  & 11,780,761 & 39.9  & 0.9 & 2.8  & 0.9 \\
YOLOv8s-seg & 960 & 86  & 11,780,761 & 39.9  & 0.4 & 2.4  & 0.9 \\
\midrule

YOLOv8m-seg & 640 & 106 & 27,224,121 & 104.3 & 1.4 & 5.4  & 1.2 \\
YOLOv8m-seg & 960 & 106 & 27,224,121 & 104.3 & 0.4 & 6.2  & 1.5 \\
\midrule

YOLOv8l-seg & 640 & 126 & 45,914,201 & 210.1 & 1.2 & 4.5  & 1.2 \\
YOLOv8l-seg & 960 & 126 & 45,914,201 & 210.1 & 0.2 & 9.2  & 2.4 \\
\midrule

YOLOv8x-seg & 640 & 126 & 71,723,545 & 328.0 & 1.4 & 5.9  & 2.0 \\
YOLOv8x-seg & 960 & 126 & 71,723,545 & 328.0 & 0.2 & 11.6 & 1.0 \\

\bottomrule
\end{tabular}

\vspace{1mm}
\begin{minipage}{0.97\textwidth}
\footnotesize
\textit{Note:} Layers, parameters, and GFLOPs correspond to fused YOLOv8 inference models. These architectural quantities are unchanged between corresponding 640- and 960-pixel configurations because increasing input resolution alters the spatial dimensions processed by the network rather than its learned architecture. Processing times correspond to the per-image values reported during validation.
\end{minipage}
\end{table*}

\textcolor{black}{YOLOv8 exhibits a particularly wide computational range. YOLOv8n contains 3.26 million parameters and requires 11.3 GFLOPs, whereas YOLOv8x contains 71.72 million parameters and requires 328.0 GFLOPs. The extra-large architecture therefore contains approximately 22 times more parameters and requires approximately 29 times more floating-point operations than the nano model. The number of fused layers increases from 86 for the n and s configurations to 106 for m and 126 for l and x.}

\textcolor{black}{The processing-time results broadly reflect this increase in complexity, although individual measurements show some non-monotonic variation attributable to validation-time execution characteristics. At 640 pixels, measured inference times are 1.2, 2.8, 5.4, 4.5, and 5.9~ms for the n, s, m, l, and x models, respectively. At 960 pixels, the corresponding values are 1.7, 2.4, 6.2, 9.2, and 11.6~ms. The higher-resolution large and extra-large configurations therefore incur particularly substantial computational penalties.}

\textcolor{black}{An interesting exception is YOLOv8s, for which the recorded 960-pixel inference time (2.4~ms) is slightly lower than the 640-pixel value (2.8~ms). Because input resolution normally increases theoretical computational workload while the network architecture remains unchanged, this isolated timing difference should be interpreted as a validation-runtime measurement effect rather than evidence that 960-pixel inference is intrinsically less computationally demanding. The broader trend across the m, l, and x variants shows the expected increase in inference latency at higher resolution.}

\begin{figure}[h!]
    \centering
    \includegraphics[width=0.99\linewidth]{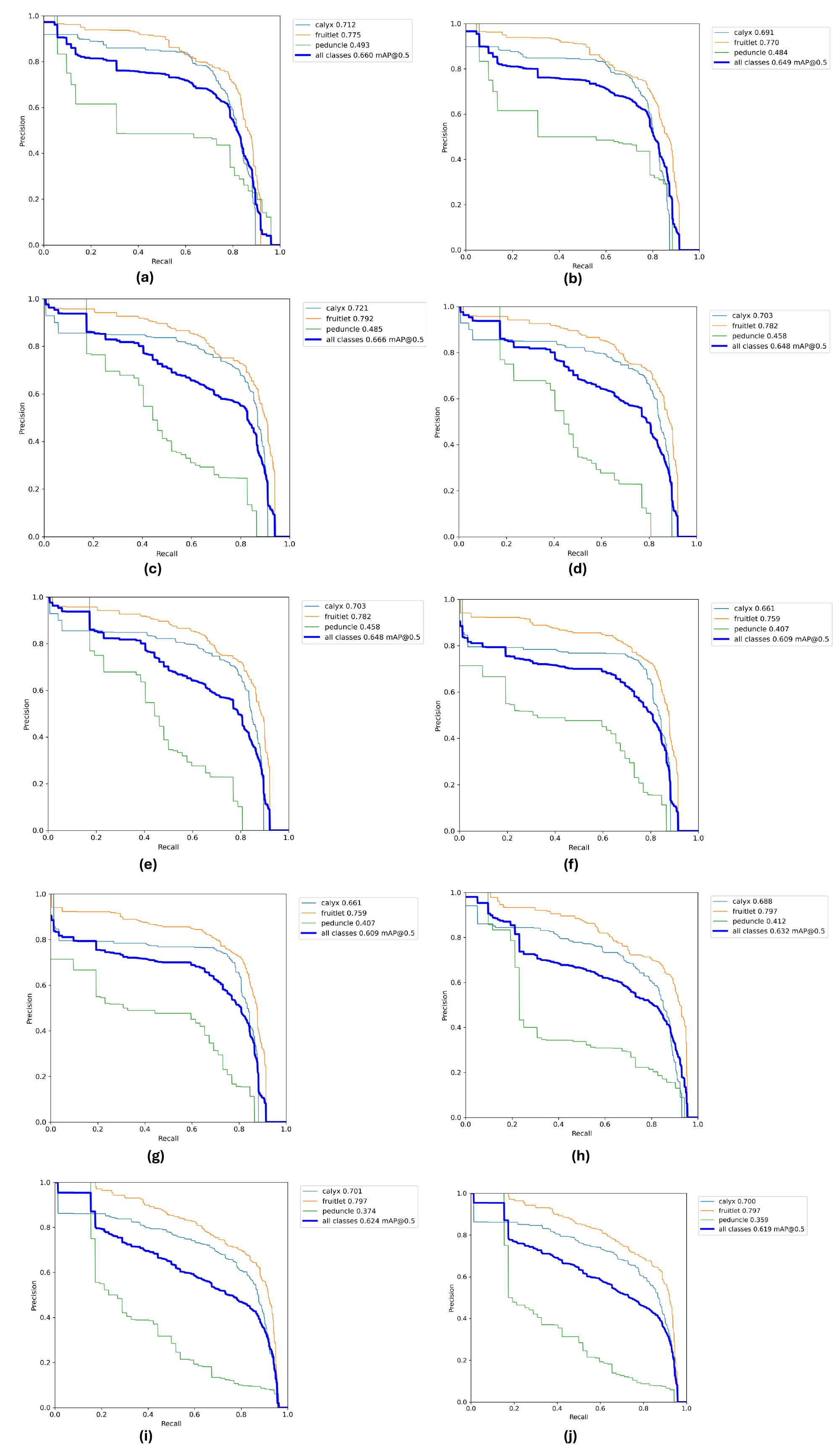}
\caption{\textcolor{black}{Training--validation precision--recall (PR) curves for YOLOv8 instance-segmentation models trained using the conventional $640\times640$-pixel input-resolution configuration: (a,b) YOLOv8n-seg, (c,d) YOLOv8s-seg, (e,f) YOLOv8m-seg, (g,h) YOLOv8l-seg, and (i,j) YOLOv8x-seg. Panels (a,c,e,g,i) present bounding-box PR curves, whereas panels (b,d,f,h,j) present instance-mask PR curves.}}
    \label{fig:8640}
\end{figure}

\begin{figure}[h!]
    \centering
    \includegraphics[width=0.99\linewidth]{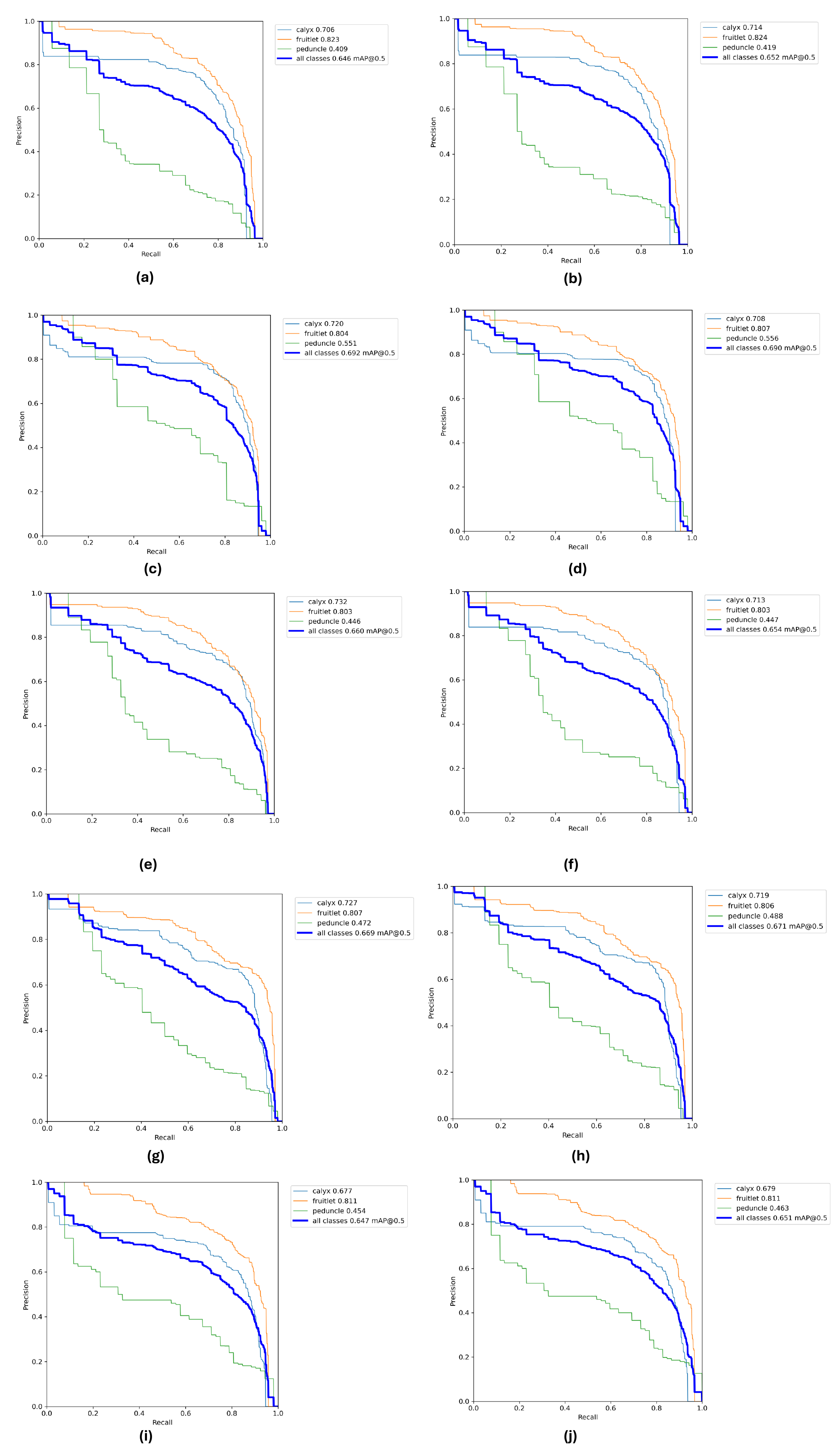}
\caption{\textcolor{black}{Training--validation precision--recall (PR) curves for YOLOv8 instance-segmentation models trained using the small-object-focused $960\times960$-pixel input-resolution configuration: (a,b) YOLOv8n-seg, (c,d) YOLOv8s-seg, (e,f) YOLOv8m-seg, (g,h) YOLOv8l-seg, and (i,j) YOLOv8x-seg. Panels (a,c,e,g,i) present bounding-box PR curves, whereas panels (b,d,f,h,j) present instance-mask PR curves.}}
    \label{fig:8960}
\end{figure}

\textcolor{black}{The relationship between predictive accuracy and computational complexity identifies YOLOv8s-960 as the most favorable overall operating point within the YOLOv8 family. This configuration achieved the highest recall and the strongest performance across all four AP-based measures while requiring only 11.78 million parameters and 39.9~GFLOPs, with a measured inference latency of 2.4~ms. In comparison, YOLOv8x-960 required 71.72 million parameters, 328.0~GFLOPs, and 11.6~ms of inference time, yet produced lower box mAP$_{50}$ (0.647 versus 0.692), box mAP$_{50:95}$ (0.404 versus 0.419), mask mAP$_{50}$ (0.651 versus 0.690), and mask mAP$_{50:95}$ (0.380 versus 0.390). Thus, increasing the model from the small to the extra-large scale resulted in an approximately six-fold increase in parameter count and more than an eight-fold increase in GFLOPs without a corresponding improvement in detection or instance-segmentation accuracy. These results demonstrate that increased network capacity alone is not sufficient to improve fine-grained perception when the principal challenge is the limited spatial representation of small anatomical structures.}

\textcolor{black}{The precision--recall characteristics shown in Figs.~\ref{fig:8640} and~\ref{fig:8960} provide additional insight into this behavior across the complete YOLOv8 model family. Figure~\ref{fig:8640} presents the bounding-box and instance-mask PR curves for the five YOLOv8 scales trained using the conventional $640\times640$-pixel configuration, whereas Fig.~\ref{fig:8960} shows the corresponding curves obtained using the small-object-focused $960\times960$-pixel configuration. These curves complement the aggregate performance metrics by illustrating the precision--recall behavior across confidence thresholds for both localization and mask prediction. In particular, the results show that increasing model capacity from nano through extra-large does not produce a uniform outward shift of the PR curves, confirming that larger architectures do not systematically dominate their smaller counterparts. Instead, the favorable PR characteristics of the small and intermediate configurations indicate that appropriately preserving spatial information can be more consequential than simply increasing network depth and capacity for the present fine-grained orchard perception task.}

\textcolor{black}{Comparison of Figs.~\ref{fig:8640} and~\ref{fig:8960} further supports the quantitative evidence that the higher-resolution training strategy improves the ability of YOLOv8 to represent difficult small-object instances. Although the magnitude of improvement varies with model scale and class, the $960\times960$-pixel configuration improves the stringent box and mask mAP$_{50:95}$ measures for every YOLOv8 scale. This consistency is particularly important because mAP$_{50:95}$ evaluates predictions over progressively stricter spatial-overlap thresholds and is therefore more sensitive to localization and mask-boundary quality than mAP$_{50}$. The combined numerical results and PR-curve behavior consequently suggest that the principal benefit of the 960-pixel configuration is not simply an increase in detections, but improved localization and segmentation quality across a broader range of operating thresholds.}

\textcolor{black}{Taken together, the YOLOv8 experiments provide an important earlier-generation confirmation of the trends observed for YOLO11 and YOLO26. Increasing model size does not consistently improve fine-grained orchard perception, whereas increasing the training input resolution from $640\times640$ to $960\times960$ pixels more reliably improves stringent localization and mask-quality measures. Most notably, box and mask mAP$_{50:95}$ increase under the 960-pixel configuration for every YOLOv8 model scale, while YOLOv8s-960 emerges as the strongest overall accuracy--efficiency configuration despite being substantially smaller than the large and extra-large variants. The quantitative results, together with the PR-curve comparisons in Figs.~\ref{fig:8640} and~\ref{fig:8960}, therefore indicate that preserving fine spatial information during training provides greater practical benefit for this dataset than maximizing network capacity alone. This recurring pattern across YOLOv8, YOLO11, and YOLO26 motivates the subsequent cross-generation synthesis, where the effects of architectural generation, model scale, and input-resolution-based small-object optimization are evaluated jointly rather than interpreting each YOLO family in isolation.}

\subsection{Class-Wise Performance for Calyx, Fruitlet, and Peduncle}
\label{subsec:classwise_performance}

Aggregate performance provides an overall measure of model capability, but it does not fully characterize the difficulty of the individual anatomical targets considered in this study. The three classes differ substantially in size, morphology, visibility, and foreground--background separability. The fruitlet represents the comparatively largest target and typically occupies a sufficiently large image region to preserve recognizable shape and texture after resizing. The calyx is substantially smaller, irregular in geometry, orientation dependent, and frequently partially occluded by the fruit body or nearby canopy structures. The peduncle represents the most challenging target because it is extremely thin and elongated, frequently occupies only a small number of pixels, and can be visually similar to surrounding stems, petioles, shoots, and woody structures.

These anatomical differences are particularly relevant to robotic thinning because whole-fruit localization alone does not fully describe the local attachment geometry required for selective physical interaction. Accurate delineation of the calyx and especially the peduncle provides more detailed structural information that can potentially support target prioritization, attachment-point reasoning, manipulation planning, and collision-aware end-effector placement. The following class-wise analyses therefore examine not only absolute performance but also whether increasing input resolution from 640$\times$640 to 960$\times$960 pixels preferentially improves the smaller anatomical targets.

\subsubsection{YOLO26 Performance}
\label{subsubsec:yolo26_classwise}
\begin{figure}[ht!]
     \centering
     \includegraphics[width=\linewidth]{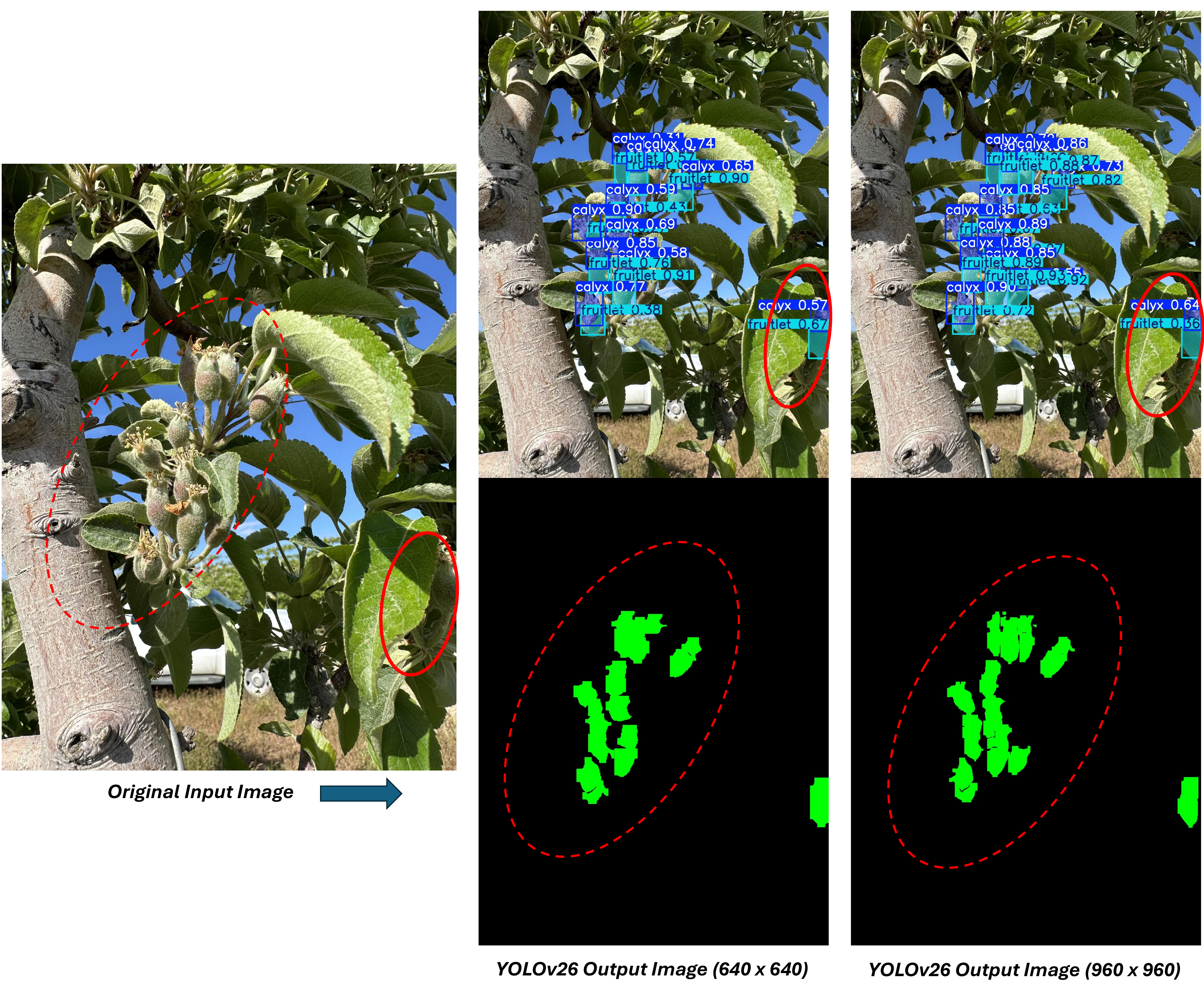}
\caption{\textcolor{black}{Qualitative comparison of YOLO26 instance-segmentation results at $640\times640$ and $960\times960$ training resolutions. Red dotted circles highlight additional calyx and fruitlet instances recovered by the higher-resolution configuration, whereas the peduncle remains undetected at both resolutions. Corresponding heatmaps illustrate the enhanced spatial response and anatomical-feature representation achieved with $960\times960$ training.}}
    \label{fig:yolo26imagediscussion}
\end{figure}

\textcolor{black}{The class-wise YOLO26 results reveal a clear hierarchy of perception difficulty, with fruitlet exhibiting the strongest performance, followed by calyx, while peduncle remained the most challenging anatomical class. As qualitatively illustrated in Fig.~\ref{fig:yolo26imagediscussion}, the $960\times960$ configuration detected and segmented additional fruitlet and calyx instances that were missed at $640\times640$, particularly within the regions highlighted by red dotted circles.}

\textcolor{black}{Fig.~\ref{fig:yolo26imagediscussion} also demonstrates an important case in the lower-right region, highlighted by the solid red circle, where an extremely small and visually indistinct fruitlet was successfully detected and segmented by YOLO26 at both resolutions, including its calyx and fruit body. However, neither configuration successfully identified the associated peduncle, despite recovering the other anatomical components of the same fruitlet. This persistent difficulty is consistent with the substantial class imbalance in the dataset, which contains 3,706 calyx and 3,714 fruitlet instances but only 839 peduncle annotations. These qualitative observations suggest that increasing training resolution improves the representation and recovery of small calyx and fruitlet structures, whereas peduncle perception remains constrained by its extremely small spatial footprint, thin geometry, frequent occlusion, and comparatively limited training representation. Consequently, increasing the number and diversity of manually annotated peduncle instances, particularly for small, partially occluded, and low-contrast examples, represents an important direction for further improving anatomical-level segmentation performance.}

\paragraph{Fruitlet performance.}

\textcolor{black}{Table~\ref{tab:yolo26_fruitlet_class} summarizes YOLO26 performance for the fruitlet class. Across most configurations, fruitlet box mAP$_{50}$ approaches or exceeds 0.79, and mask mAP$_{50}$ similarly remains close to or above 0.79. These results indicate that the fruit body retains sufficient spatial representation even under conventional resizing.}

\begin{table*}[ht!]
\centering
\caption{\textbf{Class-wise detection and instance-segmentation performance for the fruitlet class using YOLO26 models under 640$\times$640 and 960$\times$960 input resolutions.} Bold values indicate the highest value for each metric.}
\label{tab:yolo26_fruitlet_class}
\setlength{\tabcolsep}{3.8pt}
\renewcommand{\arraystretch}{1.14}
\small
\begin{tabular}{llcccccccc}
\toprule
\textbf{Model} & \textbf{Resolution} &
\multicolumn{4}{c}{\textbf{Bounding Box}} &
\multicolumn{4}{c}{\textbf{Mask}}\\
\cmidrule(lr){3-6}\cmidrule(lr){7-10}
& & \textbf{P} & \textbf{R} & \textbf{mAP$_{50}$} & \textbf{mAP$_{50:95}$}
& \textbf{P} & \textbf{R} & \textbf{mAP$_{50}$} & \textbf{mAP$_{50:95}$}\\
\midrule
YOLO26n-seg & 640 & 0.643 & 0.841 & 0.791 & 0.479 & 0.643 & 0.841 & 0.795 & 0.476\\
YOLO26n-seg & 960 & \textbf{0.755} & 0.742 & 0.803 & 0.541 & \textbf{0.754} & 0.759 & 0.811 & 0.519\\
\midrule
YOLO26s-seg & 640 & 0.703 & 0.801 & 0.806 & 0.530 & 0.694 & 0.802 & 0.801 & 0.509\\
YOLO26s-seg & 960 & 0.715 & 0.826 & 0.803 & 0.554 & 0.738 & 0.777 & 0.808 & \textbf{0.534}\\
\midrule
YOLO26m-seg & 640 & 0.727 & 0.796 & \textbf{0.813} & \textbf{0.558} & 0.730 & 0.799 & \textbf{0.818} & 0.525\\
YOLO26m-seg & 960 & 0.673 & 0.820 & 0.786 & 0.553 & 0.652 & 0.829 & 0.790 & 0.533\\
\midrule
YOLO26l-seg & 640 & 0.709 & 0.799 & 0.794 & 0.539 & 0.713 & 0.796 & 0.789 & 0.503\\
YOLO26l-seg & 960 & 0.679 & 0.817 & 0.789 & 0.548 & 0.681 & 0.820 & 0.792 & 0.521\\
\midrule
YOLO26x-seg & 640 & 0.683 & \textbf{0.845} & 0.789 & 0.535 & 0.685 & \textbf{0.848} & 0.791 & 0.502\\
YOLO26x-seg & 960 & 0.737 & 0.779 & 0.787 & 0.542 & 0.737 & 0.779 & 0.787 & 0.519\\
\bottomrule
\end{tabular}
\end{table*}

\textcolor{black}{The fruitlet results reveal clear performance saturation. YOLO26m-640 provides the highest box mAP$_{50}$ (0.813), box mAP$_{50:95}$ (0.558), and mask mAP$_{50}$ (0.818), while YOLO26s-960 provides the highest mask mAP$_{50:95}$ (0.534). Increasing resolution therefore does not systematically improve fruitlet performance. For example, YOLO26n benefits considerably in precision and strict AP, but its recall decreases. Similarly, YOLO26m and YOLO26x show only marginal or mixed changes. These results indicate that the fruitlet body is sufficiently large for conventional 640-pixel training to preserve most of the spatial information required for reliable localization and segmentation.}

\paragraph{Calyx performance.}

\textcolor{black}{The calyx presents a more difficult perception problem because of its smaller spatial footprint, irregular morphology, variable orientation, and frequent occlusion. Table~\ref{tab:yolo26_calyx_class} shows that 960-pixel training improves several strict localization and segmentation measures, although gains are not uniform across model scales.}

\begin{table*}[ht!]
\centering
\caption{\textbf{Class-wise detection and instance-segmentation performance for the calyx class using YOLO26 models under 640$\times$640 and 960$\times$960 input resolutions.} Bold values indicate the highest value for each metric.}
\label{tab:yolo26_calyx_class}
\setlength{\tabcolsep}{3.8pt}
\renewcommand{\arraystretch}{1.14}
\small
\begin{tabular}{llcccccccc}
\toprule
\textbf{Model} & \textbf{Resolution} &
\multicolumn{4}{c}{\textbf{Bounding Box}} &
\multicolumn{4}{c}{\textbf{Mask}}\\
\cmidrule(lr){3-6}\cmidrule(lr){7-10}
& & \textbf{P} & \textbf{R} & \textbf{mAP$_{50}$} & \textbf{mAP$_{50:95}$}
& \textbf{P} & \textbf{R} & \textbf{mAP$_{50}$} & \textbf{mAP$_{50:95}$}\\
\midrule
YOLO26n-seg & 640 & 0.625 & \textbf{0.813} & 0.726 & 0.360 & 0.601 & 0.781 & 0.698 & 0.328\\
YOLO26n-seg & 960 & \textbf{0.746} & 0.771 & \textbf{0.756} & 0.397 & 0.717 & 0.755 & \textbf{0.737} & 0.371\\
\midrule
YOLO26s-seg & 640 & 0.618 & 0.771 & 0.736 & 0.401 & 0.600 & 0.755 & 0.718 & 0.352\\
YOLO26s-seg & 960 & 0.664 & 0.792 & 0.710 & 0.397 & 0.704 & 0.757 & 0.707 & 0.366\\
\midrule
YOLO26m-seg & 640 & 0.672 & 0.742 & 0.698 & 0.375 & 0.666 & 0.736 & 0.676 & 0.337\\
YOLO26m-seg & 960 & 0.622 & 0.803 & 0.709 & \textbf{0.409} & 0.590 & 0.803 & 0.702 & \textbf{0.380}\\
\midrule
YOLO26l-seg & 640 & 0.693 & 0.771 & 0.719 & 0.396 & 0.692 & 0.758 & 0.709 & 0.343\\
YOLO26l-seg & 960 & 0.635 & 0.809 & 0.677 & 0.379 & 0.638 & \textbf{0.812} & 0.684 & 0.349\\
\midrule
YOLO26x-seg & 640 & 0.664 & 0.798 & 0.688 & 0.358 & 0.674 & 0.811 & 0.697 & 0.340\\
YOLO26x-seg & 960 & 0.718 & 0.752 & 0.711 & 0.401 & \textbf{0.718} & 0.752 & 0.707 & 0.368\\
\bottomrule
\end{tabular}
\end{table*}

\textcolor{black}{The clearest benefit occurs for YOLO26n, where higher resolution increases box mAP$_{50}$ from 0.726 to 0.756 and mask mAP$_{50:95}$ from 0.328 to 0.371. YOLO26m shows a different but equally important response: box recall increases from 0.742 to 0.803, mask recall from 0.736 to 0.803, box mAP$_{50:95}$ from 0.375 to 0.409, and mask mAP$_{50:95}$ from 0.337 to 0.380. These gains indicate that additional spatial resolution can improve both calyx recovery and high-IoU boundary quality. However, the response becomes mixed for larger configurations, particularly YOLO26l and YOLO26x, confirming that input resolution and model capacity interact rather than contributing independently.}

\paragraph{Peduncle performance.}

\textcolor{black}{Peduncle is the most challenging YOLO26 class. Its thin and elongated structure, limited pixel coverage, partial visibility, and similarity to neighboring stems and woody structures reduce both foreground discrimination and boundary stability. Table~\ref{tab:yolo26_peduncle_class} shows that several of the largest resolution-related improvements occur for this class.}

\begin{table*}[ht!]
\centering
\caption{\textbf{Class-wise detection and instance-segmentation performance for the peduncle class using YOLO26 models under 640$\times$640 and 960$\times$960 input resolutions.} Bold values indicate the highest value for each metric.}
\label{tab:yolo26_peduncle_class}
\setlength{\tabcolsep}{3.8pt}
\renewcommand{\arraystretch}{1.14}
\small
\begin{tabular}{llcccccccc}
\toprule
\textbf{Model} & \textbf{Resolution} &
\multicolumn{4}{c}{\textbf{Bounding Box}} &
\multicolumn{4}{c}{\textbf{Mask}}\\
\cmidrule(lr){3-6}\cmidrule(lr){7-10}
& & \textbf{P} & \textbf{R} & \textbf{mAP$_{50}$} & \textbf{mAP$_{50:95}$}
& \textbf{P} & \textbf{R} & \textbf{mAP$_{50}$} & \textbf{mAP$_{50:95}$}\\
\midrule
YOLO26n-seg & 640 & 0.434 & 0.457 & 0.468 & 0.250 & 0.415 & 0.438 & 0.441 & 0.201\\
YOLO26n-seg & 960 & 0.483 & 0.577 & 0.487 & 0.303 & 0.478 & 0.577 & 0.480 & 0.263\\
\midrule
YOLO26s-seg & 640 & 0.375 & 0.365 & 0.339 & 0.200 & 0.333 & 0.337 & 0.304 & 0.163\\
YOLO26s-seg & 960 & 0.477 & 0.635 & 0.549 & 0.325 & 0.586 & 0.558 & 0.542 & 0.290\\
\midrule
YOLO26m-seg & 640 & 0.545 & 0.385 & 0.450 & 0.278 & 0.490 & 0.346 & 0.394 & 0.220\\
YOLO26m-seg & 960 & 0.469 & 0.561 & 0.500 & 0.294 & 0.441 & \textbf{0.673} & 0.502 & 0.271\\
\midrule
YOLO26l-seg & 640 & \textbf{0.626} & 0.611 & \textbf{0.575} & 0.330 & 0.588 & 0.538 & 0.503 & 0.273\\
YOLO26l-seg & 960 & 0.604 & 0.469 & 0.537 & \textbf{0.358} & \textbf{0.629} & 0.488 & \textbf{0.559} & 0.308\\
\midrule
YOLO26x-seg & 640 & 0.387 & \textbf{0.654} & 0.529 & 0.322 & 0.387 & 0.654 & 0.538 & \textbf{0.313}\\
YOLO26x-seg & 960 & 0.610 & 0.362 & 0.483 & 0.299 & 0.610 & 0.362 & 0.480 & 0.257\\
\bottomrule
\end{tabular}
\end{table*}

\textcolor{black}{The response of YOLO26s is particularly notable. Increasing input resolution raises peduncle box recall from 0.365 to 0.635, box mAP$_{50}$ from 0.339 to 0.549, and box mAP$_{50:95}$ from 0.200 to 0.325. Mask mAP$_{50}$ increases from 0.304 to 0.542 and mask mAP$_{50:95}$ from 0.163 to 0.290. YOLO26m also exhibits a substantial increase in mask recall, from 0.346 to 0.673, indicating that higher resolution strongly improves sensitivity to thin attachment structures in this medium-capacity model.}

\textcolor{black}{The effect is less consistent for the largest networks. YOLO26l-960 improves strict box AP and mask AP but loses recall, whereas YOLO26x-960 experiences a pronounced precision--recall reversal: precision increases while recall decreases from 0.654 to 0.362. Consequently, higher resolution is most beneficial for the nano, small, and medium configurations, while very large models exhibit diminishing or mixed returns.}

\textcolor{black}{To summarize the anatomical effect, Table~\ref{tab:yolo26_class_resolution_summary} reports the mean 960-minus-640 change across all five YOLO26 scales.}

\begin{table*}[ht!]
\centering
\caption{\textbf{Mean absolute change in class-wise YOLO26 performance when increasing input resolution from 640$\times$640 to 960$\times$960 pixels across all five model scales.}}
\label{tab:yolo26_class_resolution_summary}
\setlength{\tabcolsep}{4pt}
\renewcommand{\arraystretch}{1.15}
\small
\begin{tabular}{lrrrrrrrr}
\toprule
\textbf{Class} &
$\Delta$\textbf{Box P} & $\Delta$\textbf{Box R} &
$\Delta$\textbf{Box mAP$_{50}$} & $\Delta$\textbf{Box mAP$_{50:95}$} &
$\Delta$\textbf{Mask P} & $\Delta$\textbf{Mask R} &
$\Delta$\textbf{Mask mAP$_{50}$} & $\Delta$\textbf{Mask mAP$_{50:95}$}\\
\midrule
Fruitlet & +0.019 & -0.020 & -0.005 & +0.019 & +0.019 & -0.024 & -0.001 & +0.022\\
Calyx & +0.023 & +0.006 & -0.001 & +0.019 & +0.027 & +0.008 & +0.008 & +0.027\\
Peduncle & \textbf{+0.055} & \textbf{+0.026} & \textbf{+0.039} & \textbf{+0.040} &
\textbf{+0.106} & \textbf{+0.069} & \textbf{+0.077} & \textbf{+0.044}\\
\bottomrule
\end{tabular}
\end{table*}

\textcolor{black}{The averaged differences provide clear evidence of an object-scale-dependent resolution effect. Fruitlet shows essentially no mean change in mask mAP$_{50}$ ($-0.001$), whereas peduncle improves by 0.077. Mean peduncle mask precision and recall increase by 0.106 and 0.069, respectively, and strict mask mAP$_{50:95}$ improves by 0.044. Calyx exhibits an intermediate response. Thus, for YOLO26, the principal advantage of the 960-pixel configuration is not a uniform improvement across all anatomical classes but a disproportionate gain for the structure most vulnerable to spatial-information loss.}

\subsubsection{YOLO11 Performance}
\label{subsubsec:yolo11_classwise}

\begin{figure}[ht!]
     \centering
     \includegraphics[width=\linewidth]{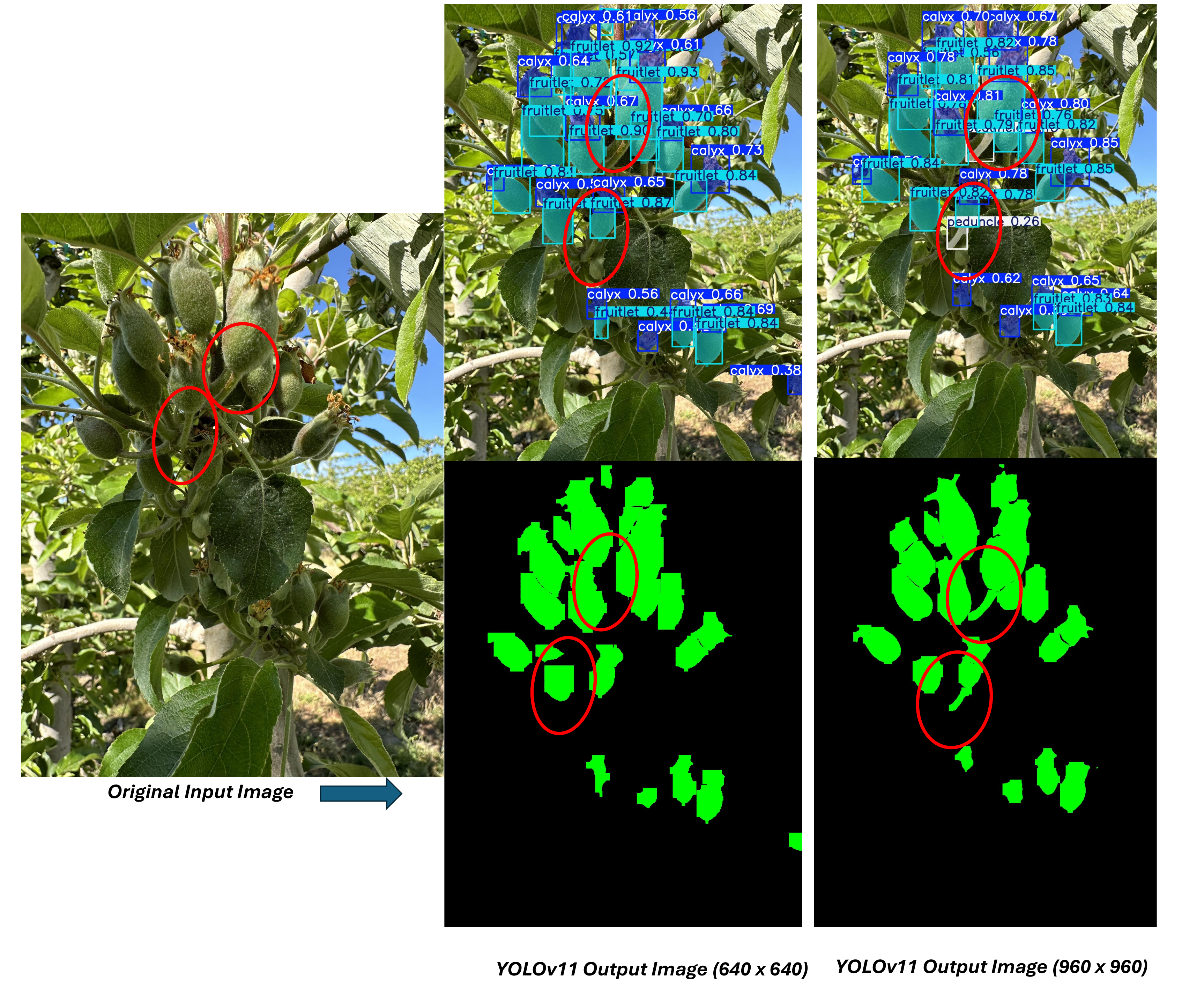}
\caption{\textcolor{black}{Qualitative comparison of YOLO11 segmentation at $640\times640$ and $960\times960$ training resolutions. Solid red circles highlight peduncles successfully detected and segmented only by the $960\times960$ configuration, demonstrating improved fine-scale anatomical perception while both configurations effectively identify fruitlet and calyx structures.}}
    \label{fig:yolo11imagediscussion}
\end{figure}

\textcolor{black}{The class-wise YOLO11 results exhibit a similar anatomical hierarchy, with fruitlet and calyx demonstrating comparatively strong detection and segmentation performance, while peduncle remains the most challenging target because of its substantially smaller spatial footprint, elongated geometry, and limited visual contrast. Importantly, the qualitative comparison in Fig.~\ref{fig:yolo11imagediscussion} demonstrates that increasing the training resolution from $640\times640$ to $960\times960$ substantially improved the representation of this difficult anatomical structure.}

\textcolor{black}{Within the densely clustered early-stage fruitlets, the two regions indicated by solid red circles in Fig.~\ref{fig:yolo11imagediscussion} show peduncles that were successfully detected and segmented by the YOLO11 model trained at $960\times960$, whereas the corresponding $640\times640$ model failed to identify any peduncle in the same image. Both resolution configurations nevertheless remained effective in recognizing the comparatively larger fruitlet bodies and associated calyx structures. These observations indicate that the principal qualitative benefit of the higher-resolution YOLO11 configuration is concentrated on the smallest anatomical target: preserving greater spatial detail during training enabled recovery of peduncle structures that were otherwise lost at the conventional resolution. The visual evidence therefore supports the class-wise quantitative analysis showing that small-object-focused $960\times960$ training can provide particularly meaningful improvements for fine-scale peduncle perception rather than uniformly benefiting all anatomical classes.}

\paragraph{Fruitlet performance.}

\begin{table*}[ht!]
\centering
\caption{\textbf{Class-wise detection and instance-segmentation performance for the fruitlet class using YOLO11 models under 640$\times$640 and 960$\times$960 input resolutions.}}
\label{tab:yolo11_fruitlet_class}
\setlength{\tabcolsep}{3.8pt}
\renewcommand{\arraystretch}{1.14}
\small
\begin{tabular}{llcccccccc}
\toprule
\textbf{Model} & \textbf{Resolution} &
\multicolumn{4}{c}{\textbf{Bounding Box}} &
\multicolumn{4}{c}{\textbf{Mask}}\\
\cmidrule(lr){3-6}\cmidrule(lr){7-10}
& & P & R & mAP$_{50}$ & mAP$_{50:95}$ & P & R & mAP$_{50}$ & mAP$_{50:95}$\\
\midrule
YOLO11n-seg & 640 & 0.677 & 0.851 & 0.814 & 0.512 & 0.674 & 0.848 & 0.810 & 0.485\\
YOLO11n-seg & 960 & 0.686 & 0.841 & \textbf{0.835} & 0.545 & 0.685 & 0.841 & \textbf{0.834} & 0.523\\
YOLO11s-seg & 640 & 0.670 & 0.829 & 0.794 & 0.519 & 0.668 & 0.826 & 0.798 & 0.491\\
YOLO11s-seg & 960 & \textbf{0.753} & 0.780 & 0.817 & 0.562 & \textbf{0.753} & 0.780 & 0.820 & 0.540\\
YOLO11m-seg & 640 & 0.629 & 0.890 & 0.819 & 0.538 & 0.667 & 0.884 & 0.823 & 0.496\\
YOLO11m-seg & 960 & 0.719 & 0.796 & 0.802 & 0.550 & 0.722 & 0.798 & 0.804 & 0.528\\
YOLO11l-seg & 640 & 0.708 & 0.811 & 0.790 & 0.519 & 0.708 & 0.811 & 0.786 & 0.485\\
YOLO11l-seg & 960 & 0.621 & 0.884 & 0.812 & 0.551 & 0.628 & 0.893 & 0.815 & 0.525\\
YOLO11x-seg & 640 & 0.716 & 0.790 & 0.802 & 0.522 & 0.719 & 0.793 & 0.804 & 0.491\\
YOLO11x-seg & 960 & 0.595 & \textbf{0.927} & 0.822 & \textbf{0.565} & 0.593 & \textbf{0.924} & 0.822 & \textbf{0.541}\\
\bottomrule
\end{tabular}
\end{table*}

\textcolor{black}{Fruitlet performance remains strong at both resolutions. YOLO11n-960 achieves the highest mAP$_{50}$ values, whereas YOLO11x-960 provides the highest recall and strict mAP$_{50:95}$. Across the five scales, the mean changes are modest relative to peduncle: box mAP$_{50}$ improves by only 0.014 on average, while mask mAP$_{50}$ improves by 0.015. This again indicates that fruitlet representation is already comparatively mature at 640 pixels.}

\paragraph{Calyx performance.}

\begin{table*}[ht!]
\centering
\caption{\textbf{Class-wise detection and instance-segmentation performance for the calyx class using YOLO11 models under 640$\times$640 and 960$\times$960 input resolutions.}}
\label{tab:yolo11_calyx_class}
\setlength{\tabcolsep}{3.8pt}
\renewcommand{\arraystretch}{1.14}
\small
\begin{tabular}{llcccccccc}
\toprule
\textbf{Model} & \textbf{Resolution} &
\multicolumn{4}{c}{\textbf{Bounding Box}} &
\multicolumn{4}{c}{\textbf{Mask}}\\
\cmidrule(lr){3-6}\cmidrule(lr){7-10}
& & P & R & mAP$_{50}$ & mAP$_{50:95}$ & P & R & mAP$_{50}$ & mAP$_{50:95}$\\
\midrule
YOLO11n-seg & 640 & 0.630 & 0.815 & 0.746 & 0.372 & 0.617 & 0.799 & 0.731 & 0.315\\
YOLO11n-seg & 960 & 0.668 & 0.831 & \textbf{0.759} & 0.393 & 0.670 & 0.836 & \textbf{0.771} & 0.370\\
YOLO11s-seg & 640 & 0.640 & 0.815 & 0.750 & 0.382 & 0.630 & 0.802 & 0.736 & 0.348\\
YOLO11s-seg & 960 & 0.713 & 0.720 & 0.744 & \textbf{0.400} & 0.716 & 0.723 & 0.743 & 0.369\\
YOLO11m-seg & 640 & 0.548 & 0.860 & 0.743 & 0.393 & 0.570 & 0.843 & 0.729 & 0.344\\
YOLO11m-seg & 960 & \textbf{0.730} & 0.745 & 0.706 & 0.386 & \textbf{0.728} & 0.742 & 0.698 & 0.343\\
YOLO11l-seg & 640 & 0.653 & 0.783 & 0.692 & 0.368 & 0.645 & 0.774 & 0.675 & 0.318\\
YOLO11l-seg & 960 & 0.587 & 0.864 & 0.691 & 0.389 & 0.588 & 0.866 & 0.691 & 0.354\\
YOLO11x-seg & 640 & 0.677 & 0.755 & 0.675 & 0.345 & 0.686 & 0.764 & 0.681 & 0.303\\
YOLO11x-seg & 960 & 0.550 & \textbf{0.882} & 0.722 & 0.399 & 0.555 & \textbf{0.885} & 0.724 & \textbf{0.373}\\
\bottomrule
\end{tabular}
\end{table*}

\textcolor{black}{Calyx shows a moderate benefit from increased resolution. YOLO11n-960 improves both box and mask AP, while YOLO11x-960 achieves the highest recall and mask mAP$_{50:95}$. Mean mask mAP$_{50:95}$ improves by 0.036 across the five scales, exceeding the corresponding mean mAP$_{50}$ gain of 0.015. This difference suggests that higher spatial resolution is particularly useful for more stringent calyx boundary delineation.}

\paragraph{Peduncle performance.}

\begin{table*}[ht!]
\centering
\caption{\textbf{Class-wise detection and instance-segmentation performance for the peduncle class using YOLO11 models under 640$\times$640 and 960$\times$960 input resolutions.}}
\label{tab:yolo11_peduncle_class}
\setlength{\tabcolsep}{3.8pt}
\renewcommand{\arraystretch}{1.14}
\small
\begin{tabular}{llcccccccc}
\toprule
\textbf{Model} & \textbf{Resolution} &
\multicolumn{4}{c}{\textbf{Bounding Box}} &
\multicolumn{4}{c}{\textbf{Mask}}\\
\cmidrule(lr){3-6}\cmidrule(lr){7-10}
& & P & R & mAP$_{50}$ & mAP$_{50:95}$ & P & R & mAP$_{50}$ & mAP$_{50:95}$\\
\midrule
YOLO11n-seg & 640 & 0.332 & 0.404 & 0.351 & 0.188 & 0.332 & 0.404 & 0.324 & 0.152\\
YOLO11n-seg & 960 & 0.432 & 0.442 & 0.404 & 0.250 & 0.432 & 0.442 & 0.401 & 0.207\\
YOLO11s-seg & 640 & 0.349 & 0.361 & 0.324 & 0.190 & 0.367 & 0.380 & 0.328 & 0.164\\
YOLO11s-seg & 960 & 0.540 & 0.423 & \textbf{0.521} & \textbf{0.318} & 0.540 & 0.423 & \textbf{0.512} & \textbf{0.297}\\
YOLO11m-seg & 640 & \textbf{0.657} & 0.346 & 0.479 & 0.280 & \textbf{0.747} & 0.284 & 0.481 & 0.244\\
YOLO11m-seg & 960 & 0.505 & 0.385 & 0.478 & 0.307 & 0.507 & 0.385 & 0.480 & 0.276\\
YOLO11l-seg & 640 & 0.550 & 0.442 & 0.453 & 0.278 & 0.550 & 0.442 & 0.467 & 0.227\\
YOLO11l-seg & 960 & 0.465 & \textbf{0.519} & 0.453 & 0.302 & 0.470 & \textbf{0.519} & 0.457 & 0.273\\
YOLO11x-seg & 640 & 0.481 & 0.518 & 0.458 & 0.285 & 0.481 & 0.518 & 0.450 & 0.222\\
YOLO11x-seg & 960 & 0.566 & 0.476 & 0.451 & 0.286 & 0.564 & 0.473 & 0.459 & 0.246\\
\bottomrule
\end{tabular}
\end{table*}

\textcolor{black}{Peduncle exhibits the strongest mean response to higher resolution in the YOLO11 family. YOLO11s is particularly notable: box mAP$_{50}$ increases from 0.324 to 0.521, box mAP$_{50:95}$ from 0.190 to 0.318, mask mAP$_{50}$ from 0.328 to 0.512, and mask mAP$_{50:95}$ from 0.164 to 0.297. These gains substantially exceed those observed for fruitlet.}

\begin{table*}[ht!]
\centering
\caption{\textbf{Mean absolute change in class-wise YOLO11 performance when increasing input resolution from 640$\times$640 to 960$\times$960 pixels across all five model scales.}}
\label{tab:yolo11_class_resolution_summary}
\setlength{\tabcolsep}{4pt}
\renewcommand{\arraystretch}{1.15}
\small
\begin{tabular}{lrrrrrrrr}
\toprule
\textbf{Class} &
$\Delta$\textbf{Box P} & $\Delta$\textbf{Box R} &
$\Delta$\textbf{Box mAP$_{50}$} & $\Delta$\textbf{Box mAP$_{50:95}$} &
$\Delta$\textbf{Mask P} & $\Delta$\textbf{Mask R} &
$\Delta$\textbf{Mask mAP$_{50}$} & $\Delta$\textbf{Mask mAP$_{50:95}$}\\
\midrule
Fruitlet & -0.005 & +0.011 & +0.014 & +0.033 & -0.011 & +0.015 & +0.015 & +0.042\\
Calyx & +0.020 & +0.003 & +0.003 & +0.021 & +0.022 & +0.014 & +0.015 & +0.036\\
Peduncle & \textbf{+0.028} & \textbf{+0.035} & \textbf{+0.048} & \textbf{+0.048} &
+0.007 & \textbf{+0.043} & \textbf{+0.052} & \textbf{+0.058}\\
\bottomrule
\end{tabular}
\end{table*}

\textcolor{black}{The mean paired comparison provides particularly strong evidence of disproportionate peduncle improvement. Peduncle produces the largest mean gain in box mAP$_{50}$ (+0.048), box mAP$_{50:95}$ (+0.048), mask mAP$_{50}$ (+0.052), and mask mAP$_{50:95}$ (+0.058). Calyx exhibits intermediate gains, while fruitlet improvements remain relatively modest. Thus, among the three generations evaluated, YOLO11 provides one of the clearest demonstrations that increased spatial resolution preferentially benefits the smallest anatomical structure.}

\subsubsection{YOLOv8 Performance}
\label{subsubsec:yolov8_classwise}

\begin{figure}[ht!]
     \centering
     \includegraphics[width=\linewidth]{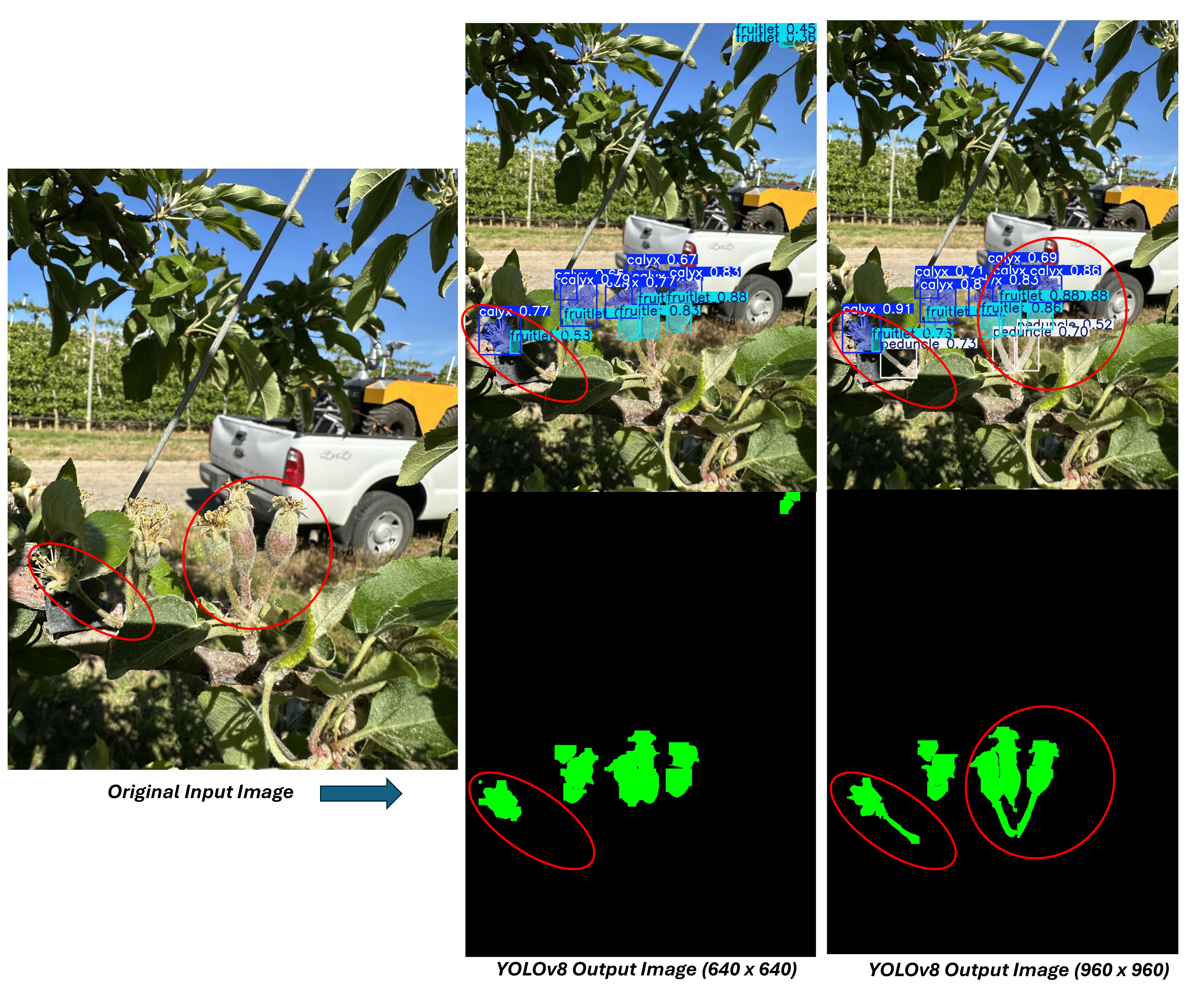}
\caption{\textcolor{black}{Qualitative comparison of YOLOv8 instance segmentation at $640\times640$ and $960\times960$ training resolutions. Solid red circles highlight peduncles missed by the $640\times640$ model but successfully detected and segmented at $960\times960$, demonstrating improved preservation and recognition of fine-scale anatomical structures under higher-resolution training.}}
    \label{fig:yolov8imagediscussion}
\end{figure}

\textcolor{black}{The class-wise YOLOv8 results provide an earlier-generation reference and confirm that the observed anatomical hierarchy is consistent across the evaluated YOLO families: fruitlet remains the comparatively easiest target, calyx exhibits intermediate difficulty, and peduncle represents the most challenging anatomical structure. This behavior is clearly illustrated in Fig.~\ref{fig:yolov8imagediscussion}, where both resolution configurations effectively detect and segment the larger fruitlet bodies and associated calyx structures within a densely clustered orchard scene.}

\textcolor{black}{More importantly, the two regions highlighted by solid red circles in Fig.~\ref{fig:yolov8imagediscussion} demonstrate a distinct resolution-dependent improvement in peduncle perception. The YOLOv8 model trained at $640\times640$ failed to detect the highlighted peduncles, whereas the corresponding $960\times960$ configuration successfully detected and segmented both structures in the same image. This qualitative evidence indicates that preserving greater spatial information during higher-resolution training can improve the representation of thin, low-contrast peduncles that may otherwise be lost at conventional resolution. Although the quantitative response to increased resolution varies among YOLOv8 model scales because of precision--recall trade-offs, the visual results demonstrate that the $960\times960$ small-object-focused configuration can recover anatomically important structures missed at $640\times640$, supporting its value for fine-grained orchard perception.}

\paragraph{Fruitlet performance.}

\begin{table*}[ht!]
\centering
\caption{\textbf{Class-wise detection and instance-segmentation performance for the fruitlet class using YOLOv8 models under 640$\times$640 and 960$\times$960 input resolutions.}}
\label{tab:yolov8_fruitlet_class}
\setlength{\tabcolsep}{3.8pt}
\renewcommand{\arraystretch}{1.14}
\small
\begin{tabular}{llcccccccc}
\toprule
\textbf{Model} & \textbf{Resolution} &
\multicolumn{4}{c}{\textbf{Bounding Box}} &
\multicolumn{4}{c}{\textbf{Mask}}\\
\cmidrule(lr){3-6}\cmidrule(lr){7-10}
& & P & R & mAP$_{50}$ & mAP$_{50:95}$ & P & R & mAP$_{50}$ & mAP$_{50:95}$\\
\midrule
YOLOv8n-seg & 640 & 0.734 & 0.787 & 0.775 & 0.492 & 0.726 & 0.777 & 0.770 & 0.458\\
YOLOv8n-seg & 960 & \textbf{0.765} & 0.765 & \textbf{0.823} & \textbf{0.557} & \textbf{0.771} & 0.771 & \textbf{0.824} & \textbf{0.532}\\
YOLOv8s-seg & 640 & 0.749 & 0.771 & 0.792 & 0.516 & 0.747 & 0.765 & 0.782 & 0.477\\
YOLOv8s-seg & 960 & 0.684 & 0.838 & 0.804 & 0.543 & 0.689 & 0.845 & 0.807 & 0.523\\
YOLOv8m-seg & 640 & 0.724 & 0.799 & 0.759 & 0.494 & 0.724 & 0.802 & 0.759 & 0.476\\
YOLOv8m-seg & 960 & 0.713 & 0.799 & 0.803 & 0.537 & 0.713 & 0.799 & 0.803 & 0.523\\
YOLOv8l-seg & 640 & 0.619 & \textbf{0.881} & 0.795 & 0.518 & 0.623 & \textbf{0.887} & 0.797 & 0.476\\
YOLOv8l-seg & 960 & 0.666 & 0.872 & 0.807 & 0.529 & 0.663 & 0.869 & 0.806 & 0.516\\
YOLOv8x-seg & 640 & 0.635 & 0.866 & 0.797 & 0.519 & 0.625 & 0.882 & 0.797 & 0.488\\
YOLOv8x-seg & 960 & 0.688 & 0.826 & 0.811 & 0.547 & 0.682 & 0.823 & 0.811 & 0.531\\
\bottomrule
\end{tabular}
\end{table*}

\textcolor{black}{Fruitlet remains highly detectable under both resolutions. YOLOv8n-960 achieves the highest AP-based performance, with box mAP$_{50:95}=0.557$ and mask mAP$_{50:95}=0.532$. Averaged across model scales, mask mAP$_{50}$ improves by 0.029 and mask mAP$_{50:95}$ by 0.050. These gains are meaningful, but they occur from an already comparatively high baseline.}

\paragraph{Calyx performance.}

\begin{table*}[ht!]
\centering
\caption{\textbf{Class-wise detection and instance-segmentation performance for the calyx class using YOLOv8 models under 640$\times$640 and 960$\times$960 input resolutions.}}
\label{tab:yolov8_calyx_class}
\setlength{\tabcolsep}{3.8pt}
\renewcommand{\arraystretch}{1.14}
\small
\begin{tabular}{llcccccccc}
\toprule
\textbf{Model} & \textbf{Resolution} &
\multicolumn{4}{c}{\textbf{Bounding Box}} &
\multicolumn{4}{c}{\textbf{Mask}}\\
\cmidrule(lr){3-6}\cmidrule(lr){7-10}
& & P & R & mAP$_{50}$ & mAP$_{50:95}$ & P & R & mAP$_{50}$ & mAP$_{50:95}$\\
\midrule
YOLOv8n-seg & 640 & 0.691 & 0.748 & 0.712 & 0.328 & 0.691 & 0.748 & 0.691 & 0.300\\
YOLOv8n-seg & 960 & 0.691 & 0.755 & 0.706 & 0.362 & 0.696 & 0.761 & 0.714 & 0.337\\
YOLOv8s-seg & 640 & \textbf{0.722} & 0.764 & 0.721 & 0.358 & \textbf{0.715} & 0.755 & 0.703 & 0.332\\
YOLOv8s-seg & 960 & 0.677 & 0.831 & 0.720 & 0.376 & 0.671 & 0.825 & 0.708 & 0.351\\
YOLOv8m-seg & 640 & 0.687 & 0.771 & 0.658 & 0.357 & 0.696 & 0.783 & 0.661 & 0.315\\
YOLOv8m-seg & 960 & 0.687 & 0.774 & \textbf{0.732} & 0.376 & 0.685 & 0.771 & 0.713 & \textbf{0.353}\\
YOLOv8l-seg & 640 & 0.560 & \textbf{0.854} & 0.705 & 0.346 & 0.547 & \textbf{0.834} & 0.688 & 0.305\\
YOLOv8l-seg & 960 & 0.661 & 0.815 & 0.727 & \textbf{0.380} & 0.666 & 0.822 & \textbf{0.719} & 0.349\\
YOLOv8x-seg & 640 & 0.598 & 0.825 & 0.701 & 0.354 & 0.570 & 0.828 & 0.700 & 0.326\\
YOLOv8x-seg & 960 & 0.658 & 0.755 & 0.677 & 0.365 & 0.655 & 0.761 & 0.679 & 0.337\\
\bottomrule
\end{tabular}
\end{table*}

Calyx shows a moderate resolution response. YOLOv8n improves strict box mAP$_{50:95}$ from 0.328 to 0.362 and strict mask AP from 0.300 to 0.337. YOLOv8m-960 provides the highest mask mAP$_{50:95}$ (0.353), while YOLOv8l-960 provides the highest box mAP$_{50:95}$ (0.380) and mask mAP$_{50}$ (0.719). Across all scales, mean mask mAP$_{50:95}$ improves by 0.030.

\paragraph{Peduncle performance.}

\begin{table*}[ht!]
\centering
\caption{\textbf{Class-wise detection and instance-segmentation performance for the peduncle class using YOLOv8 models under 640$\times$640 and 960$\times$960 input resolutions.}}
\label{tab:yolov8_peduncle_class}
\setlength{\tabcolsep}{3.8pt}
\renewcommand{\arraystretch}{1.14}
\small
\begin{tabular}{llcccccccc}
\toprule
\textbf{Model} & \textbf{Resolution} &
\multicolumn{4}{c}{\textbf{Bounding Box}} &
\multicolumn{4}{c}{\textbf{Mask}}\\
\cmidrule(lr){3-6}\cmidrule(lr){7-10}
& & P & R & mAP$_{50}$ & mAP$_{50:95}$ & P & R & mAP$_{50}$ & mAP$_{50:95}$\\
\midrule
YOLOv8n-seg & 640 & 0.459 & \textbf{0.654} & 0.493 & 0.249 & 0.473 & \textbf{0.673} & 0.484 & 0.216\\
YOLOv8n-seg & 960 & 0.402 & 0.365 & 0.409 & 0.251 & 0.402 & 0.365 & 0.419 & 0.243\\
YOLOv8s-seg & 640 & 0.497 & 0.442 & 0.485 & 0.295 & 0.499 & 0.442 & 0.458 & 0.259\\
YOLOv8s-seg & 960 & 0.455 & \textbf{0.654} & \textbf{0.551} & \textbf{0.338} & 0.455 & 0.654 & \textbf{0.556} & \textbf{0.296}\\
YOLOv8m-seg & 640 & 0.473 & 0.635 & 0.460 & 0.276 & 0.445 & 0.600 & 0.407 & 0.220\\
YOLOv8m-seg & 960 & \textbf{0.521} & 0.335 & 0.446 & 0.272 & \textbf{0.516} & 0.328 & 0.447 & 0.236\\
YOLOv8l-seg & 640 & 0.306 & 0.538 & 0.407 & 0.216 & 0.305 & 0.538 & 0.412 & 0.200\\
YOLOv8l-seg & 960 & 0.458 & 0.404 & 0.472 & 0.274 & 0.468 & 0.424 & 0.488 & 0.265\\
YOLOv8x-seg & 640 & 0.415 & 0.288 & 0.374 & 0.222 & 0.391 & 0.308 & 0.359 & 0.191\\
YOLOv8x-seg & 960 & 0.418 & 0.577 & 0.454 & 0.300 & 0.428 & 0.596 & 0.463 & 0.273\\
\bottomrule
\end{tabular}
\end{table*}

The small model exhibits the clearest improvement. YOLOv8s-960 raises peduncle box recall from 0.442 to 0.654, box mAP$_{50}$ from 0.485 to 0.551, mask mAP$_{50}$ from 0.458 to 0.556, and mask mAP$_{50:95}$ from 0.259 to 0.296. YOLOv8l and YOLOv8x also exhibit substantial gains in mask AP. However, YOLOv8n and YOLOv8m experience large recall reductions, indicating that the higher-resolution effect is more scale dependent in YOLOv8 than in YOLO11.

\begin{table*}[ht!]
\centering
\caption{\textbf{Mean absolute change in class-wise YOLOv8 performance when increasing input resolution from 640$\times$640 to 960$\times$960 pixels across all five model scales.}}
\label{tab:yolov8_class_resolution_summary}
\setlength{\tabcolsep}{4pt}
\renewcommand{\arraystretch}{1.15}
\small
\begin{tabular}{lrrrrrrrr}
\toprule
\textbf{Class} &
$\Delta$\textbf{Box P} & $\Delta$\textbf{Box R} &
$\Delta$\textbf{Box mAP$_{50}$} & $\Delta$\textbf{Box mAP$_{50:95}$} &
$\Delta$\textbf{Mask P} & $\Delta$\textbf{Mask R} &
$\Delta$\textbf{Mask mAP$_{50}$} & $\Delta$\textbf{Mask mAP$_{50:95}$}\\
\midrule
Fruitlet & +0.011 & -0.001 & +0.026 & +0.035 & +0.015 & -0.001 & +0.029 & \textbf{+0.050}\\
Calyx & +0.023 & -0.006 & +0.013 & +0.023 & +0.031 & -0.002 & +0.018 & +0.030\\
Peduncle & +0.021 & -0.044 & +0.023 & +0.035 & \textbf{+0.031} & -0.039 & \textbf{+0.051} & +0.045\\
\bottomrule
\end{tabular}
\end{table*}

\textcolor{black}{Although the mean YOLOv8 response is less uniform, peduncle still exhibits the largest increase in mask mAP$_{50}$ (+0.051), substantially exceeding calyx (+0.018) and fruitlet (+0.029). Its strict mask mAP$_{50:95}$ also improves by 0.045. The negative mean recall change results primarily from the substantial recall losses of the nano and medium configurations and therefore highlights an important distinction between improved spatial accuracy and improved object recovery.}

\textcolor{black}{Across all three generations, the class-wise analysis establishes a consistent anatomical hierarchy. Fruitlet is comparatively easy to detect and segment and therefore exhibits substantial performance saturation at 640 pixels. Calyx remains more sensitive to spatial resolution because of its smaller and irregular geometry, particularly when performance is evaluated under strict IoU thresholds. Peduncle constitutes the dominant perception bottleneck but also shows the largest resolution-dependent improvements in several AP measures. For YOLO26, mean peduncle mask mAP$_{50}$ increases by 0.077 compared with $-0.001$ for fruitlet; for YOLO11, the corresponding gain is 0.052 compared with 0.015; and for YOLOv8 it is 0.051 compared with 0.029. These cross-generation results therefore support the central hypothesis that increasing effective spatial resolution is particularly beneficial for anatomically small structures whose representation is most severely degraded during resizing.}

\textcolor{black}{At the same time, the observed precision--recall reversals demonstrate that the benefit of higher resolution cannot be interpreted as universally positive. Large and extra-large models frequently exhibit diminishing returns or shifts toward either greater selectivity or greater sensitivity rather than uniform AP improvement. The most consistent benefits occur in lightweight and intermediate-capacity models, especially for peduncle segmentation. This finding has direct implications for robotic green-fruit thinning: the perception architecture should not be selected solely according to whole-fruit accuracy or network size, because anatomically relevant attachment structures impose a substantially more demanding small-object segmentation problem. A configuration capable of preserving fine spatial information while maintaining moderate computational complexity may therefore provide a more useful perception substrate for downstream target selection and manipulation than a substantially larger network optimized primarily for aggregate detection performance.}

\subsection{Quantitative Evaluation of Detection and Segmentation Performance}
\label{subsec:quantitative_resolution_evaluation}

\textcolor{black}{The preceding generation-wise and class-wise analyses demonstrate that increased input resolution can improve fine-grained orchard perception, but they do not directly quantify whether the magnitude of this effect depends on model generation and network scale. To isolate the effect of spatial input resolution, corresponding 640$\times$640- and 960$\times$960-pixel configurations were therefore compared in a paired manner for all 15 model architectures spanning YOLOv8, YOLO11, and YOLO26. For a selected metric $M$, the resolution-induced change was defined as}

\begin{equation}
\Delta M = M_{960}-M_{640},
\label{eq:delta_metric}
\end{equation}

\textcolor{black}{where a positive $\Delta M$ indicates improved performance at 960 pixels and a negative value indicates degradation. In addition to precision, recall, and box and mask average precision, an aggregate detection F1-score was calculated from the reported bounding-box precision ($P$) and recall ($R$) as}

\begin{equation}
F_1 = 2\frac{PR}{P+R}.
\label{eq:f1}
\end{equation}

\textcolor{black}{The F1-score was derived post hoc because it was not explicitly reported in the training logs. Training duration and per-image inference latency were additionally compared to quantify the computational cost associated with increasing spatial resolution. This paired formulation is particularly useful because it separates the effect of input resolution from differences attributable to architecture generation or model capacity.}

\begin{table*}[ht!]
\centering
\caption{\textbf{Paired change in aggregate detection and instance-segmentation performance when increasing input resolution from 640$\times$640 to 960$\times$960 pixels.}
Changes are calculated as $\Delta M=M_{960}-M_{640}$. F1 was derived from aggregate bounding-box precision and recall. Positive values denote improvement under the 960-pixel configuration.}
\label{tab:paired_resolution_accuracy}

\setlength{\tabcolsep}{3.8pt}
\renewcommand{\arraystretch}{1.14}
\small

\begin{tabular}{lrrrrrrr}
\toprule
\textbf{Model} &
$\Delta$\textbf{P} &
$\Delta$\textbf{R} &
$\Delta$\textbf{F1} &
$\Delta$\textbf{Box mAP$_{50}$} &
$\Delta$\textbf{Box mAP$_{50:95}$} &
$\Delta$\textbf{Mask mAP$_{50}$} &
$\Delta$\textbf{Mask mAP$_{50:95}$} \\
\midrule

\multicolumn{8}{l}{\textit{YOLOv8}}\\
YOLOv8n-seg & -0.009 & -0.102 & -0.052 & -0.014 & +0.034 & +0.003 & +0.046\\
YOLOv8s-seg & -0.051 & +0.115 & +0.022 & +0.026 & +0.029 & +0.042 & +0.034\\
YOLOv8m-seg & +0.012 & -0.099 & -0.039 & +0.034 & +0.019 & +0.045 & +0.034\\
YOLOv8l-seg & +0.100 & -0.061 & +0.043 & +0.033 & +0.034 & +0.039 & +0.050\\
YOLOv8x-seg & +0.039 & +0.059 & +0.048 & +0.023 & +0.039 & +0.032 & +0.045\\
\midrule

\multicolumn{8}{l}{\textit{YOLO11}}\\
YOLO11n-seg & +0.049 & +0.015 & +0.036 & +0.029 & +0.039 & +0.047 & +0.049\\
YOLO11s-seg & +0.115 & -0.027 & +0.049 & +0.071 & +0.062 & +0.071 & +0.068\\
YOLO11m-seg & +0.039 & -0.057 & -0.006 & -0.018 & +0.010 & -0.017 & +0.021\\
YOLO11l-seg & -0.079 & +0.077 & -0.015 & +0.007 & +0.025 & +0.011 & +0.041\\
YOLO11x-seg & -0.055 & +0.075 & -0.002 & +0.020 & +0.033 & +0.024 & +0.048\\
\midrule

\multicolumn{8}{l}{\textit{YOLO26}}\\
YOLO26n-seg & +0.093 & -0.007 & +0.050 & +0.021 & +0.051 & +0.032 & +0.049\\
YOLO26s-seg & +0.053 & +0.105 & \textbf{+0.075} & +0.060 & +0.048 & \textbf{+0.079} & +0.056\\
YOLO26m-seg & -0.060 & +0.087 & +0.006 & +0.011 & +0.016 & +0.036 & +0.034\\
YOLO26l-seg & -0.037 & -0.029 & -0.033 & -0.028 & +0.006 & +0.011 & +0.020\\
YOLO26x-seg & +0.111 & -0.134 & +0.000 & -0.009 & +0.009 & -0.017 & -0.003\\

\bottomrule
\end{tabular}
\end{table*}

\textcolor{black}{Table~\ref{tab:paired_resolution_accuracy} demonstrates that the effect of increasing input resolution is strongly dependent on both architecture generation and model scale. The most consistent result across the complete benchmark is the improvement in the stringent mAP$_{50:95}$ metrics. For YOLOv8, all five model scales improve in both box and mask mAP$_{50:95}$ at 960 pixels. The same pattern is observed for YOLO11. YOLO26 similarly improves box mAP$_{50:95}$ for all five scales, while mask mAP$_{50:95}$ improves for n, s, m, and l and decreases only marginally for the extra-large model ($\Delta=-0.003$). Thus, increased input resolution most consistently improves spatial agreement under strict IoU criteria rather than universally increasing precision or recall.}

\textcolor{black}{The strongest individual response is observed in the small-scale configurations. YOLO11s exhibits the largest improvement in box mAP$_{50}$ (+0.071), box mAP$_{50:95}$ (+0.062), and one of the largest mask improvements (+0.071 and +0.068 for mAP$_{50}$ and mAP$_{50:95}$, respectively). YOLO26s similarly improves across all reported aggregate performance metrics, including precision (+0.053), recall (+0.105), F1 (+0.075), box mAP$_{50}$ (+0.060), box mAP$_{50:95}$ (+0.048), mask mAP$_{50}$ (+0.079), and mask mAP$_{50:95}$ (+0.056). YOLOv8s also shows a substantial recall increase (+0.115) together with positive changes in all four AP measures. Therefore, the small configuration is the only model scale for which the higher-resolution representation produces consistently strong AP gains across all three architecture generations.}

\textcolor{black}{The nano configurations show a more generation-dependent response. YOLO11n benefits uniformly, with all seven performance indicators increasing at 960 pixels. YOLO26n also shows clear improvement in F1 (+0.050), box mAP$_{50:95}$ (+0.051), and mask mAP$_{50:95}$ (+0.049), although recall remains nearly unchanged. In contrast, YOLOv8n exhibits a substantial reduction in recall ($-0.102$) and F1 ($-0.052$), despite improvements of +0.034 and +0.046 in box and mask mAP$_{50:95}$, respectively. Thus, higher spatial resolution appears to interact more favorably with the nano architectures of the newer YOLO generations, while the earlier YOLOv8 nano network exhibits a stronger sensitivity--localization trade-off.}

\textcolor{black}{The medium-scale models exhibit substantially weaker and less consistent gains. YOLOv8m improves all four AP measures but loses 0.099 in recall, resulting in a lower F1-score. YOLO11m exhibits the weakest AP response within its family: box mAP$_{50}$ decreases by 0.018 and mask mAP$_{50}$ by 0.017, although both stringent mAP$_{50:95}$ measures increase. YOLO26m produces modest AP gains and a substantial recall increase (+0.087), but precision decreases by 0.060. These results indicate that medium-capacity models primarily undergo a redistribution of the precision--recall operating point rather than a uniformly positive accuracy shift.}

\textcolor{black}{A similar saturation becomes apparent for the large and extra-large architectures. YOLO26l loses 0.028 in box mAP$_{50}$ and 0.033 in F1, while gaining only 0.006 and 0.020 in box and mask mAP$_{50:95}$, respectively. YOLO26x provides the clearest example of diminishing returns: precision increases by 0.111, but recall decreases by 0.134, leaving F1 essentially unchanged; box mAP$_{50}$ decreases by 0.009 and mask mAP$_{50}$ by 0.017. YOLO11l and YOLO11x similarly show pronounced precision--recall shifts, although their stringent AP metrics remain improved. These results suggest that once substantial representational capacity is already available, additional input resolution contributes progressively smaller improvements and may primarily alter prediction confidence and sensitivity.}

The scale dependence becomes clearer when the paired changes are averaged across the three YOLO generations, as shown in Table~\ref{tab:scale_mean_delta}.

\begin{table*}[ht!]
\centering
\caption{\textbf{Mean resolution-induced performance change for each model scale across YOLOv8, YOLO11, and YOLO26.}
Values represent the mean $\Delta M$ for the three architecture generations.}
\label{tab:scale_mean_delta}

\setlength{\tabcolsep}{4.5pt}
\renewcommand{\arraystretch}{1.15}
\small

\begin{tabular}{lrrrrrrr}
\toprule
\textbf{Scale} &
$\Delta$\textbf{P} &
$\Delta$\textbf{R} &
$\Delta$\textbf{F1} &
$\Delta$\textbf{Box mAP$_{50}$} &
$\Delta$\textbf{Box mAP$_{50:95}$} &
$\Delta$\textbf{Mask mAP$_{50}$} &
$\Delta$\textbf{Mask mAP$_{50:95}$}\\
\midrule

Nano (n)  & +0.044 & -0.031 & +0.011 & +0.012 & +0.041 & +0.027 & +0.048\\
Small (s) & +0.039 & \textbf{+0.064} & \textbf{+0.049} & \textbf{+0.052} &
\textbf{+0.046} & \textbf{+0.064} & \textbf{+0.053}\\
Medium (m)& -0.003 & -0.023 & -0.013 & +0.009 & +0.015 & +0.021 & +0.030\\
Large (l) & -0.005 & -0.004 & -0.002 & +0.004 & +0.022 & +0.020 & +0.037\\
Extra-large (x) & +0.032 & +0.000 & +0.015 & +0.011 & +0.027 & +0.013 & +0.030\\

\bottomrule
\end{tabular}
\end{table*}

\textcolor{black}{The cross-generation scale averages provide strong evidence for a resolution--capacity interaction. The small configuration exhibits the largest mean improvement in every AP metric: +0.052 in box mAP$_{50}$, +0.046 in box mAP$_{50:95}$, +0.064 in mask mAP$_{50}$, and +0.053 in mask mAP$_{50:95}$. It also produces the largest mean F1 improvement (+0.049) and recall gain (+0.064). By comparison, medium and large architectures show mean F1 changes of $-0.013$ and $-0.002$, respectively. Therefore, the benefit of increasing resolution is not proportional to model capacity; instead, it is maximized in a relatively compact capacity regime.}

\textcolor{black}{One interpretation is that the small models possess sufficient feature-extraction capacity to exploit the additional fine-scale information retained at 960 pixels, while still being sufficiently capacity constrained for that additional spatial information to materially improve representation. Nano networks can also benefit substantially, but their response is more generation dependent, potentially because extremely restricted feature capacity limits how consistently the additional information is converted into improved object recovery. Conversely, large and extra-large architectures already possess substantial representational capacity at 640 pixels, leading to diminishing returns when only input resolution is increased. This interpretation is consistent with the observation that stringent AP often improves for large models even when F1, precision, or recall does not.}

Generation-specific mean changes are summarized in Table~\ref{tab:generation_mean_delta}.

\begin{table*}[ht!]
\centering
\caption{\textbf{Mean resolution-induced performance change for each YOLO generation across the five n/s/m/l/x model scales.}}
\label{tab:generation_mean_delta}

\setlength{\tabcolsep}{4.5pt}
\renewcommand{\arraystretch}{1.15}
\small

\begin{tabular}{lrrrrrrr}
\toprule
\textbf{Generation} &
$\Delta$\textbf{P} &
$\Delta$\textbf{R} &
$\Delta$\textbf{F1} &
$\Delta$\textbf{Box mAP$_{50}$} &
$\Delta$\textbf{Box mAP$_{50:95}$} &
$\Delta$\textbf{Mask mAP$_{50}$} &
$\Delta$\textbf{Mask mAP$_{50:95}$}\\
\midrule

YOLOv8  & +0.018 & -0.018 & +0.004 & +0.020 & +0.031 & \textbf{+0.032} & +0.042\\
YOLO11 & +0.014 & +0.017 & +0.012 & \textbf{+0.022} & \textbf{+0.034} & +0.027 & \textbf{+0.045}\\
YOLO26 & \textbf{+0.032} & +0.004 & \textbf{+0.020} & +0.011 & +0.026 & +0.028 & +0.031\\

\bottomrule
\end{tabular}
\end{table*}

\textcolor{black}{Generation-wise averaging reveals a second important interaction. YOLO11 exhibits the largest mean improvement in box mAP$_{50}$ (+0.022), box mAP$_{50:95}$ (+0.034), and mask mAP$_{50:95}$ (+0.045), whereas YOLOv8 produces the largest average increase in mask mAP$_{50}$ (+0.032). YOLO26 provides the largest average precision (+0.032) and F1 (+0.020) improvements but smaller average AP gains because its large and extra-large variants exhibit substantial saturation. Consequently, newer generation alone does not imply a stronger response to higher input resolution. The interaction between generation and model scale is more informative than generation-level averaging.}

\textcolor{black}{The computational cost of these accuracy gains is shown in Table~\ref{tab:resolution_efficiency_change}. Training-time changes are reported both as absolute additional hours and as percentage increases relative to the corresponding 640-pixel run. Inference changes are similarly reported in milliseconds and percentage terms.}

\begin{table*}[ht!]
\centering
\caption{\textbf{Paired changes in training duration and validation-time inference latency between 640$\times$640 and 960$\times$960 configurations.}
Percentage changes are relative to the corresponding 640-pixel configuration. Positive values indicate increased computational cost.}
\label{tab:resolution_efficiency_change}

\setlength{\tabcolsep}{4.2pt}
\renewcommand{\arraystretch}{1.14}
\small

\begin{tabular}{lrrrrrr}
\toprule
\textbf{Model} &
\textbf{Train$_{640}$ (h)} &
\textbf{Train$_{960}$ (h)} &
$\Delta$\textbf{Train (h)} &
\textbf{Train change (\%)} &
$\Delta$\textbf{Infer. (ms)} &
\textbf{Infer. change (\%)}\\
\midrule

\multicolumn{7}{l}{\textit{YOLOv8}}\\
YOLOv8n-seg & 0.462 & 0.792 & +0.330 & +71.4 & +0.5 & +41.7\\
YOLOv8s-seg & 0.369 & 0.724 & +0.355 & +96.2 & -0.4 & -14.3\\
YOLOv8m-seg & 0.406 & 0.615 & +0.209 & +51.5 & +0.8 & +14.8\\
YOLOv8l-seg & 0.332 & 0.791 & +0.459 & +138.3 & +4.7 & +104.4\\
YOLOv8x-seg & 0.500 & 1.214 & +0.714 & +142.8 & +5.7 & +96.6\\
\midrule

\multicolumn{7}{l}{\textit{YOLO11}}\\
YOLO11n-seg & 0.462 & 0.685 & +0.223 & +48.3 & +1.8 & +180.0\\
YOLO11s-seg & 0.431 & 0.838 & +0.407 & +94.4 & +1.5 & +65.2\\
YOLO11m-seg & 0.373 & 0.939 & +0.566 & +151.7 & +1.8 & +47.4\\
YOLO11l-seg & 0.586 & 0.931 & +0.345 & +58.9 & +2.6 & +44.8\\
YOLO11x-seg & 0.836 & 1.391 & +0.555 & +66.4 & +6.3 & +95.5\\
\midrule

\multicolumn{7}{l}{\textit{YOLO26}}\\
YOLO26n-seg & 0.933 & 1.073 & +0.140 & +15.0 & -0.1 & -5.3\\
YOLO26s-seg & 1.019 & 1.076 & +0.057 & +5.6 & +1.2 & +66.7\\
YOLO26m-seg & 0.998 & 1.257 & +0.259 & +26.0 & +2.9 & +96.7\\
YOLO26l-seg & 1.051 & 1.409 & +0.358 & +34.1 & +3.8 & +97.4\\
YOLO26x-seg & 1.034 & 2.124 & +1.090 & +105.4 & +6.5 & +103.2\\

\bottomrule
\end{tabular}
\end{table*}

\textcolor{black}{Higher input resolution generally increases training duration, but the magnitude of the increase differs substantially among generations. For YOLOv8, the additional training cost ranges from 0.209~h for the medium model to 0.714~h for the extra-large model. YOLO11 shows increases between 0.223 and 0.566~h, while YOLO26 exhibits comparatively modest increases for n, s, and m but a substantially larger +1.090~h penalty for YOLO26x. The latter configuration therefore provides a particularly unfavorable resolution trade-off: training time more than doubles while mask mAP$_{50:95}$ decreases slightly and F1 remains essentially unchanged.}

\textcolor{black}{Inference latency similarly increases most strongly for the largest models. YOLO26x increases from 6.3 to 12.8~ms per image, while YOLO11x increases from 6.6 to 12.9~ms and YOLOv8x from 5.9 to 11.6~ms. In contrast, the additional latency of small models remains substantially lower. The isolated negative inference-time changes for YOLOv8s and YOLO26n should not be interpreted as evidence that higher-resolution processing is computationally cheaper. Because the architecture is unchanged and a larger spatial tensor must theoretically be processed, these small inversions are more appropriately attributed to run-to-run variation in validation-time timing measurements.}

\textcolor{black}{When accuracy gains are considered jointly with computational cost, the resolution--scale interaction becomes especially important. The small models produce the largest average AP improvements across generations while avoiding the severe computational penalties associated with the extra-large configurations. YOLO11s-960 and YOLO26s-960 are particularly favorable examples: both exhibit substantial improvements in strict box and mask AP while retaining substantially lower inference latency than their corresponding large and extra-large variants. Conversely, increasing resolution for already large-capacity models often produces relatively small strict-AP improvements at disproportionately greater training and inference cost.}

\textcolor{black}{These findings reject a simple interpretation in which either higher resolution or greater model size is universally preferable. Instead, the results identify a three-way interaction among \textit{architecture generation}, \textit{model capacity}, and \textit{input resolution}. The 960-pixel configuration consistently improves stringent spatial-accuracy measures, but the magnitude of that gain is greatest in specific capacity regimes, particularly the small models. Nano architectures also frequently benefit, whereas large and extra-large models exhibit diminishing returns and stronger precision--recall redistribution. From a robotic-perception perspective, this interaction is more consequential than the absolute maximum accuracy of an isolated configuration because a useful orchard vision model must convert additional spatial information into meaningful boundary accuracy without imposing disproportionate computational latency.}

\begin{figure*}[ht!]
\centering
\begin{tikzpicture}
\begin{axis}[
    width=0.96\textwidth,
    height=0.47\textwidth,
    xlabel={Performance metric},
    ylabel={Model configuration},
    xtick={0,1,2,3},
    xticklabels={
        Box mAP$_{50}$,
        Box mAP$_{50:95}$,
        Mask mAP$_{50}$,
        Mask mAP$_{50:95}$
    },
    ytick={0,...,14},
    yticklabels={
        YOLOv8n,
        YOLOv8s,
        YOLOv8m,
        YOLOv8l,
        YOLOv8x,
        YOLO11n,
        YOLO11s,
        YOLO11m,
        YOLO11l,
        YOLO11x,
        YOLO26n,
        YOLO26s,
        YOLO26m,
        YOLO26l,
        YOLO26x
    },
    x tick label style={font=\footnotesize, rotate=15, anchor=east},
    y tick label style={font=\footnotesize},
    point meta min=-0.03,
    point meta max=0.08,
    colorbar,
    colorbar style={
        ylabel={$\Delta$mAP = mAP$_{960}$--mAP$_{640}$},
        ylabel style={font=\footnotesize},
        tick label style={font=\scriptsize}
    },
    colormap/viridis,
    enlargelimits=false,
    axis on top,
]

\addplot[
    matrix plot*,
    mesh/cols=4,
    point meta=explicit
]
table[meta=value] {
x y value
0 0 -0.014
1 0  0.034
2 0  0.003
3 0  0.046

0 1  0.026
1 1  0.029
2 1  0.042
3 1  0.034

0 2  0.034
1 2  0.019
2 2  0.045
3 2  0.034

0 3  0.033
1 3  0.034
2 3  0.039
3 3  0.050

0 4  0.023
1 4  0.039
2 4  0.032
3 4  0.045

0 5  0.029
1 5  0.039
2 5  0.047
3 5  0.049

0 6  0.071
1 6  0.062
2 6  0.071
3 6  0.068

0 7 -0.018
1 7  0.010
2 7 -0.017
3 7  0.021

0 8  0.007
1 8  0.025
2 8  0.011
3 8  0.041

0 9  0.020
1 9  0.033
2 9  0.024
3 9  0.048

0 10 0.021
1 10 0.051
2 10 0.032
3 10 0.049

0 11 0.060
1 11 0.048
2 11 0.079
3 11 0.056

0 12 0.011
1 12 0.016
2 12 0.036
3 12 0.034

0 13 -0.028
1 13  0.006
2 13  0.011
3 13  0.020

0 14 -0.009
1 14  0.009
2 14 -0.017
3 14 -0.003
};

\end{axis}
\end{tikzpicture}

\caption{\textbf{Resolution-induced change in aggregate detection and segmentation accuracy across 15 YOLO model configurations.}
Each heatmap cell represents $\Delta$mAP = mAP$_{960}$--mAP$_{640}$ for the corresponding model and performance metric. Positive values indicate improved performance under the 960$\times$960 small-object-focused configuration, whereas negative values indicate reduced performance. The strongest and most consistent gains occur for the small-scale YOLO11s and YOLO26s configurations, while several large and extra-large configurations exhibit diminishing returns.}
\label{fig:resolution_delta_heatmap}
\end{figure*}
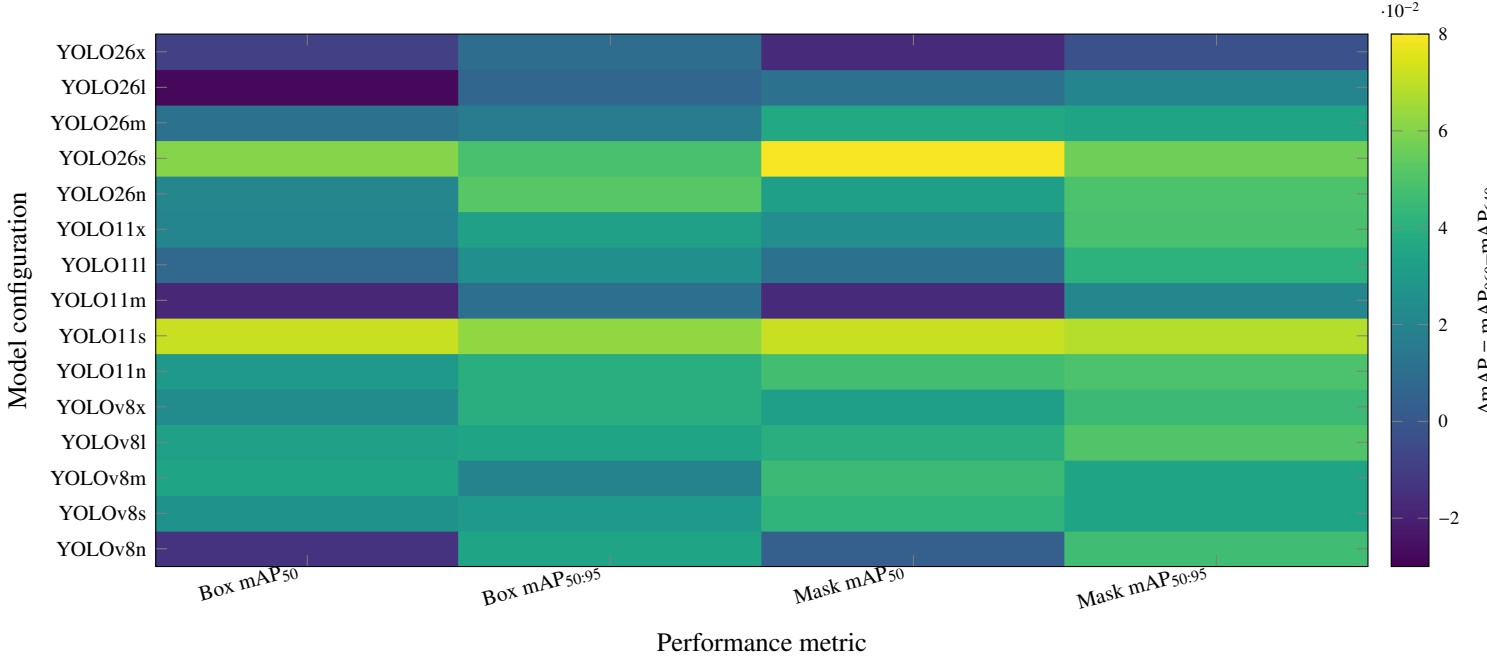

\begin{figure}[ht!]
\centering
\begin{tikzpicture}

\begin{axis}[
    width=\columnwidth,
    height=0.75\columnwidth,
    xlabel={Model scale},
    ylabel={$\Delta$ Mask mAP$_{50:95}$},
    symbolic x coords={n,s,m,l,x},
    xtick=data,
    ymin=-0.01,
    ymax=0.075,
    ytick={-0.01,0,0.01,0.02,0.03,0.04,0.05,0.06,0.07},
    grid=major,
    grid style={gray!20},
    axis line style={black},
    tick label style={font=\footnotesize},
    label style={font=\small},
    legend style={
        at={(0.5,1.03)},
        anchor=south,
        legend columns=3,
        font=\footnotesize,
        draw=none
    },
]

\addplot[
    color=blue!80!black,
    mark=*,
    mark size=2.5pt,
    line width=1.3pt,
    solid
] coordinates {
    (n,0.046)
    (s,0.034)
    (m,0.034)
    (l,0.050)
    (x,0.045)
};
\addlegendentry{YOLOv8}

\addplot[
    color=green!60!black,
    mark=square*,
    mark size=2.5pt,
    line width=1.3pt,
    dashed
] coordinates {
    (n,0.049)
    (s,0.068)
    (m,0.021)
    (l,0.041)
    (x,0.048)
};
\addlegendentry{YOLO11}

\addplot[
    color=red!80!black,
    mark=triangle*,
    mark size=3pt,
    line width=1.3pt,
    dashdotted
] coordinates {
    (n,0.049)
    (s,0.056)
    (m,0.034)
    (l,0.020)
    (x,-0.003)
};
\addlegendentry{YOLO26}

\addplot[
    black,
    dotted,
    line width=0.9pt
] coordinates {
    (n,0)
    (x,0)
};

\end{axis}
\end{tikzpicture}

\caption{\textbf{Resolution-induced change in mask mAP$_{50:95}$ across
YOLO generations and model scales.}
Each point represents
$\Delta\mathrm{mAP}_{50:95}
=\mathrm{mAP}_{50:95}^{960}
-\mathrm{mAP}_{50:95}^{640}$.
Positive values indicate improved instance-segmentation performance
when increasing input resolution from 640$\times$640 to
960$\times$960 pixels, whereas negative values indicate reduced
performance. YOLO11s exhibits the largest improvement
($\Delta=+0.068$), while YOLO26x shows a slight reduction
($\Delta=-0.003$), demonstrating that the effect of increased
spatial resolution depends jointly on architecture generation and
model scale.}
\label{fig:generation_scale_resolution_interaction}

\end{figure}
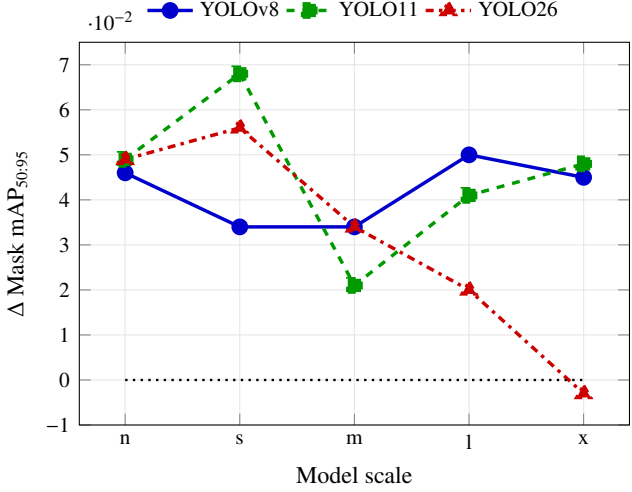

\subsection{Model Complexity, Processing Speed, and Training Efficiency}
\label{subsec:model_complexity_efficiency}

\textcolor{black}{Predictive accuracy alone is insufficient for selecting a perception model
for robotic orchard deployment because improvements in detection and
instance segmentation must be considered jointly with network complexity,
memory requirements, image-processing latency, and training cost.
This consideration is particularly important for the present application,
where fruitlet, calyx, and peduncle perception must ultimately operate as
part of a closed-loop robotic system in which image acquisition, perception,
target selection, motion planning, and manipulation compete for a finite
computational and temporal budget. Accordingly, the previously separated
analyses of architectural complexity and image-processing speed were
consolidated to characterize the accuracy--efficiency trade-off across
YOLOv8, YOLO11, and YOLO26.}

\textcolor{black}{The computational analysis considered the number of fused model layers,
trainable parameters, GFLOPs, stored checkpoint size, preprocessing time,
inference time, postprocessing time, total per-image processing time,
wall-clock training duration, and convergence behavior. The architectural
properties of a given model are identical between its 640$\times$640 and
960$\times$960 configurations because changing the training input resolution
does not add network layers or trainable parameters. Consequently,
resolution-dependent differences in processing and training time reflect the
larger spatial representation presented to the same underlying architecture,
rather than an increase in model capacity.}

\begin{table*}[ht!]
\centering
\caption{\textbf{Cross-generation architectural complexity and stored model
size of the 15 YOLO instance-segmentation architectures.}
Parameters and GFLOPs correspond to the fused Ultralytics inference models.
Model size corresponds to the optimizer-stripped best-performing checkpoint.
The architecture is identical for the corresponding 640- and 960-pixel
configurations.}
\label{tab:cross_generation_complexity}
\setlength{\tabcolsep}{6pt}
\renewcommand{\arraystretch}{1.12}
\small
\begin{tabular}{llrrrr}
\toprule
\textbf{Generation} & \textbf{Model} &
\textbf{Layers} & \textbf{Parameters (M)} &
\textbf{GFLOPs} & \textbf{Model size (MB)}\\
\midrule

\multirow{5}{*}{YOLOv8}
& YOLOv8n-seg & 86  & 3.259  & 11.3  & 6.8\\
& YOLOv8s-seg & 86  & 11.781 & 39.9  & 23.9\\
& YOLOv8m-seg & 106 & 27.224 & 104.3 & 54.8\\
& YOLOv8l-seg & 126 & 45.914 & 210.1 & 92.3\\
& YOLOv8x-seg & 126 & 71.724 & 328.0 & 144.0\\
\midrule

\multirow{5}{*}{YOLO11}
& YOLO11n-seg & 114 & 2.835  & 9.6   & 6.0\\
& YOLO11s-seg & 114 & 10.068 & 32.8  & 20.5\\
& YOLO11m-seg & 139 & 22.338 & 112.9 & 45.2\\
& YOLO11l-seg & 204 & 27.587 & 131.8 & 55.9\\
& YOLO11x-seg & 204 & 62.006 & 295.9 & 124.8\\
\midrule

\multirow{5}{*}{YOLO26}
& YOLO26n-seg & 139 & \textbf{2.689} & \textbf{9.0} & 6.6\\
& YOLO26s-seg & 139 & 10.367 & 34.1 & 23.4\\
& YOLO26m-seg & 149 & 23.510 & 121.2 & 54.5\\
& YOLO26l-seg & \textbf{207} & 27.906 & 139.4 & 63.5\\
& YOLO26x-seg & \textbf{207} & 62.730 & 313.0 & 141.8\\

\bottomrule
\end{tabular}
\end{table*}

\textcolor{black}{Table~\ref{tab:cross_generation_complexity} demonstrates that model scale
produces a substantial increase in computational demand, but the scaling
pattern differs among generations. YOLOv8n contains 3.259~M parameters and
requires 11.3~GFLOPs, whereas YOLOv8x increases to 71.724~M parameters and
328.0~GFLOPs. This corresponds to approximately 22 times more parameters and
29 times greater floating-point computational demand. YOLO11 scales from
2.835~M parameters and 9.6~GFLOPs for the nano configuration to
62.006~M parameters and 295.9~GFLOPs for the extra-large configuration.
YOLO26 follows a similar pattern, increasing from only 2.689~M parameters
and 9.0~GFLOPs for YOLO26n to 62.730~M parameters and 313.0~GFLOPs for
YOLO26x.}

\textcolor{black}{The minimum parameter count and GFLOP requirement are therefore obtained
with YOLO26n-seg, at 2.689~M parameters and 9.0~GFLOPs. These values are
particularly attractive for resource-constrained robotic systems because
they imply lower weight-storage requirements and substantially fewer
floating-point operations per forward pass. In contrast, YOLOv8x-seg
contains the largest parameter count (71.724~M), requires the greatest
computational workload (328.0~GFLOPs), and produces the largest stored
checkpoint (144.0~MB). A high parameter count represents increased learned
capacity, while a high GFLOP count indicates greater computational demand;
neither quantity should therefore be interpreted as a performance metric in
itself.}

\textcolor{black}{An important cross-generation observation is that architectural recency does
not necessarily imply greater computational complexity. At the nano scale,
YOLO26n is computationally lighter than YOLOv8n despite belonging to the
newer generation. At the extra-large scale, YOLO26x contains approximately
12.5\% fewer parameters than YOLOv8x (62.730 versus 71.724~M) and requires
4.6\% fewer GFLOPs (313.0 versus 328.0). YOLO11x is lighter still at
62.006~M parameters and 295.9~GFLOPs. These differences indicate that
cross-generation architectural development changes how representational
capacity is allocated rather than simply increasing network size.}

\textcolor{black}{The number of layers must similarly be interpreted together with parameter
count rather than independently. YOLO26l and YOLO26x both contain
207 fused layers, the largest depth among the evaluated architectures, while
YOLOv8n and YOLOv8s contain only 86 layers. However, YOLO26l contains
27.906~M parameters compared with 62.730~M for YOLO26x despite identical
reported layer counts. Network depth therefore does not directly determine
model size or computational burden; layer width, channel dimensions,
internal block design, and feature-processing operations jointly determine
the resulting parameter and FLOP requirements.}

\textcolor{black}{Stored checkpoint size closely follows parameter count. The smallest
checkpoint is YOLO11n at approximately 6.0~MB, followed by YOLO26n at
6.6~MB and YOLOv8n at 6.8~MB. At the opposite extreme, YOLOv8x,
YOLO26x, and YOLO11x require approximately 144.0, 141.8, and 124.8~MB,
respectively. The approximately 20-fold increase in checkpoint size between
nano and extra-large configurations has practical implications for model
storage, deployment, loading, memory transfer, and edge-computing
environments, even when sufficient GPU throughput is available for
inference.}

\begin{table}[ht!]
\centering
\caption{\textbf{Per-image validation processing time for YOLOv8, YOLO11,
and YOLO26 under 640$\times$640 and 960$\times$960 configurations.}
Total time is calculated as preprocessing + inference + postprocessing.
All values are milliseconds per image.}
\label{tab:cross_generation_processing}
\setlength{\tabcolsep}{4.2pt}
\renewcommand{\arraystretch}{1.10}
\scriptsize
\begin{tabular}{llrrrrrrrr}
\toprule
& & \multicolumn{4}{c}{\textbf{640$\times$640}} &
\multicolumn{4}{c}{\textbf{960$\times$960}}\\
\cmidrule(lr){3-6}\cmidrule(lr){7-10}
\textbf{Gen.} & \textbf{Model} &
\textbf{Pre} & \textbf{Infer.} & \textbf{Post} & \textbf{Total} &
\textbf{Pre} & \textbf{Infer.} & \textbf{Post} & \textbf{Total}\\
\midrule

\multirow{5}{*}{YOLO26}
& n & 0.1 & 1.9 & 0.3 & \textbf{2.3} & 0.2 & 1.8 & 0.6 & \textbf{2.6}\\
& s & 0.3 & 1.8 & 0.4 & 2.5 & 0.2 & 3.0 & 0.8 & 4.0\\
& m & 0.1 & 3.0 & 0.4 & 3.5 & 0.2 & 5.9 & 0.8 & 6.9\\
& l & 0.2 & 3.9 & 0.3 & 4.4 & 0.2 & 7.7 & 0.6 & 8.5\\
& x & 0.1 & 6.3 & 0.9 & 7.3 & 0.2 & 12.8 & 0.7 & 13.7\\
\midrule

\multirow{5}{*}{YOLO11}
& n & 0.8 & 1.0 & 2.4 & 4.2 & 0.5 & 2.8 & 1.9 & 5.2\\
& s & 0.8 & 2.3 & 3.5 & 6.6 & 0.3 & 3.8 & 1.6 & 5.7\\
& m & 0.8 & 3.8 & 2.4 & 7.0 & 0.3 & 5.6 & 0.9 & 6.8\\
& l & 1.5 & 5.8 & 2.5 & 9.8 & 0.2 & 8.4 & 1.2 & 9.8\\
& x & 1.9 & 6.6 & 2.9 & 11.4 & 0.3 & 12.9 & 1.9 & 15.1\\
\midrule

\multirow{5}{*}{YOLOv8}
& n & 0.4 & 1.2 & 2.2 & 3.8 & 0.2 & 1.7 & 2.3 & 4.2\\
& s & 0.9 & 2.8 & 0.9 & 4.6 & 0.4 & 2.4 & 0.9 & 3.7\\
& m & 1.4 & 5.4 & 1.2 & 8.0 & 0.4 & 6.2 & 1.5 & 8.1\\
& l & 1.2 & 4.5 & 1.2 & 6.9 & 0.2 & 9.2 & 2.4 & 11.8\\
& x & 1.4 & 5.9 & 2.0 & 9.3 & 0.2 & 11.6 & 1.0 & 12.8\\
\midrule

\bottomrule
\end{tabular}
\end{table}

\textcolor{black}{The runtime results in Table~\ref{tab:cross_generation_processing} show that
inference, rather than preprocessing, increasingly dominates total latency
as network scale and input resolution increase. Preprocessing ranges from
only 0.1 to 1.9~ms per image, whereas inference ranges from 1.0~ms for
YOLO11n at 640 pixels to 12.9~ms for YOLO11x at 960 pixels.
Postprocessing ranges from 0.3 to 3.5~ms. Consequently, optimization of the
forward network pass becomes progressively more important than preprocessing
for the larger models.}

\textcolor{black}{YOLO26 exhibits particularly favorable processing efficiency at
640 pixels. YOLO26n requires approximately 2.3~ms in total
(0.1~ms preprocessing, 1.9~ms inference, and 0.3~ms postprocessing),
while YOLO26s requires approximately 2.5~ms. Even YOLO26m requires only
3.5~ms. These measurements correspond to theoretical validation throughputs
of approximately 435, 400, and 286 images~s$^{-1}$, respectively, if the
reported per-image processing time is considered in isolation. These values
should not be interpreted as complete robotic-system frame rates because
camera acquisition, CPU--GPU transfer, depth processing, target association,
decision making, ROS communication, visualization, and manipulator control
would introduce additional latency. Nevertheless, they provide a consistent
basis for relative model comparison on the common experimental hardware.}

\textcolor{black}{Increasing input resolution to 960 pixels generally increases inference
cost, particularly for medium-to-extra-large architectures. YOLO26m
inference increases from 3.0 to 5.9~ms, YOLO26l from 3.9 to 7.7~ms, and
YOLO26x from 6.3 to 12.8~ms. The corresponding total processing times are
6.9, 8.5, and 13.7~ms. Similar behavior is observed for YOLO11x
(6.6 to 12.9~ms inference) and YOLOv8x (5.9 to 11.6~ms). Thus, the
computational penalty associated with higher spatial resolution becomes
particularly pronounced when it is combined with high-capacity networks.}

\textcolor{black}{The small timing inversions observed for selected configurations for
example, YOLOv8s inference decreasing from 2.8 to 2.4~ms and YOLO26n
decreasing from 1.9 to 1.8~ms, should not be interpreted as evidence that
processing a 960-pixel image is intrinsically cheaper. These differences are
small relative to the overall benchmark and can arise from GPU scheduling,
warm-up state, memory-transfer behavior, kernel selection, or timing
variability during validation. The broader scaling trend, particularly from
the medium through extra-large configurations, clearly demonstrates the
expected increase in inference cost at higher spatial resolution.}

\textcolor{black}{The relationship between computational complexity and predictive performance
is non-monotonic. This is particularly evident within YOLO26. Increasing
capacity from YOLO26m to YOLO26x raises the parameter count from
23.510~M to 62.730~M (+166.8\%) and GFLOPs from 121.2 to 313.0
(+158.3\%). At 960 pixels, inference latency simultaneously increases from
5.9 to 12.8~ms (+116.9\%). Despite this substantial computational increase,
aggregate box mAP$_{50}$ decreases slightly from 0.665 for YOLO26m to
0.660 for YOLO26x, mask mAP$_{50}$ decreases from 0.665 to 0.658,
and mask mAP$_{50:95}$ decreases from 0.395 to 0.382. Thus, scaling from
the medium to extra-large YOLO26 configuration does not produce a
commensurate accuracy improvement under the 960-pixel setting.}

\textcolor{black}{This result is particularly important for the three anatomical classes.
Additional model capacity does not remove the fundamental scale imbalance
between fruitlet, calyx, and peduncle. Fruitlet is comparatively large and
already reaches high segmentation accuracy in moderate-capacity models;
therefore, additional parameters increasingly encounter performance
saturation. Calyx benefits from improved fine-scale representation but does
not exhibit a monotonic relationship with network size. Peduncle remains
the most difficult structure because its thin elongated geometry, small
pixel footprint, occlusion, and visual similarity to surrounding stems
create a spatial-resolution problem that cannot be solved solely by adding
network capacity. Consequently, the results indicate that preserving
fine-scale image information can be more valuable for peduncle perception
than simply increasing the number of parameters.}

\begin{table}[ht!]
\centering
\caption{\textbf{Training duration and convergence behavior of all
configurations.} ``--'' indicates that the full 200-epoch schedule was
completed and no early-stopping best epoch was reported in the corresponding
summary. YOLO26 completed all 200 epochs in every configuration.}
\label{tab:training_efficiency_all}
\setlength{\tabcolsep}{4pt}
\renewcommand{\arraystretch}{1.08}
\scriptsize
\begin{tabular}{llrrrrrr}
\toprule
\textbf{Generation} & \textbf{Model} &
\multicolumn{3}{c}{\textbf{640$\times$640}} &
\multicolumn{3}{c}{\textbf{960$\times$960}}\\
\cmidrule(lr){3-5}\cmidrule(lr){6-8}
& & \textbf{Completed} & \textbf{Best} & \textbf{Time (h)}
& \textbf{Completed} & \textbf{Best} & \textbf{Time (h)}\\
\midrule

\multirow{5}{*}{YOLO26}
& n & 200 & -- & 0.933 & 200 & -- & 1.073\\
& s & 200 & -- & 1.019 & 200 & -- & 1.076\\
& m & 200 & -- & 0.998 & 200 & -- & 1.257\\
& l & 200 & -- & 1.051 & 200 & -- & 1.409\\
& x & 200 & -- & 1.034 & 200 & -- & 2.124\\
\midrule

\multirow{5}{*}{YOLO11}
& n & 200 & -- & 0.462 & 163 & 63 & 0.685\\
& s & 200 & -- & 0.431 & 161 & 61 & 0.838\\
& m & 148 & 48 & 0.373 & 197 & 97 & 0.939\\
& l & 200 & -- & 0.586 & 168 & 68 & 0.931\\
& x & 200 & -- & 0.836 & 158 & 58 & 1.391\\
\midrule

\multirow{5}{*}{YOLOv8}
& n & 200 & -- & 0.462 & 200 & -- & 0.792\\
& s & 200 & -- & 0.369 & 183 & 83 & 0.724\\
& m & 200 & -- & 0.406 & 147 & 47 & 0.615\\
& l & 131 & 31 & 0.332 & 137 & 37 & 0.791\\
& x & 140 & 40 & 0.500 & 149 & 49 & 1.214\\

\bottomrule
\end{tabular}
\end{table}

\textcolor{black}{Training efficiency further distinguishes the three generations
(Table~\ref{tab:training_efficiency_all}). All ten YOLO26 runs completed
the full 200-epoch schedule; therefore, no early termination was observed
under the configured patience criterion. At 640 pixels, YOLO26 training
duration remains surprisingly compact across scale, ranging from 0.933~h
for YOLO26n to 1.051~h for YOLO26l, with YOLO26x requiring 1.034~h.
At 960 pixels, however, the effect of model scale becomes more pronounced:
training requires 1.073, 1.076, 1.257, 1.409, and 2.124~h for n, s, m, l,
and x, respectively. YOLO26x therefore requires almost twice the
wall-clock training time of YOLO26s at 960 pixels without providing a
corresponding increase in aggregate detection or segmentation accuracy.}

\textcolor{black}{YOLOv8 and YOLO11 exhibit substantially different convergence behavior.
Several runs reach the early-stopping condition before the 200-epoch
maximum. For YOLOv8 at 640 pixels, the l and x configurations terminate
after 131 and 140 epochs, with best epochs of 31 and 40, respectively.
At 960 pixels, YOLOv8s, m, l, and x terminate after 183, 147, 137, and
149 epochs, with best epochs of 83, 47, 37, and 49. YOLO11 shows an even
stronger resolution-dependent convergence pattern: all five 960-pixel
models terminate early, at 163, 161, 197, 168, and 158 epochs for n through
x, respectively. Because patience was 100 epochs, these completed-epoch
counts are consistent with the corresponding best epochs reported in the
training records.}

\textcolor{black}{An early termination should not be interpreted as inferior training.
Instead, it indicates that the monitored validation criterion failed to
improve sufficiently during the configured patience interval. In this
context, the best epoch identifies the checkpoint retained for subsequent
evaluation, while the completed epoch identifies when training was
terminated. For example, YOLO11s-960 achieves its best checkpoint at
epoch 61 and terminates at epoch 161. This behavior is particularly useful
from a training-efficiency perspective because continuing to epoch 200
would consume additional computation without improving the retained model.}

\textcolor{black}{The lowest recorded wall-clock training duration is 0.332~h for
YOLOv8l-640; however, this value must be interpreted together with its
early termination at epoch 131 and therefore cannot be directly interpreted
as the fastest architecture for a fixed 200-epoch workload. Similarly,
YOLO11m-640 requires only 0.373~h because it terminates after 148 epochs.
In contrast, the largest recorded duration is 2.124~h for YOLO26x-960,
which completes all 200 epochs. Training duration therefore reflects both
per-epoch computational demand and the number of epochs actually completed,
and comparisons based solely on wall-clock duration can be misleading
without convergence information.}

\textcolor{black}{Taken together, model complexity, processing latency, training behavior,
and predictive accuracy reveal a clear accuracy--efficiency frontier rather
than a monotonic advantage for increasingly large models. Nano models
provide the lowest memory and computational requirements and are therefore
attractive when processing resources are severely constrained. Their
limitation is primarily predictive capacity for difficult small anatomical
structures, particularly peduncle. Extra-large models occupy the opposite
extreme: they provide high representational capacity but require
approximately 60--72~M parameters, 296--328~GFLOPs, checkpoints exceeding
120~MB, and substantially greater high-resolution inference time. The
accuracy analyses show that this additional computational burden does not
consistently translate into superior aggregate or class-wise performance.}

\textcolor{black}{The small and medium configurations therefore occupy the most practically
important operating region. In particular, the preceding resolution
analysis showed that the small models respond strongly to 960-pixel
small-object-focused optimization. This is especially relevant for calyx
and peduncle, for which preserving spatial detail is more consequential
than increasing network width alone. YOLO26s-960, for example, requires
only 10.367~M parameters, 34.1~GFLOPs, a 23.4~MB checkpoint, and
approximately 4.0~ms total validation processing time while producing
substantial improvements in the small-object-focused performance measures.
This represents only approximately one-sixth of the parameters and
one-ninth of the GFLOPs of YOLO26x, while avoiding the strong diminishing
returns observed at the extra-large scale.}

\textcolor{black}{YOLO26m represents a second relevant operating point when additional
representational capacity is available. Its 23.510~M parameters and
121.2~GFLOPs remain substantially below the 62.730~M parameters and
313.0~GFLOPs required by YOLO26x. At 960 pixels, its total measured
processing time is 6.9~ms compared with 13.7~ms for YOLO26x, yet its
aggregate mask mAP$_{50:95}$ is slightly higher (0.395 versus 0.382).
Thus, for this dataset, moving from YOLO26m to YOLO26x more than doubles
the inference burden and increases parameter count by approximately
2.67$\times$ without producing a meaningful accuracy advantage.}

\textcolor{black}{From the perspective of robotic green-fruit thinning, this distinction is
critical. Fruitlet localization is comparatively mature across many of the
evaluated architectures, whereas reliable calyx and especially peduncle
segmentation requires preservation of fine spatial information. The
computational results therefore favor models that allocate sufficient
capacity to exploit higher-resolution inputs without entering the
high-cost saturation regime of the largest networks. Lightweight
YOLO26n configurations are appropriate where memory, power, or inference
resources are highly constrained; YOLO26s-960 provides a particularly
strong candidate for high-speed small-object-aware perception; and
YOLO26m-960 provides a higher-capacity alternative while retaining a
substantially more favorable computational profile than extra-large
architectures. Large and extra-large models may remain useful when
computational resources are abundant, but the present results do not
support selecting them solely on the assumption that increased network
capacity will produce superior anatomical-level orchard perception.}

\textcolor{black}{Overall, the experiments demonstrate that computational scale and predictive
performance are only partially coupled. Higher layer counts, parameter
counts, GFLOPs, checkpoint sizes, and inference times represent increased
computational capacity and cost, but their highest values do not correspond
systematically to the highest detection or segmentation accuracy. Conversely,
the lowest-complexity architecture is computationally attractive but does
not necessarily provide sufficient representation for the most difficult
thin anatomical targets. The most favorable deployment region therefore
lies between these extremes, where increased spatial resolution can be
exploited by small-to-medium networks without incurring the disproportionate
computational cost and diminishing accuracy returns associated with
extra-large configurations.}

\section{Conclusion}

\textcolor{black}{This study established a comprehensive cross-generation benchmark for fine-grained small-object detection and instance segmentation under a challenging real-world green-on-green orchard condition. Using an early-stage apple fruitlet dataset, YOLOv8, YOLO11, and YOLO26 were systematically compared across five model scales: nano (n), small (s), medium (m), large (l), and extra-large (x) and two training configurations, resulting in 15 architectures and 30 model configuration experiments. The conventional configuration used $640\times640$-pixel inputs, whereas the small-object-focused $960\times960$ configuration combined increased spatial resolution with targeted scale- and mosaic-related training controls. Beyond conventional whole-fruit perception, the benchmark considered three anatomically distinct targets---fruitlet, calyx, and peduncle---providing a progressively more demanding evaluation from comparatively larger fruit bodies to small, thin, and visually ambiguous attachment structures.}

\textcolor{black}{The results demonstrated that neither progression toward newer YOLO generations nor increasing network capacity consistently produced higher predictive performance. Instead, performance reflected interactions among model generation, network scale, training configuration, and target morphology. Fruitlet generally exhibited the strongest detection and segmentation performance, whereas peduncle remained the principal perception bottleneck because of its limited pixel footprint, elongated geometry, occlusion, weak foreground--background contrast, and comparatively limited representation in the training dataset. The small-object-focused $960\times960$ configuration generally produced stronger stringent box and mask mAP$_{50:95}$ values, particularly for several compact model variants. YOLO11s-960 achieved the highest observed aggregate mask mAP$_{50:95}$ of 0.402, while YOLO26s-960 achieved a closely comparable value of 0.397 using 10.37~M parameters and 34.1~GFLOPs. In contrast, YOLO26x-960 required 62.73~M parameters, 313~GFLOPs, and 12.8~ms inference time without exceeding the aggregate segmentation performance of substantially smaller configurations. These findings indicate that increasing network capacity alone may provide diminishing returns and that compact-to-moderate architectures combined with small-object-focused training can offer favorable accuracy--efficiency trade-offs for fine-grained orchard perception. Given the single-run experimental design, small numerical differences among closely performing configurations should be interpreted as observed benchmark outcomes rather than evidence of statistical superiority.}

\textcolor{black}{Several limitations define the scope of these findings. The dataset comprised 600 images acquired from a single commercial orchard in Prosser, Washington, on May 6, 2024, using an iPhone 14 Pro, and therefore represents one location, acquisition period, phenological stage, and imaging platform rather than the full variability of commercial orchards. The dataset was also class-imbalanced, with substantially fewer peduncle annotations than fruitlet and calyx instances, which may have contributed to the lower peduncle performance. Furthermore, the 30 model--configuration experiments were evaluated using single training runs because of their computational scale, and computational performance was measured on a GPU workstation rather than through onboard robotic deployment. Future work should therefore expand the dataset across multiple orchards, cultivars, seasons, phenological stages, illumination conditions, and sensing platforms; increase annotation of visible and occluded peduncles; conduct repeated-seed experiments to quantify performance variability; and validate selected models on embedded computing hardware and operational robotic thinning platforms. Accordingly, the present results should be interpreted as a rigorous benchmark under a particularly challenging real-world orchard condition and as a foundation for broader multi-environment validation.}

\section*{Acknowledgement}
This work was supported in part by the National Science Foundation (NSF) and the United States Department of Agriculture (USDA), National Institute of Food and Agriculture (NIFA), through the “Artificial Intelligence (AI) Institute for Agriculture” program under Award Numbers AWD003473 and AWD004595, and USDA-NIFA Accession Number 1029004 for the project titled “Robotic Blossom Thinning with Soft Manipulators.” Additional support was provided through USDA-NIFA Grant Number 2024-67022-41788, Accession Number 1031712, under the project “ExPanding UCF AI Research To Novel Agricultural EngineeRing Applications (PARTNER)”. We acknowledge Allan Brothers Fruit Company for granting access to their commercial orchards across Washington State, USA. Special thanks to Astrid Wimmer, Randall Cason, Diego Lopez, Giulio Diracca, and Priyanka Upadhyaya for their
invaluable efforts in data preparation and logistical support throughout this project.
\scriptsize
\bibliographystyle{unsrt}  
\bibliography{references}

\end{document}